%% file: _main.tex
\input{_constants}

\arxiv % \review OR \arxiv OR \cameraready
\pdfoutput=1
\documentclass[10pt,twocolumn,letterpaper]{article}
\input{format_header}

\begin{document}
%% TITLE
\title{\paperTitle}
\author{\authorBlock}
\maketitle

\input{00_abstract}

\input{01_intro}

\input{02_related_work}

\input{03_our_method}

\input{04_quality_evaluation}
\input{05_conclusion}

\balance
{\small
\bibliographystyle{ieee_fullname}
\bibliography{_references}
}

\ifarxiv \clearpage \input{_appendix} \fi
\end{document}

%% file: _constants.tex
\def\cvprPaperID{}
\def\confName{}
\def\confYear{}

\def\paperTitle{Composite Diffusion \\ $whole >= \Sigma parts$}

\def\authorBlock{

Vikram Jamwal\\
TCS Research, India\\
{\tt\small vikram.jamwal@tcs.com}

\and
Ramaneswaran S \thanks{Work performed while working at TCS Research. }\footnotemark[1] \\
NVIDIA, India\\
{\tt\small ramanr@nvidia.com}

}

\newif\ifreview 
\newif\ifarxiv \newcommand{\arxiv}{\arxivtrue}
\newif\ifcamera 
\newif\ifrebuttal

%% file: format_header.tex
\ifarxiv \usepackage[pagenumbers]{format} \fi

\usepackage{graphicx}
\usepackage{amsmath}
\usepackage{amssymb}
\usepackage{booktabs}

\input{_macros}  % Add packages to _macros.tex

\usepackage{xr-hyper}

\makeatletter
\newcommand*{\addFileDependency}[1]{
  \typeout{(#1)}
  \@addtofilelist{#1}
  \IfFileExists{#1}{}{\typeout{No file #1.}}
}

\makeatother

\usepackage[pagebackref,breaklinks,colorlinks]{hyperref}
\usepackage[capitalize]{cleveref}
\crefname{section}{Sec.}{Secs.}
\crefname{table}{Table}{Tables}
\crefname{figure}{Fig.}{Figs.}

%% file: _macros.tex
\usepackage{times}
\usepackage{microtype}
\usepackage{epsfig}
\usepackage[table,xcdraw]{xcolor}
\usepackage{caption}
\usepackage{float}
\usepackage{placeins}
\usepackage{color, colortbl}
\usepackage{stfloats}
\usepackage{enumitem}
\usepackage{tabularx}
\usepackage{xstring}
\usepackage{multirow}
\usepackage{xspace}
\usepackage{url}
\usepackage{subcaption}
\usepackage[hang,flushmargin]{footmisc}

\usepackage{graphicx}
\usepackage{amsmath}
\usepackage{amssymb}
\usepackage{booktabs}
\usepackage[linesnumbered]{algorithm2e}
\RestyleAlgo{ruled}

\usepackage{lmodern,babel,adjustbox,booktabs,multirow}
\usepackage{bm}
\usepackage{balance}

\ifarxiv  \fi

\usepackage[pagebackref,breaklinks,colorlinks]{hyperref}

\usepackage[capitalize]{cleveref}
\crefname{section}{Sec.}{Secs.}
\Crefname{section}{Section}{Sections}
\Crefname{table}{Table}{Tables}
\crefname{table}{Tab.}{Tabs.}

\usepackage{amssymb,pifont}
\SetAlFnt{\small}
\SetAlCapFnt{\small}
\SetAlCapNameFnt{\small}
\SetAlCapHSkip{0pt}

\SetCommentSty{mycommfont}
\SetNlSty{mycommfont}{}{}
\SetKwComment{tcp}{$\triangleleft$\ }{}

%% file: 00_abstract.tex
\begin{abstract}
For an artist or a graphic designer, the spatial layout of a scene is a critical design choice. However, existing text-to-image diffusion models provide limited support for incorporating spatial information. This paper introduces \textbf{Composite Diffusion} as a means for artists to generate high-quality images by composing from the sub-scenes. The artists can specify the arrangement of these sub-scenes through a flexible free-form segment layout. They can describe the content of each sub-scene primarily using natural text and additionally by utilizing reference images or control inputs such as line art, scribbles, human pose, canny edges, and more.

We provide a comprehensive and modular method for Composite Diffusion that enables alternative ways of generating, composing, and harmonizing sub-scenes. Further, we wish to evaluate the composite image for effectiveness in both image quality and achieving the artist's intent. We argue that existing image quality metrics lack a holistic evaluation of image composites. To address this, we propose novel quality criteria especially relevant to composite generation. 

We believe that our approach provides an intuitive method of art creation. Through extensive user surveys, quantitative and qualitative analysis, we show how it achieves greater spatial, semantic, and creative control over image generation. In addition,  our methods do not need to retrain or modify the architecture of the base diffusion models and can work in a plug-and-play manner with the fine-tuned models.
\end{abstract}

%% file: 01_intro.tex
\section{Introduction}
\label{sec:introduction}

\begin{figure}[t!]
    \centering
   
    \includegraphics[width = \linewidth]{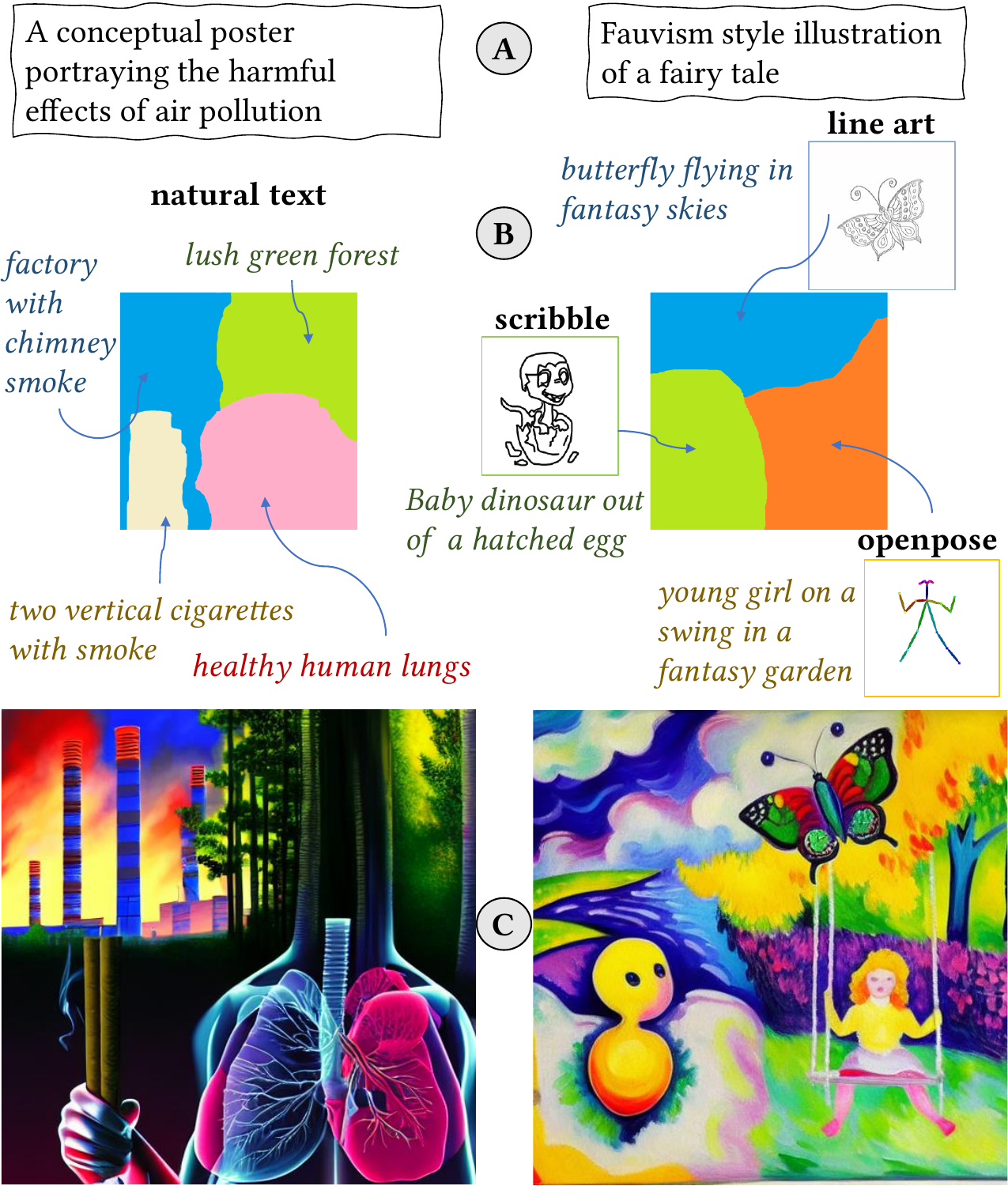}
    \caption {Image generation using Composite Diffusion: The artist's intent (A) is manually converted into input specification (B) for the model in the form of a \textit{free-form sub-scene layout} and conditioning information for each sub-scene. The conditioning information can be  \textit{natural text description}, and any other \textit{control condition}. The model generates composite images (C) based on these inputs.} 
    
    \label{fig:teaser}
    
\end{figure}

Recent advances in diffusion models \cite{diffusion_models_dhariwal}, such as Dalle-2 \cite{dalle}, Imagen \cite{imagegen}, and Stable Diffusion \cite{stable_diffusion} have enabled artists to generate vivid imagery by describing their envisioned scenes with natural language phrases.
However, it is cumbersome and occasionally even impossible to specify spatial information or sub-scenes within an image solely by text descriptions. Consequently, artists have limited or no direct control over the layout, placement, orientation, and properties of the individual objects within a scene. These creative controls are indispensable for artists seeking to express their creativity \cite{solidarnyh2023} and are crucial in various content creation domains, including illustration generation, graphic design, and advertisement production. Frameworks like Controlnets \cite{zhang2023adding} offer exciting new capabilities by training parallel conditioning networks within diffusion models to support numerous control conditions. Nevertheless, as we show in this paper, creating a complex scene solely based on control conditions can still be challenging.  As a result, achieving the desired imagery may require several hours of labor or maybe only be partially attainable through pure text-driven or control-condition-driven techniques.

To overcome these challenges, we propose \textbf{Composite-Diffusion}  as a method for creating composite images by combining spatially distributed segments or sub-scenes. These segments are generated and harmonized through independent diffusion processes to produce a final composite image. The \textit{artistic intent} in Composite Diffusion is conveyed through the following two means:

\textbf{(i) Spatial Intent:} Artists can flexibly arrange sub-scenes using a free-form spatial layout. A unique color identifies each sub-scene.

\textbf{(ii) Content intent:}Artists can specify the desired content within each sub-scene through text descriptions. They can augment this information by using examples images and other control methods such as scribbles, line drawings, pose indicators, etc.

We believe, and our initial experience has shown, that this approach offers a powerful and intuitive method for visual artists to stipulate their artwork.

This paper seeks to answer two primary research questions: First, how can native diffusion models facilitate composite creation using the diverse input modalities we described above? Second, how do we assess the quality of images produced using Composite Diffusion methods? Our paper \textbf{contributes} in the following novel ways:

\textbf{1.}We present a comprehensive, modular, and flexible method for creating composite images, where the individual segments (or sub-scenes) can be influenced not only by textual descriptions, but also by various control modalities such as line art, scribbles, human pose, canny images, and reference images. The method also enables the simultaneous use of different control conditions for different segments. 

\textbf{2.} Recognizing the inadequacy of existing image quality metrics such as FID (Frechet Inception Distance) and Inception Scores \cite{FID_NIPS, IS_NIPS} for evaluating the quality of composite images, we introduce a new set of quality criteria. While principally relying  on human evaluations for quality assessments,  we also develop new methods of automated evaluations suitable for these quality criteria.

We rigorously evaluate our methods using various techniques including quantitative user evaluations, automated assessments, artist consultations, and qualitative visual comparisons with alternative approaches. In the following sections, we delve into related work (Section \ref{sec:related_work}), detail our method (Section \ref{sec:our_method}), and discuss the evaluation and implications of our approach (Section \ref{sec:quality_criteria}, and \ref{sec:conclusion}).

%% file: 02_related_work.tex
\section{Related work}
\label{sec:related_work}
In this section, we discuss the approaches that are related to our work from multiple perspectives.

\subsection{Text-to-Image generative models} The field of text-to-image generation has recently seen rapid advancements, driven primarily by the evolution of powerful neural network architectures. Approaches like DALL·E \cite{dalle} and VQ-GAN \cite{vqgan} proposed a two-stage method for image generation. These methods employ a discrete variational auto-encoder (VAE) to acquire comprehensive semantic representations, followed by a transformer architecture to autoregressively model text and image tokens. Subsequently, diffusion-based approaches, such as Guided Diffusion \cite{openai-guided-diff} \cite{diffusion_models_dhariwal}, have showcased superior image sample quality compared to previous GAN-based techniques. Dalle-2 \cite{dalle2} and Imagen \cite{imagegen} perform the diffusion process in the pixel-image space while Latent Diffusion Models such as Stable Diffusion \cite{stable_diffusion} perform the diffusion process in a more computationally suitable latent space.  However, in all these cases, relying on single descriptions to depict complex scenes restricts the level of control users possess over the generation process.

\begin{figure*}[t!]
\centering
\includegraphics[width= \textwidth]{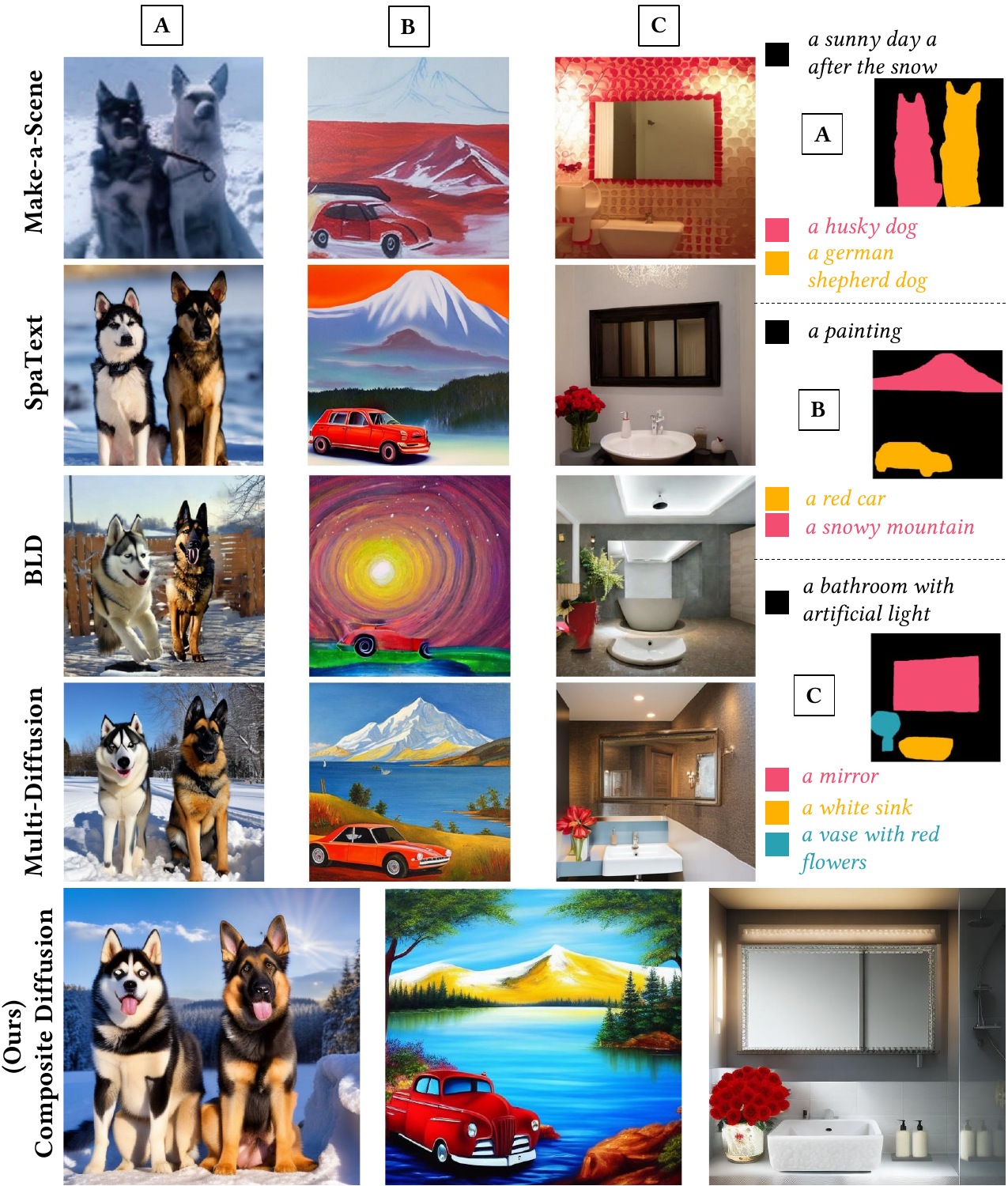}
\caption{The figure provides a visual comparison of the outputs of Composite Diffusion with other related approaches - using the same segment layouts and text prompts. Note that these input specifications are from the related-work literature. Given a choice, our approach to creating segment layout and text prompts would vary slightly - we would partition the image space into distinct \textit{sub-scenes} that fully partition the image space, and we will not have background masks or prompts.}
\label{fig:relevant_comparison}
\end{figure*}

\subsection{Spatial control models} Some past works on image generation have employed segments for spatial control but were limited to domain-specific segments. For example, GauGAN \cite{gaugan} introduced spatially-adaptive normalization to incorporate semantic segments to generate high-resolution images. PoE-GAN \cite{poe_gan} utilized the product of experts method to integrate semantic segments and a global text prompt to enhance the controllability of image generation. However, both approaches rely on GAN architectures and are constrained to specific domains with a fixed segment vocabulary. Make-A-Scene \cite{make_a_scene} utilized an optional set of dense segmentation maps, along with a global text prompt, to aid in the spatial controllability of generation. VQ-GAN \cite{vqgan} can be trained to use semantic segments as inputs for image generation. No-Token-Left-Behind \cite{paiss2022token} employed explainability-based methods to implement spatial conditioning in VQ-GAN; they propose a method that conditions a text-to-image model on spatial locations using an optimization approach. The approaches discussed above are also limited by training only on a fixed set of dense segments.

\subsection{Inpainting} 
\label{sec:rw_inpainting}
The work that comes closest to our approach in diffusion models is in-painting. Almost all the popular models \cite{dalle2}, \cite{imagegen},  \cite{stable_diffusion} support some form of inpainting. The goal of inpainting is to modify a portion in an image specified by a segment-mask (and optional accompanying textual description) while retaining the information outside the segment. Some of the approaches for inpainting in the recent past include repaint \cite{repaint}, blended-diffusion\cite{avrahami2022blended}, and latent-blended diffusion \cite{latent_blended}. RunwayML \cite{stable_diffusion} devises a specialized model for in-painting in Stable Diffusion, by modifying the architecture of the UNet model to include special masked inputs.  As we show in later this paper, one can conceive of an approach for Composite Diffusion using inpainting, where we can perform inpainting for each segment in a serial manner (refer to Appendix  \ref{sec:serial_inpainting}). However, as we explain in this paper, a simple extension of localized in-painting methods for multi-segment composites presents some drawbacks. 

\subsection{Other diffusion-based composition methods}
Some works look at the composition or editing of images through a different lens. These include prompt-to-prompt editing \cite{hertz2022prompt, mokady2022null}, composing scenes through composable prompts \cite{liu2022compositional}, and methods for personalization of subjects in a generative model \cite{ruiz2022dreambooth}.  Composable Diffusion \cite{Composable-Diffusion} takes a structured approach to generate images where separate diffusion models generate distinct components of an image. As a result, they can generate more complex imagery than seen during the training. Composed GLIDE \cite{liu2022compositional} is a composable diffusion implementation that builds upon the GLIDE model \cite{glide} and utilizes compositional operators to combine textual operations. Dreambooth \cite{ruiz2022dreambooth} allows the personalization of subjects in a text-to-image diffusion model through fine-tuning. The learned subjects can be put in totally new contexts such as scenes, poses, and lighting conditions. Prompt-to-prompt editing techniques \cite{couairon2022diffedit, hertz2022prompt, mokady2022null} exploit the information in cross-attention layers of a diffusion model by pinpointing areas that spatially correspond to particular words in a prompt. These areas can then be modified according to the change of the words in the prompt. 
Our method is complementary to these advances. We concentrate specifically on composing the spatial segments specified via a spatial layout. So, in principle, our methods can be supplemented with these capabilities (and vice versa). 

\subsection{Spatial layout and natural text-based models }
In this section, we discuss three related concurrent works: SpaText \cite{Avrahami_2023_CVPR}, eDiffi \cite{balaji2023ediffi}, and Multi-diffusion \cite{bar2023multidiffusion}.  All these works provide some method of creating images from spatially free-form layouts with natural text descriptions. 

SpaText \cite{Avrahami_2023_CVPR} achieves spatial control by training the model to be space-sensitive by additional CLIP-based spatial-textual representation. The approach requires the creation of a training dataset and extensive model training, both of which are costly. Their layout schemes differ slightly from ours as they are guided towards creating outlines of the objects, whereas we focus on specifying the sub-scene. 

eDiffi \cite{balaji2023ediffi} proposes a method called paint-with-words which exploits the cross-attention mechanism of U-Net in the diffusion model to specify the spatial positioning of objects. Specifically, it associates certain phrases in the global text prompt with particular regions by manipulating the cross-attention matrix. Similar to our work, they do not require pre-training for a segment-based generation. However, they must create an explicit control for the objects in the text description for spatial control. We use the inherent capability of U-net's cross-attention layers to guide the relevant image into the segments through step-inpainting and other techniques.

Multi-diffusion \cite{bar2023multidiffusion} proposes a mechanism for controlling the image generation in a region by providing the abstraction of an optimization loss between an ideal output by a single diffusion generator and multiple diffusion processes that generate different parts of an image. It also provides an application of this abstraction to segment layout and natural-text-based image generation. This approach has some similarities to ours in that they also build their segment generation by step-wise inpainting. They also use bootstrapping to anchor the image and then use the later stages for blending. However, our approach is more generic, has a wider scope, and is more detailed.
For example, we don't restrict the step composition to a particular method. Our scaffolding stage has a much wider significance as our principal goal is to create segments independent of each other, and the goal of the harmonization stage is to create segments in the context of each other. We provide alternative means of handling both the scaffolding and harmonization stages. 

Further,  in comparison to all the above approaches, we achieve \textit{additional control over the orientation and placement of objects within a segment} through reference images and control conditions specific to the segment.

%% file: 03_our_method.tex
\section{Our Composite Diffusion method}
\label{sec:our_method}
We present our method for Composite Diffusion. It can directly utilize a pre-trained text-conditioned diffusion model or a control-conditioned model without the need to retrain them. We first formally define our goal. We will use the term \textit{`segment'} particularly to denote a \textit{sub-scene}.

\subsection{Goal definition} 
\label{sec:goal-definition}
We want to generate an image $\mathbf{x}$ which is composed entirely based on two types of input specifications: 
\begin{enumerate}
    \item \textbf{Segment Layout}: a set of free-form segments $S=[s^1, s^2, ..., s^n]$, and 
    \item \textbf{Segment Content}: a set of natural text descriptions, $D = [d^1, d^2, ..., d^n]$, and optional additional control conditions, $C = [c^1, c^2, ..., c^n]$. 
\end{enumerate}

Each segment $s^j$ in $S$ describes the spatial form of a sub-scene and has a corresponding natural text description $d^j$ in $D$, and optionally a corresponding control condition $c^j$ in $C$. The segments don't non-overlap and fully partition the image space of $\mathbf{x}$. Additionally, we convert the segment layout to  segment-specific masks, $M=[m^1, m^2, ..., m^n]$, as one-hot encoding vectors. The height and width dimensions of the encoding vector are the same as that of $\mathbf{x}$. `$1$s' in the encoded mask vector indicate the presence of image pixels corresponding to a segment, and `$0$'s indicate the absence of pixel information in the complementary image area (Refer to Appendix Figure \ref{fig:running-example}).

Our method divides the generative process of a diffusion model into two successive temporal stages: (a) the Scaffolding stage and (b) the Harmonization stage. We explain these stages below:

\begin{figure}[ht!]
\label{fig:serial}
    \centering
    \fbox{
    \includegraphics[width= .96\columnwidth]{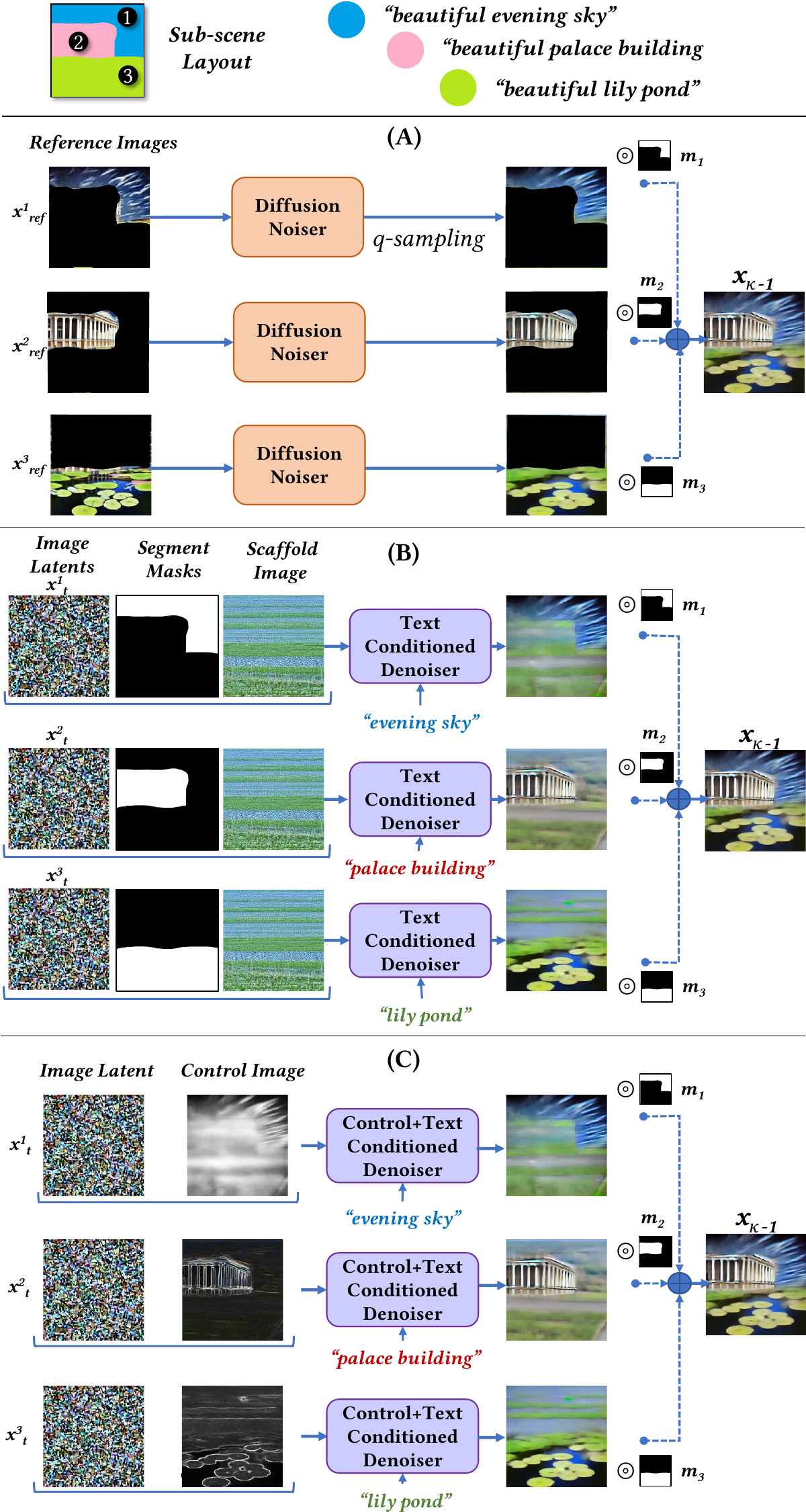}}
    \caption {Scaffolding stage step for three different cases: (A) with reference images, (B) with a scaffolding image, and (C) with control conditions. Please note that for case (A), the \textit{diffusion noising} process is only a single step, while for cases (B) and (C), the \textit{diffusion denoising} process repeats for each time step till the end of scaffolding stage at $t=\kappa$. All the segments develop independently of each other. The individual segments are composed to form an intermediate composite only at the end of the scaffolding stage.}
    \label{fig:scaff-all-conditions}
\end{figure}

\begin{algorithm}[ht!]
\caption{Composite Diffusion: Scaffolding Stage. The input is as defined in the section \ref{sec:goal-definition}.}
\label{alg:scaff}
\uIf{Segment Reference Images} 
{
\For{\textbf{all segments} $i$  \textbf{from} $1$ to $n$}
    {
     $x_{\kappa-1}^{seg_i} \leftarrow Noise(x^{ref_i}, \kappa) $ \tcp*{Q-sample reference images to last timestep of scaffolding stage. }  
    }
}
\uElseIf{Only Segment Text Descriptions}
{ 
    \For {\textbf{all} $t$ \textbf{from} $T$ to $\kappa$}
    {
     \For{\textbf{all segments} $i$  \textbf{from} $1$ to $n$} 
        {
            $x_{t}^{scaff}     \leftarrow Noise(x^{scaff}, t) $ \tcp*{Q-sample scaffold. }    

            % $x_{t-1}^{seg_i} \leftarrow Denoise(x_t \odot m^i + x_{t}^{scaff}  \odot (1 -  m^i),  d^i)$ \tcp*{Scaffold with the Scaffolding Image and denoise}
            $x_{t-1}^{seg_i} \leftarrow Denoise(x_t, x_{t}^{scaff}, m^i, d^i)$ \tcp*{Step-inpaint with the scaffolding image. }
        }
    }       
}   
\uElseIf{Text and Segment Control Conditions}
{
    \For {\textbf{all} $t$ \textbf{from} $T$ to $\kappa$}  
    { 
        \For{\textbf{all segments} $i$  \textbf{from} $1$ to $n$} 
        {
            $x_{t-1}^{seg_i} \leftarrow Denoise(x_t , m^i, d^i, c^i)$ \tcp*{ Scaffold with the control condition and denoise.}
        }
    }
}
$x^{comp}_{\kappa-1} \leftarrow \sum_{i=1}^{n} x_{\kappa-1}^{seg_i} \odot m^i$ \tcp*{Merge segments.}
\Return  $x^{comp}_{\kappa-1}$
\end{algorithm}

\begin{figure*}[t!]
    \centering
    \includegraphics[width = \linewidth]{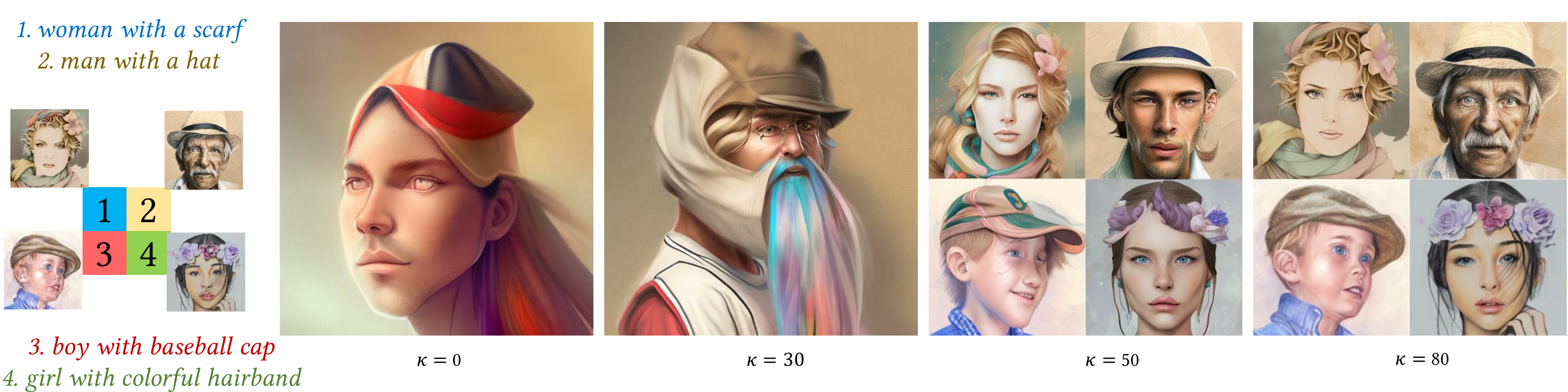}
    \caption {Use of reference images for scaffolding: The scaffolding factor ($\kappa$) (Section \ref{sec:scaffolding_factor}) controls the influence of reference images on the final composite image. At low  $\kappa$ values, the reference images are heavily noised and exercise little control; the segments merge drastically. At high $\kappa$ values, the reference images are lightly noised and the resulting image is nearer to the reference images. A middle $\kappa$ value balances the influences of reference images and textual descriptions.}
    \label{fig:reference-images}
\end{figure*}

\subsection{Scaffolding stage}
We introduce the concept of \textit{scaffolding}, which we define as a mechanism for guiding image generation within a segment with some external help. We borrow the term `scaffolding' from the construction industry \cite{scaffolding_2022_ref}, where it refers to the temporary structures that facilitate the construction of the main building or structure. These scaffolding structures are removed in the building construction once the construction work is complete or has reached a stage where it does not require external help. Similarly, we may drop the scaffolding help after completing the scaffolding stage.

The external structural help, in our case, can be provided by any means that help generate or anchor the appropriate image within a segment. We provide this help through either (i) \textit{scaffolding reference image} - in the case where reference example images are provided for the segments, (ii) \textit{a scaffolding image} - in the case where only text descriptions are available as conditioning information for the segments, or (iii) a \textit{scaffolding control condition} - in the case where the base generative model supports conditioning controls and additional control inputs are available for the segments.

\subsubsection{Segment generation using a scaffolding reference image}
An individual segment may be provided with an example image called \textit{scaffolding reference image} to gain specific control over the segment generation. This conditioning is akin to using image-to-image translation \cite{stable_diffusion} to guide the production of images in a particular segment. 

Algorithmically, we directly noise the reference image (refer to Q-sampling in Appendix \ref{sec:forward_diff}) to the time-stamp $t=\kappa$ that depicts the last time-step of the scaffolding stage in the generative diffusion process (Algo. \ref{alg:scaff}, 1-4, and Fig. \ref{fig:scaff-all-conditions}, A).  The generated segment can be made more or less in the likeness of the reference image by varying the initializing noising levels of the reference images. Refer to Fig. \ref{fig:reference-images} for an example of scaffolding using segment-specific reference images.

\begin{figure*}[htp!]
    \centering
    \fbox{
    \includegraphics[width=.98\textwidth]{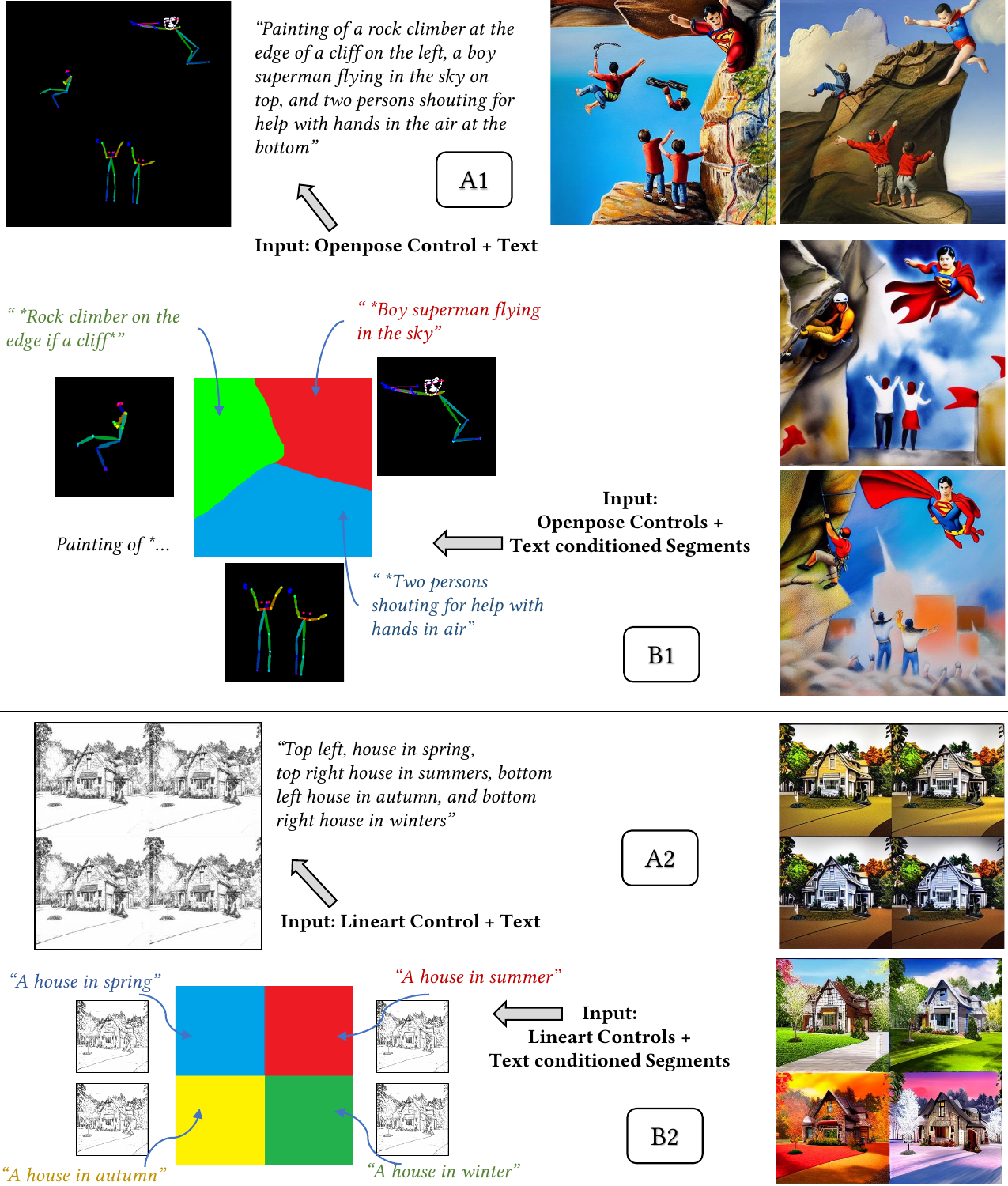}}
    \caption {\textit{Control+Text} conditioned composite generations: For the two cases shown in the figure, getting correct compositions is extremely difficult with text-to-image models or even (text+control)-to-image models (For example, in A1 the image elements don't cohere, and in A2 the fours seasons do not show in the output image). Composite Diffusion with \textit{scaffolding control conditions} can effectively influence sub-scene generations and create the desired overall composite images(B1, B2). }
    \label{fig:control-seg-examples}
\end{figure*}

\subsubsection{Segment generation with a scaffolding image} 
This case is applicable when we have only text descriptions for each segment. The essence of this method is the use of a predefined image called \textit{scaffolding image} ($x^{scaff}$), to help with the segment generation process. Refer to Algo. \ref{alg:scaff}, 5-11 and Fig. \ref{fig:scaff-all-conditions}, B. 

Algorithmically, to generate one segment at any timestep $t$ : (i) we apply the segment mask $m$ to the noisy image latent $x_t$ to isolate the area  $x_t \odot m$ where we want generation, (ii) we apply a complementary mask $(1-m)$ to an appropriately noised (q-sampled to timestep $t$) version of scaffold image $x^{scaff}_{t}$  to isolate a complementary area $x^{scaff}_{t} \odot (1-m)$, and (iii) we merge these two complementary isolated areas and denoise the composite directly through the denoiser along with the corresponding textual description for the segment. Refer to Appendix \ref{sec:our_method_implementation} Fig. \ref{fig:our-algo}(a) for an illustration of the single-step generation. We then replicate this process for all the segments. 

These steps are akin to performing an inpainting \cite{latent_blended} step on each segment but in the context of a scaffolding image. Please note that our method step (Algo. \ref{alg:scaff}, 9) is generic and flexible to allow the use of any inpainting method, including the use of a specially trained model (e.g., RunwayML Stable Diffusion inpainting 1.5 \cite{stable_diffusion}) that can directly generate inpainted segments.   

We repeat this generative process for successive time steps till the time step $t=\kappa$. The choice of scaffolding image can be arbitrary. Although convenient, we do not restrict keeping the same scaffolding image for every segment.

\subsubsection{Segment generation with a scaffolding control} 
This case is applicable where the base generative model supports conditioning controls, and, besides the text-conditioning, additional control inputs are available for the segment. In this method, we do away with the need for a scaffolding image. Instead of a scaffolding image, an artist provides a scaffolding control input for the segment. The control conditioning input can be a line art, an open pose model, a scribble,  a canny image, or any other supported control input that can guide image generation in a generative diffusion process.

Algorithmically, we proceed as follows: (i) We use a control input specifically tailored to the segment's dimensions, or we apply the segment mask $m$ to the control condition input $c^i$ to restrict the control condition only to the segment where we want generation, (ii) The image latent  $x_t$ is directly denoised through a suitable control-denoiser along with conditioning inputs of natural text and control inputs for the particular segment. We then repeat the process for all segments and for all the timesteps till $t=\kappa$. Refer to Algo.\ref{alg:scaff}, 12-17, and Fig. \ref{fig:scaff-all-conditions}, C. 

Note that since each segment is denoised independently, the algorithm supports the use of different specialized denoisers for different segments. For example, refer to Fig. \ref{fig:teaser} where we use three distinct control inputs, viz., scribble, lineart, and openpose. Combining control conditions into Composite Diffusion enables capabilities more powerful than both - the text-to-image diffusion models \cite{stable_diffusion} and the control-conditioned models\cite{zhang2023adding}. Fig. \ref{fig:control-seg-examples} refers to two example cases where we accomplish image generation tasks that are not feasible through either of these two models. 

At the end of the scaffolding stage, we construct an intermediate composite image by composing from the segment-specific latents. For each segment specific latent, we retain the region corresponding to the segment masks and discard the complementary region (Refer to Fig. \ref{fig:scaff-all-conditions} and Algo. \ref{alg:scaff}, 20-21).  The essence of the scaffolding stage is that \textit {each segment develops independently and has no influence on the development of the other segments}. We next proceed to the  `harmonization' stage,  where the intermediate composite serves as the starting point for further diffusion steps.

\begin{figure}[ht!]
\label{fig:serial}
    \centering
    \fbox{
     \includegraphics[width= .95\columnwidth]{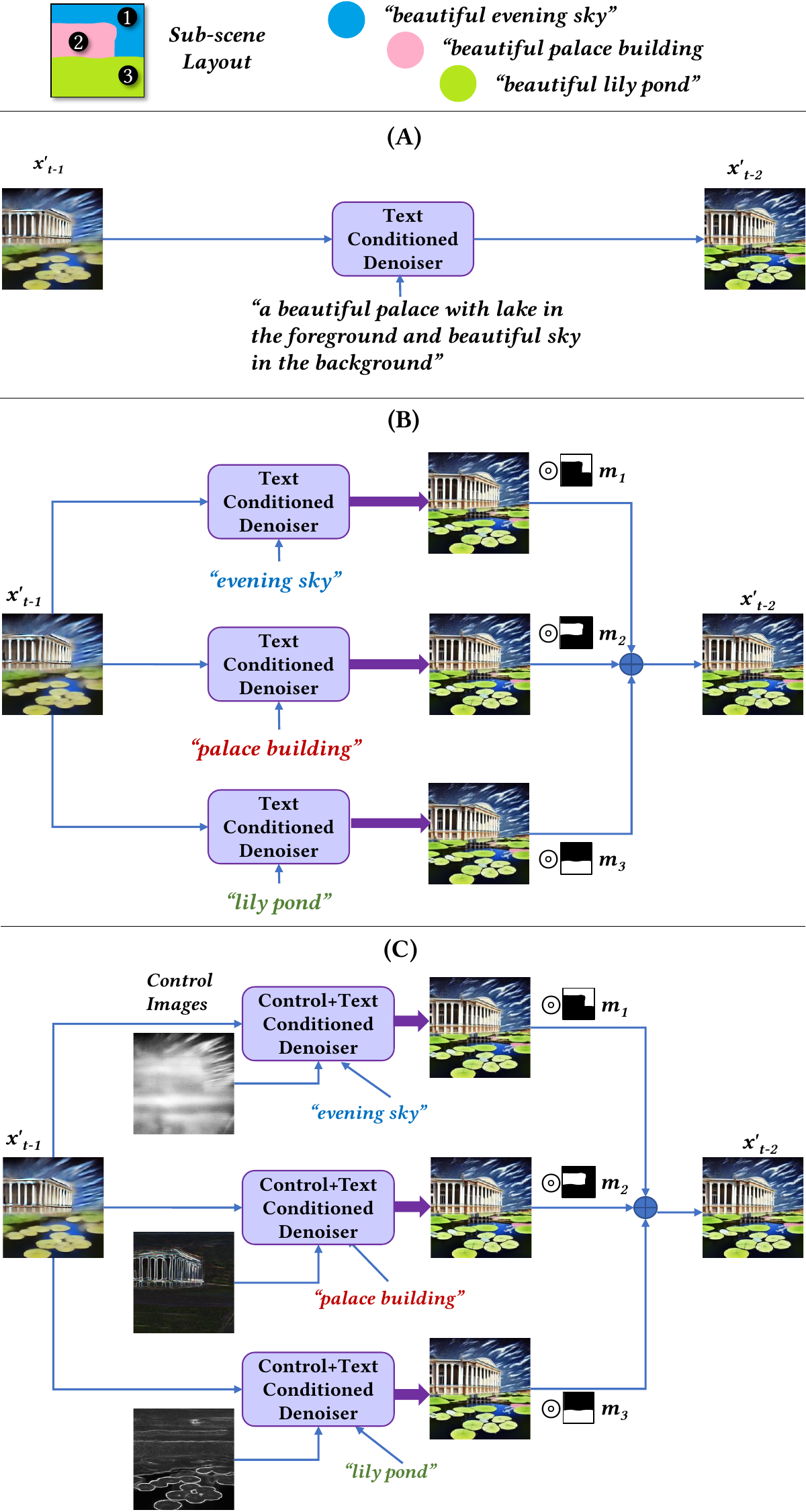}}
    \caption {Harmonization stage step for three different cases: (A) a single global text description, (B) sub-scene specific text description, and (C) sub-scene specific text description and control condition. Please note that for all the cases, the harmonization stage starts with the output of the scaffolding stage composite latent. For case (A), there is no composition step, while for cases (B) and (C), the composition step follows the denoising steps for every timestep.}
    \label{fig:harmony-all-conditions}
\end{figure}

\begin{algorithm}[ht!]
\caption{Composite Diffusion: Harmonization Stage. Input same as Algo. \ref{alg:scaff}, plus  $x^{comp}_{\kappa-1}$}
\label{alg:harmon}

\For {\textbf{all} $t$ \textbf{from} $\kappa - 1$ to $0$} 
{ 
    \uIf{Global Text Conditioning}
        {
            $x_{t-1} \leftarrow Denoise(x_t, D)$ \tcp*{Base Denoiser}              
        }
    
    \uElseIf {Segment Text Conditioning}
        {
            \For{\textbf{all segments} $i$  \textbf{from} $1$ to $n$}              
            {
                $x_{t-1}^{seg_i} \leftarrow Denoise(x_t, d^i)$ \tcp*{Base Denoiser}
                % Alternatively:
                % $z_{t-1}^{seg_i} \leftarrow Denoise(x_t, d^i, c^i)$  
            }
             $x^{comp}_{t-1} \leftarrow \sum_{i=1}^{n} x_{t-1}^{seg_i} \odot m^i$ \tcp*{Merge segments}
            
        }
  
    \uElseIf{Segment Control+Text Conditioning}
        {
            \For{\textbf{all segments} $i$  \textbf{from} $1$ to $n$}              
                {
                    $x_{t-1}^{seg_i} \leftarrow Denoise(x_t, d^i, c^i )$ \tcp*{Controlled Denoiser}
                
                }
             $x^{comp}_{t-1} \leftarrow \sum_{i=1}^{n} x_{t-1}^{seg_i} \odot m^i$ \tcp*{Merge segments}
        }
 
}
\Return  $x^{comp} \leftarrow  (x^{comp}_{-1})$ \tcp*{Final Composite}
\end{algorithm}

\subsection{Harmonizing stage}
The above method, if applied to all diffusion steps, can produce good composite images. However, because the segments are being constructed independently, the composite tends to be less harmonized and less well-blended at the segment edges. To alleviate this problem, we introduce a new succeeding stage called the \textit{`harmonization stage'}. The essential difference from the preceding scaffolding stage is that in this stage \textit{each segment develops in the context of the other segments}. We also drop any help through scaffolding images in this stage.

We can further develop the intermediate composite from the previous stage in the following ways: (i) by direct denoising the composite image latent via a global prompt (Algo. \ref{alg:harmon}, 2-3, and Fig. \ref{fig:harmony-all-conditions}, A), or (ii) by denoising the intermediate composite latent separately with each segment specific conditioning and then composing the denoised segment-specific latents. The segment-specific conditions can be either pure natural text descriptions or may include additional control conditions (Refer to Algo. \ref{alg:harmon}, 4-8 and 9-13, and Fig. \ref{fig:harmony-all-conditions}, B and C). 

\begin{figure*}[htp!]
\centering
\includegraphics[width= .98\textwidth]{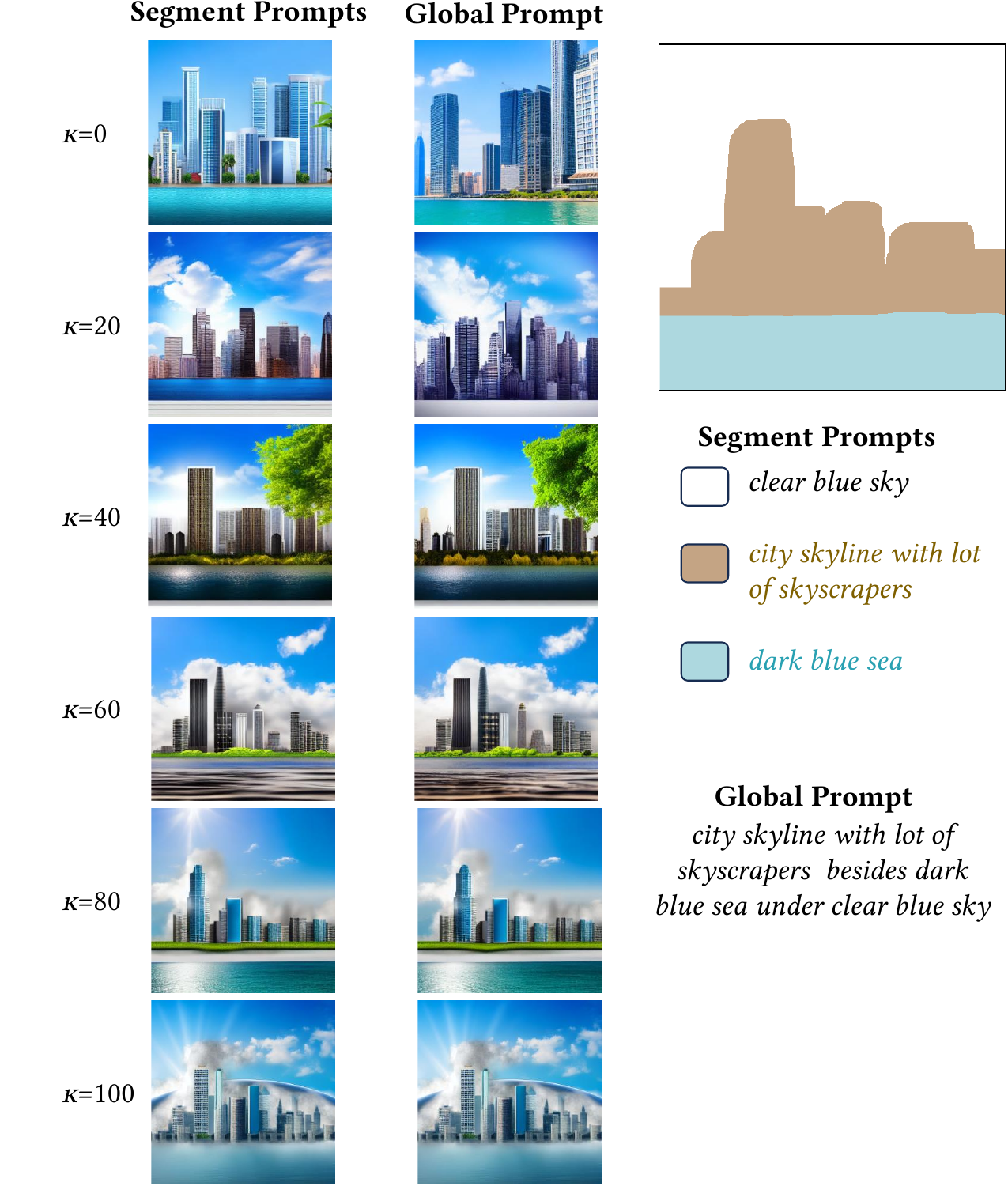}
\caption{A visual comparison of the generations using  \textit{segment-specific prompts} and \textit{global prompts} for the Harmonization stage. \textit{Harmony:} Our results show that both achieve comparable harmony with global prompts having a slight edge. \textit{Detailing:} For detailing within a segment, the segment-specific prompts provide a slight edge. Since both these methods apply only to the harmonization stage, for lower scaffolding values (e.g. $\kappa =0,  20$), the outputs vary noticeably, while at the higher values, since the number of steps for diffusion is reduced, the outputs are very close to each other.}
\label{fig:global_prompt_comparison}
\end{figure*}

While using global prompts, the output of each diffusion step is a single latent and we do not need any compositional step. For harmonization using segment-specific conditions, the compositional step of merging different segment latents at every time step (Algo. \ref{alg:harmon}, 8 and 13) ensures that the context of all the segments is available for the next diffusion step. This leads to better blending and harmony among segments after each denoising iteration. Our observation is that both these methods lead to a natural coherence and convergence among the segments of the composite image (Fig. \ref{fig:global_prompt_comparison} provides an example illustration).

\begin{figure}[t!]
    \centering
    \includegraphics[width= .92\columnwidth]{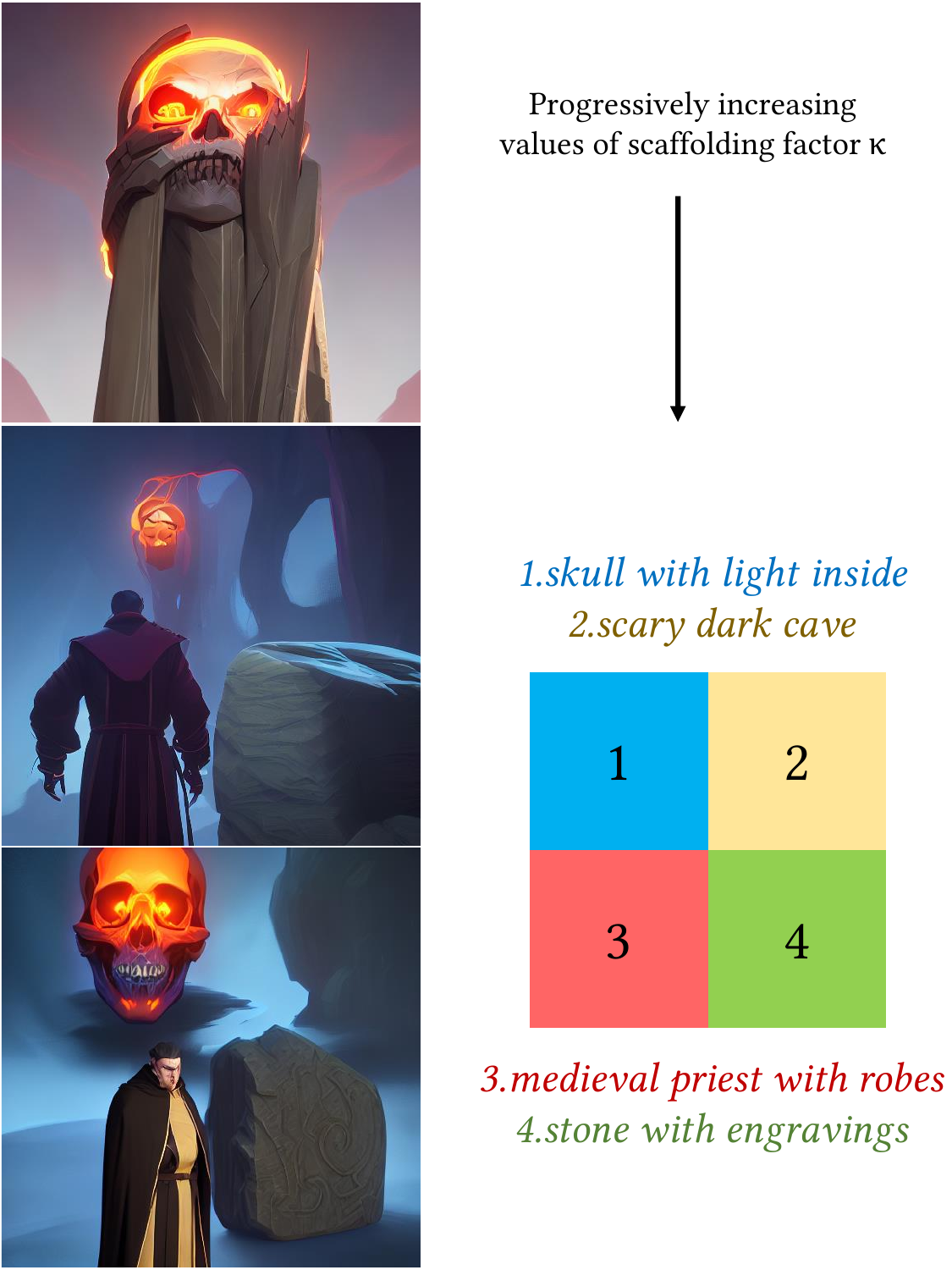}
    \caption {Effect of scaffolding factor on \textit{Artworks}. For the given inputs and generations from top to bottom:  At the lower extreme, $\kappa=0$, we get an image that merges the concepts of text descriptions for different segments. At the higher end, $\kappa=80$, we get a collage-like effect. In the middle, $\kappa=40$,  we hit a sweet spot for a well-blended image suitable for a story illustration.}
    \label{fig:kappa}
\end{figure}

\subsection{Scaffolding factor $\kappa$: }
\label{sec:scaffolding_factor}
We define a parameter called the scaffolding factor, denoted by $\kappa$ (kappa), whose value determines the percentage of the diffusion process that we assign to the scaffolding stage.  
$\kappa = \frac{\text{number of scaffolding steps}} {\text{total diffusion steps}} \times 100$.
The number of harmonization steps is calculated as total diffusion steps minus the scaffolding steps. If we increase the $\kappa$ value, we allow the segments to develop independently longer. This gives better conformance with the segment boundaries while reducing the blending and harmony of the composite image. If we decrease the $\kappa$ value, the individual segments may show a weaker anchoring of the image and lesser conformance to the mask boundaries. However, we see increased harmony and blending among the segments.

Our experience has shown that the appropriate value of $\kappa$ depends upon the domain and the creative needs of an artist. Typically, we find that values of kappa around 20-50 are sufficient to anchor an image in the segments. 
Figure \ref{fig:kappa} illustrates the impact of $\kappa$ on image generation that gives artists an interesting creative control on segment blending. Appendix Table \ref{tab:kappa_ablation} provides a quantitative evaluation of the impact of the scaffolding factor on the various parameters of image quality.

%% file: 04_quality_evaluation.tex
\section{Quality criteria and evaluation}
\label{sec:quality_criteria}
As stated earlier, one of the objectives of this research is to ask the question:  Is the quality of the composite greater than or equal to the sum total of the quality of the individual segments? In other words, the individual segments in the composite should not appear unconnected but should work together as a whole in meeting the artist's intent and quality goals. 

In this section, we lay out the quality criteria, and evaluation approach and discuss the results of our implementations.

\subsection{Quality criteria}
We find that the present methods of evaluating the image quality of a generated image are not sufficient for our purposes. For example, methods such as FID, Inception Score, Precision, and Recall \cite{FID_NIPS, IS_NIPS, sajjadi2018assessing, borji2022pros} are traditionally used for measuring the quality and diversity of generated images, but only with respect to a large set of reference images. Further, they do not evaluate some key properties of concern to us such as conformity of the generated images to the provided inputs, the harmonization achieved when forming images from sub-scenes, and the overall aesthetic and technical quality of the generated images. These properties are key to holistically evaluating the Composite Diffusion approach. To this end, we propose the following set of quality criteria:

\textbf{1. CF:} \textit{Content Fidelity:} 
 The purpose of the text prompts is to provide a natural language description of what needs to be generated in a particular region of the image. The purpose of the control conditions is to specify objects or visual elements within a sub-scene. This parameter measures how well the generated image represents the textual prompts (or control conditions)  used to describe the sub-scene.

\textbf{2. SF:} \textit{Spatial Layout Fidelity:} The purpose of the spatial layout is to provide spatial location guidance to various elements of the image. This parameter measures how well the parts of the generated image conform to the boundaries of specified segments or sub-scenes. 

\textbf{3. BH:} \textit{Blending and Harmony:}
When we compose an image out of its parts, it is important that the different regions blend together well and we do not get abrupt transitions between any two regions.
Also, it is important that the image as a whole appears harmonious, i.e., the contents, textures, colors, etc. of different regions form a unified whole. This parameter measures the smoothness of the transitions between the boundaries of the segments, and the harmony among different segments of the image. 

\textbf{4. QT:} \textit{Technical Quality:}
The presence of noise and unwanted artifacts that can appear in the image generations can be distracting and may reduce the visual quality of the generated image. This parameter measures how clean the image is from the unwanted noise, color degradation, and other unpleasant artifacts like lines, patches, and ghosting appearing on the mask boundaries or other regions of the image.

\textbf{5. QA:} \textit{Aesthetics Quality:} 
 Aesthetics refers to the visual appeal of an image. Though subjective in nature, this property plays a great part in the acceptability or consumption of the image by the viewers or the users.  This parameter measures the visual appeal of the generated image to the viewer. 

\subsection{Evaluation approach}
\label{sec:evaluation}
   In this section, we provide details about our evaluation approach. We first provide information on the baselines used for comparison and then information on the methods used for evaluation such as user studies, automated evaluations, and artist's consultation and feedback. 
   
\begin{figure}[t!]
    \centering
    \includegraphics[width= \linewidth]{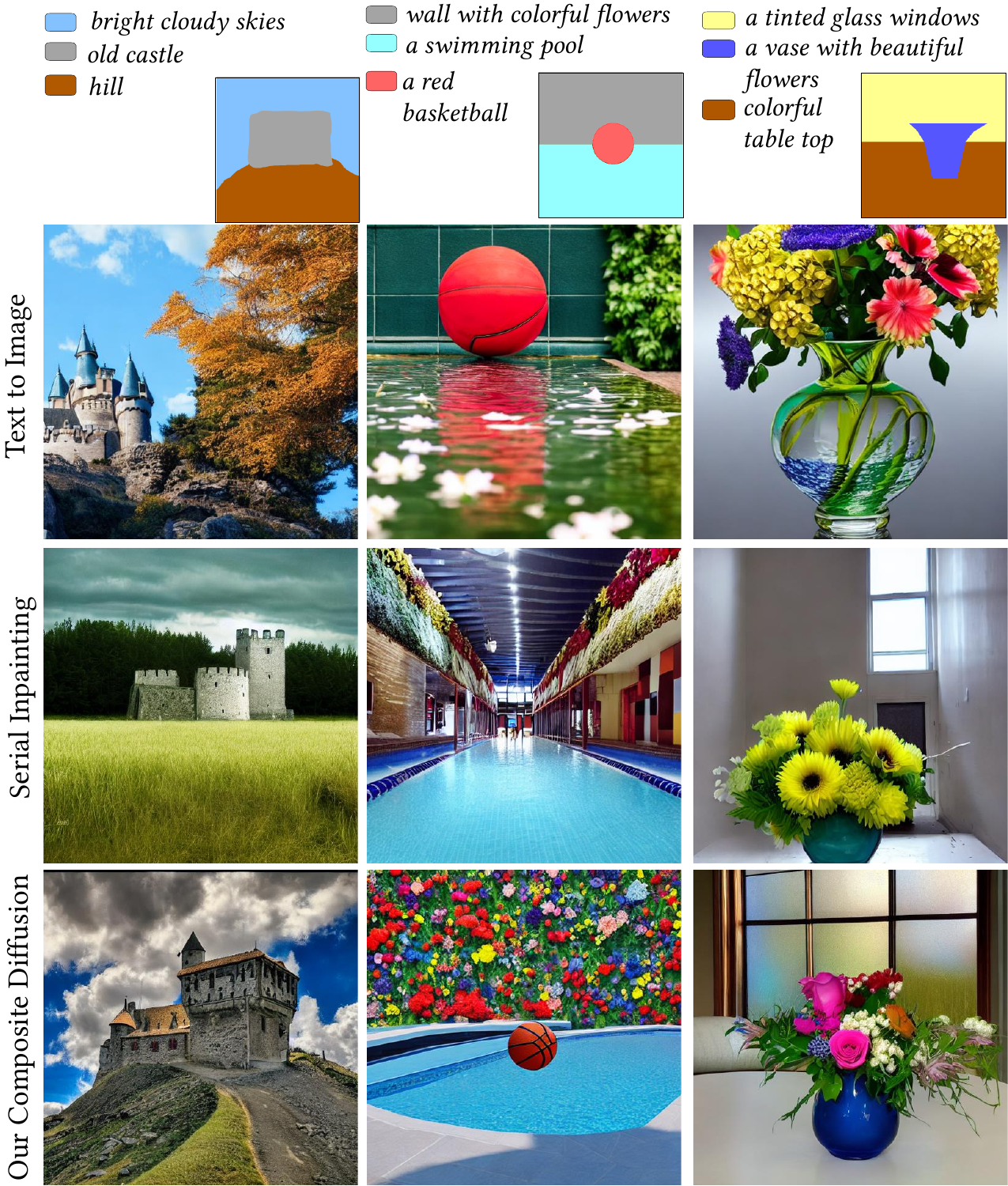}
    
    \caption{A comparison of composite images generated through text-2-image, serial-inpainting, and Composite Diffusion methods.}
    \label{fig:image-compare}
\end{figure}

We deploy the following two baselines for comparison with our methods:
\begin{itemize}
\item \textbf{Baseline 1 (B1)} - \textit{Text to Image:}  This is the base diffusion model that takes only text prompts as the input. Since this input is unimodal, the spatial information is provided solely through natural language descriptions. 
    
\item \textbf{Baseline 2 (B2)} - \textit{Serial Inpainting:}  As indicated in the section \ref{sec:rw_inpainting}, we should be able to achieve a composite generation by serially applying inpainting to an appropriate background image and generating one segment at a time.

\end{itemize}
A sample of images from different algorithms is shown in Figure \ref{fig:image-compare}. We have implemented our algorithms using Stable Diffusion 1.5 \cite{stable_diffusion}as our base diffusion model, and Controlnets 1.1 \cite{zhang2023adding} as our base for implementing controls. The implementation details for our algorithms and two baselines are available in Appendix \ref{sec:experimental_setup}, \ref{sec:serial_inpainting}, \&  \ref{sec:our_method_implementation}.

We measure the performance of our approach against the two baselines using the above-mentioned quality criteria. Specifically, we perform four different kinds of evaluations:

\textbf {(i)} \textit{Human evaluations:}
We created a survey where users were shown the input segment layout and textual descriptions and the corresponding generated image. The users were then asked to rate the image on a scale of 1 to 5 for the five different quality criteria. We utilized social outreach and Amazon MTurk to conduct the surveys and used two different sets of participants: (i) a set of General Population (GP) comprised of people from diverse backgrounds, and (ii) a set of Artists and Designers (AD) comprised of people with specific background and skills in art and design field. 

We found the current methods of automated metrics  \cite{FID_NIPS, IS_NIPS, sajjadi2018assessing, borji2022pros} inadequate for evaluating the particular quality requirements of Composite Diffusion. Hence, we consider and improvise a few automated methods that can give us the closest measure of these qualities. We adopt CLIP-based similarity \cite{clip} to measure content(text) fidelity and spatial layout fidelity. We use Gaussian noise as an indicator of technical degradation in generation and estimate it \cite{chen2015efficient} to measure the technical quality of the generated image. For aesthetic quality evaluation,  we use a CLIP-based aesthetic scoring model \cite{laion_aesthetic} that was trained on -  a dataset of 4000 AI-generated images and their corresponding human-annotated aesthetic scores. ImageReward \cite{xu2023imagereward} is a text-image human preference reward model trained on human preference ranking of over 100,000 images; we utilize this model  to estimate human preference for a comparison set of generated images. 

Additionally, we also do \textbf{(iii)} a qualitative visual comparison of images (e.g., Figures \ref{fig:relevant_comparison},  and  \ref{fig:image-compare}), and \textbf{(iv)} an informal validation by consulting with an artist. We refer readers to Appendix \ref{sec:survey_results}, \ref{sec:automated_evaluation_methods_details}, and \ref{appendix_sec:artworks} for more details on the human and automated evaluation methods. 

\begin{figure}[t!]
    \centering
    \includegraphics[width =  \linewidth]{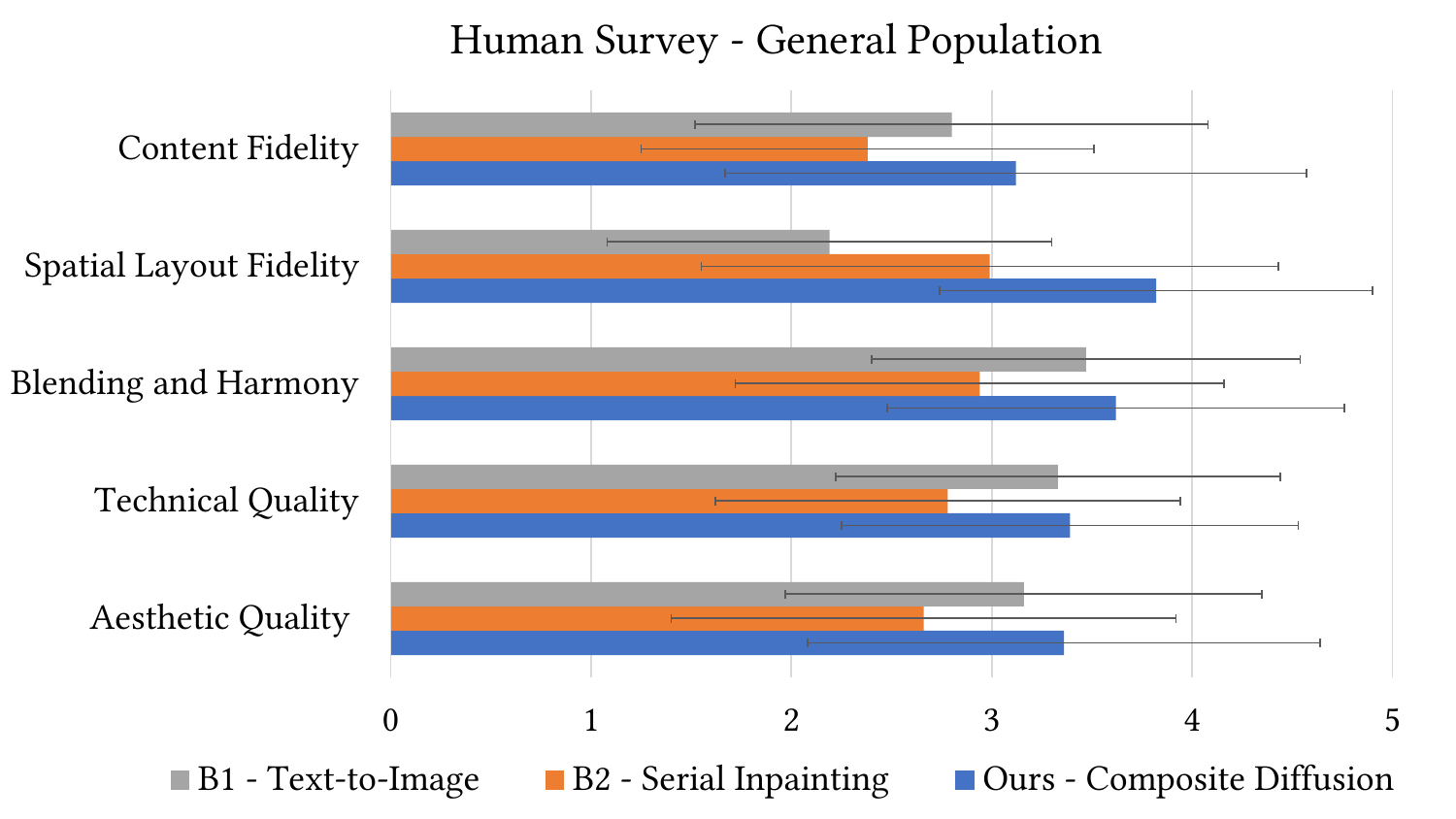}  % \includegraphics[width =  \textwidth]{figs/Evaluation Figure - Final.pdf}
    \caption {Human evaluation results from the set - General Population(GP)}
    \label{fig:evaluation-gp}
\end{figure}

\begin{figure}[t!]
    \centering
    \includegraphics[width =  \linewidth]{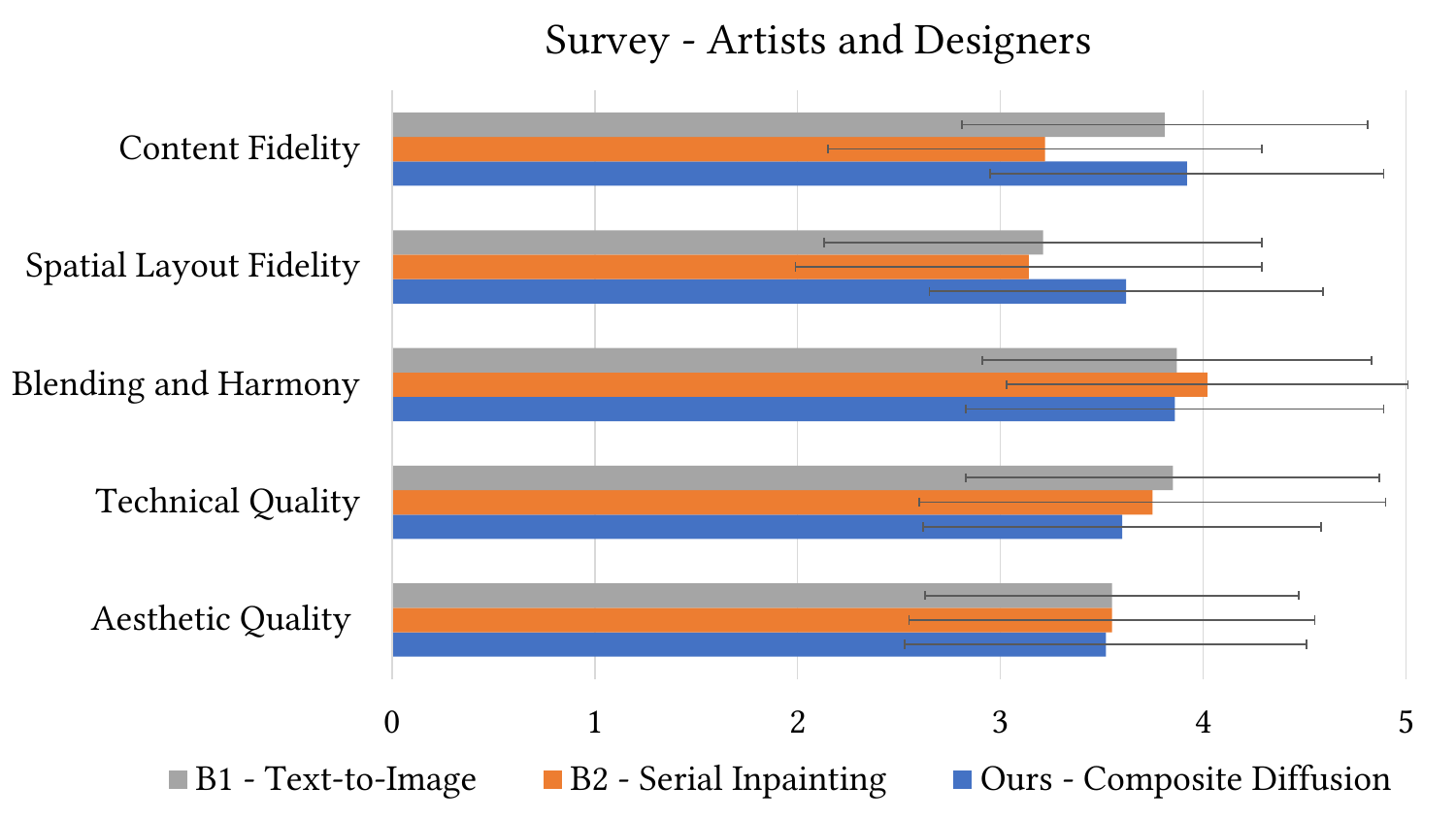}
    \caption {Human evaluation results  from the set - Artists/Designers(AD) }
    \label{fig:evaluation-ad}
\end{figure}

\subsection{Results and discussion}
\label{sec:results}
In this section, we summarize the results from the different types of evaluations and provide our analysis for each quality criterion.

\subsubsection{Content Fidelity} In both types of human evaluations, GP and AD, Composite Diffusion(CD) scores are higher than the two baselines. Composite Diffusion also gets a higher score for content fidelity on automated evaluation methods. 

\textbf{Our take:} This can be attributed to the rich textual descriptions used for describing each image segment, resulting in an overall increase in semantic information and control in the generation process. One can argue that similar rich textual descriptions are also available for the serial inpainting method (B2). However, B2 might get several limitations: (i) There is a dependency on the initial background image that massively influences the inpainting process, (ii) There is a sequential generation of the segments, which would mean that the segments that are generated earlier are not aware of the full context of the image. (iii) The content in textual prompts may sometimes be missed as the the prompts for  inpainting apply to the whole scene than a sub-scene generation.

\begin{table}[t!]
\centering
\caption{Automated evaluation results. The best performing algorithm in a category is marked in bold}
\label{tab:automated_result}
\begin{small}
\begin{tabular}{|c|c|c|c|}
\hline
                          & \textbf{B1} & \textbf{B2} & \textbf{Ours} \\ \hline
Content Fidelity $\uparrow$    & 0.2301          & 0.2485          & \textbf{0.2554}          \\ \hline
Spatial Layout Fidelity $\uparrow$ & 0.2395          & 0.2632          & \textbf{0.2735}          \\ \hline
Blending \& Harmony $\downarrow$         & 6903            & \textbf{725}             & {7404}          \\ \hline
Technical Quality $\downarrow$     & 1.34            & 2.6859          & \textbf{1.2438}        \\ \hline
Aesthetic Quality $\uparrow$   & 6.3448          & 5.5069          & \textbf{6.3492}          \\ \hline
Human Preference $\downarrow$ & 3               & 2               & \textbf{1}               \\ \hline
\end{tabular}
\end{small}
\end{table}

\subsubsection{Spatial Fidelity} This is a key parameter for our evaluation. All three types of evaluation methods - Human evaluation GP and AD, and automated methods - reveal a superior performance of Composite Diffusion.  

\textbf{Our take:}This is on expected lines. Text-to-Image (B1) provides no explicit control over the spatial layout apart from using natural language to describe the relative position of objects in a scene.  B2 could have spatial fidelity better than B1 and equivalent to Composite Diffusion. It does show an improvement over B1 in human evaluation-AD and automated methods. Its lower spacial conformance compared to Composite Diffusion can be attributed to the same reasons that we discussed in the previous section.

\subsubsection{Blending and Harmony} Human-GP evaluation rates our method as the best, while Human-AD evaluation and automated methods give an edge to the serial inpainting method. 

\textbf{Our take:} Text-to-Image (B1) generates one holistic image, and we expect it to produce a well-harmonized image. This higher rating for the serial-inpainting method could be due to the particular implementation of inpainting that we use in our project. This inpainting implementation (RunwayML SD 1.5 \cite{stable_diffusion})is especially fine-tuned to provide seamless filling of a masked region by direct inference similar to text-to-image generation. Further, in Composite Diffusion, the blending and harmonization are affected by the chosen scaffolding value, as shown in Appendix table  \ref{tab:kappa_ablation}.

\subsubsection{Technical Quality} Human evaluation-GP gives our method a better score, while Human evaluation-AP gives a slight edge to the other methods. The automated evaluation method considers only one aspect of technical quality, viz., the presence of noise; our algorithm shows lesser noise artifacts.

\textbf{Our Take:} Both serial-inpainting and Composite Diffusion build upon the base model B1. Any derivative approach risks losing the technical quality while attempting to introduce control. Hence, we expect the best-performing methods to maintain the technical quality displayed by B1. However, repeated application of inpainting to cover all the segments in B2 may amplify any noisy artifact introduced in the early stages. We also observed that for Composite Diffusion, if the segment masks do not have well-demarcated boundaries, we might get unwanted artifacts in the generated composites. 

\subsubsection{Aesthetical Quality} Human evaluation-GP gives Composite Diffusion method a clear edge over baseline methods, while Human evaluation-AP results show a comparable performance. The automated evaluation methods rate our method higher than the serial inpainting and only marginally higher than the text-to-image baseline. 

\textbf{Our take:} These results indicate that our approach does not cause any loss of aesthetic quality but may even enhance it.   The good performance of Composite Diffusion in aesthetic evaluation can be due to the enhanced detail and nuance with both textual and spatial controls. The lack of global context of all the segments in serial inpainting and the dependence on an appropriate background image put it at a slight disadvantage. Aesthetics is a subjective criterion that can be positively influenced by having more meaningful generations and better placements of visual elements. Hence, combining segment layouts and content conditioning in Composite Diffusion may lead to compositions with more visually pleasing signals. 

We further did a qualitative validation with an external artist. We requested the artist to specify her intent in the form of freehand drawings with labeled descriptions. We manually converted the artist's intent to bimodal input of segment layout and textual descriptions suitable for our model.   We then created artwork through Composite Diffusion and asked the artist to evaluate them qualitatively. The feedback was largely positive and encouraging. The artist's inputs, the generated artwork, and the artist's feedback are available in the Appendix section \ref{appendix_sec:artworks}. 

 We also present a qualitative visual comparison of our generated outputs with the baselines and other related approaches in Figures \ref{fig:image-compare} and \ref{fig:relevant_comparison} respectively. Summarizing the results of multiple modes of evaluation, we can affirm that our Composite Diffusion methods perform holistically and well across all the different quality criteria.

%% file: 05_conclusion.tex
\section{Conclusion}
\label{sec:conclusion}
In this paper, we introduced composite generation as a method for generating an image by composing from its constituent \textit{sub-scenes}. The method vastly enhances the capabilities of text-to-image models. It enables a new mode for the artists to create their art by specifying (i) \textit{spatial intent} using free-form sub-scene layout, and (ii)  \textit{content intent} for the sub-scenes using natural language descriptions and different forms of control conditions such as scribbles, line art, and human pose. 

To provide artists with better affordances, we propose that the spatial layout should be viewed as a coarse-grained layout for \textit{sub-scenes} rather than an outline of individual fine-grained objects. For a finer level of control within a sub-scene, it is best to apply sub-scene-specific control conditions. We strongly feel that this arrangement is intuitive for artists and easy to use for novices.  

We implemented composite generation in the context of diffusion models and called it \textit{Composite Diffusion}. We showed that the model generates quality images while adhering to the spatial and semantic constraints imposed by the input modalities of free-form segments, natural text, and other control conditions. Our methods do not require any retraining of models or change in  the core architecture of the pre-trained models. Further, they work seamlessly with any fine-tuning of the base generative model.

We recommend modularizing the process of composite diffusion into two stages: scaffolding and harmonizing. With this separation of concerns, researchers can independently develop and improve the respective stages in the future. We observe that diffusion processes are inherently harmonizing in nature and we can achieve a more natural blending and harmonization of an image by exploiting this property than through other external means. 

We also highlighted the need for better \textit{quality criteria} for generative image generations. We devised one such \textit{quality criteria} suitable for evaluating the results of Composite Diffusion in this paper.  To evaluate using these criteria, we conducted both human evaluations and automated evaluations. Although the automated evaluation methods for Composite Diffusion are limited in their scope and are in an early stage of development,  we nevertheless found an interesting positive correlation between human evaluations and automated evaluations. 

We make an essential observation about benchmarking: The strength of the base model heavily influences the quality of generated composite. Base model strength, in turn, depends upon the parameter strength, architecture, and quality of the training data of the base model. Hence, any evaluation of the quality of generated composite images should be in relation to the base model image quality. For example, in this paper, we use Stable Diffusion v1.5 as the standard base for all types of generations, viz., text-to-image, repeated inpainting, and composite diffusion generations.

Finally, we demonstrated that our approach achieves greater spatial, semantic, and creative control in comparison to the baselines and other approaches. This gives us confidence that with careful application, the holistic quality of an image generated through Composite Diffusion would indeed be greater than or equal to ($>=$) the sum of the quality of its constituent parts. 

\subsection{Future work}
We discuss some of the interesting future research problems and possibilities related to this work. 

We implemented Composite Diffusion in the context of Stable Diffusion \cite{stable_diffusion}. It would instructive to explore the application of Composite Diffusion in the context of different architectures like Dalle-E \cite{dalle2},  Imagen \cite{imagegen},  or other open sources models such as Deep Flyod \cite{deep_floyd}. Since the definition of Composite Generation (with input modality as defined in this paper) is generic, it can also be applied to other generative models, such as GANs or any future visual generative models. 

In this work, we have experimented with only two sampling methods - DDPM  \cite{ho2020denoising} and  DDIM \cite{song2020denoising}; all the generations in this paper use DDIM. It would be interesting to study the impact of different sampling methods, such as Euler, DPM, LMS, etc. \cite{samplers_sd_andrew, samplers_sd_agata},  on the Composite Diffusion. 

For evaluation purposes, we faced the challenge of a relevant dataset for Composite Diffusion. There are no ready data sets that provide \textit{ free-form sub-scene layout} along with the \textit{natural-language descriptions} of those sub-scenes. We handcrafted a 100-image data set of sub-scene layouts and associated sub-scene captions. The input dataset and the associated generated images helped us benchmark and evaluate different composite generation methods.   By doing multiple generations for each set of inputs, we can effectively enhance the size of the data set for evaluation. We strongly feel that this dataset should be augmented further for size and diversity - through community help or automatic means. A larger data set, curated on the above lines, will be extremely useful for benchmarking and future work.

%% file: _appendix.tex
\appendix
\label{sec:appendix}
\input{supp_00_supp_org}
\input{supp_01_background.tex}
\clearpage
\input{supp_02_base_setup}

\input{supp_02b_serial_inpainting}
\clearpage
\input{supp_03_our_method_details}

\input{supp_03b_features_limitations_societal}
\clearpage
\input{supp_04_evaluation_details}

\input{supp_04b_artworks}

%% file: supp_00_supp_org.tex
\section{Appendix organization}
\label{sec:appendix_org}

In this appendix, we provide the supplemental material to the paper: Composite Diffusion:  $whole >= \Sigma parts$. It is organized into the following four main parts:

\begin{description}
    \item [1. Background for methods]
    Appendix-\ref{sec:background} provides the mathematical background for image generation using diffusion models relevant to this paper.

    \item [2. Our base setup and serial inpainting method] 
    Appendix-\ref{sec:experimental_setup} provides the details of our experimental setup, the features and details of the base implementation model, and text-to-image generation through the base model which also serves as our baseline 1. Appendix-\ref{sec:serial_inpainting} provides the details of our implementation of the serial inpainting method which also serves as our baseline 2.

    \item [3. Our method: details and features] Appendix-\ref{sec:our_method_implementation} covers the additional implementation details of our Composite Diffusion method discussed in the main paper. Appendix-\ref{sec:personalizing_at_scale} discusses the implication of Composite Diffusion in personalizing content generation at a scale. Appendix-\ref{sec:limitations} discusses some of the limitations of our approach and Appendix-\ref{sec:societal_impact} discusses the possible societal impact of our work.

    \item [4. Evaluation details] Appendix-\ref{sec:survey_results} provides the additional details of the surveys in the human evaluation, Appendix-\ref{sec:automated_evaluation_methods_details} of the automated methods for evaluation, and Appendix-\ref{appendix_sec:artworks} of the validation exercise with an external artist.
    
\end{description}

%% file: supp_01_background.tex
\section{Background for methods}
\label{sec:background}
In this section, we provide an overview of diffusion-based generative models and diffusion guidance mechanisms that serve as the foundational blocks of the methods in this paper. The reader is referred to \cite{dieleman2022guidance, ho2020denoising, weng2021diffusion} for any further details and mathematical derivations. 

\subsection{Diffusion models(DM)} In the context of image generation, DMs are a type of generative model with two diffusion processes: (i) a \textit{ forward diffusion process}, where we define a Markov chain by gradually adding a small amount of random noise to the image at each time step, and (ii)a \textit{reverse diffusion process}, where the model learns to generate the desired image, starting from a random noise sample. 

\subsubsection{Forward diffusion process}
\label{sec:forward_diff}
Given a real distribution $q(\mathbf{x})$, we sample an image $\mathbf{x}_0$ from it  ($\mathbf{x}_0 \sim q(\mathbf{x})$). We gradually add Gaussian noise to it with a variance schedule $\{\beta_t \in (0, 1)\}_{t=1}^T$ over $T$ steps to get progressively noisier versions of the image $\mathbf{x}_1, \dots, \mathbf{x}_T$. The conditional distribution at each time step $t$ with respect to its previous timestep $t-1$ is given by the diffusion kernel:

\begin{equation}
q(\mathbf{x}_{1:T} ) = q(\mathbf{x}_0)\prod^T_{t=1} q(\mathbf{x}_t \vert \mathbf{x}_{t-1}) 
\end{equation}

The features in $\mathbf{x}_0$ are gradually lost as step $t$ becomes larger. When $T$ is sufficiently large, $T \to \infty$, then $\mathbf{x}_T$ approximates an isotropic Gaussian distribution. 

\textbf{Q-sampling:} An interesting property of the forward diffusion process is that we can also sample $\mathbf{x}_t$  directly from $\mathbf{x}_0$ in the closed form. If we let  $\alpha_t =  1 - \beta_t$, $\bar{\alpha_t} = \prod^t_{s=1} \alpha_s$, we get:

\begin{equation}
q(\mathbf{x}_t \vert \mathbf{x}_0) = \mathcal{N}(\mathbf{x}_t; \sqrt{\bar{\alpha}_t} \mathbf{x}_0, (1 - \bar{\alpha}_t)\mathbf{I}) 
\end{equation}

Further, for $\bm{\epsilon} \sim \mathcal{N} (\mathbf{0}, \mathbf{I})$, $\mathbf{x}_t$ can be expressed as a linear combination of $\mathbf{x}_0$ and $\bm{\epsilon}$:

\begin{equation}
\mathbf{x}_t = \sqrt{\bar{\alpha}_t}\mathbf{x}_0 + \sqrt{1 - \bar{\alpha}_t}{\bm{\epsilon}}
\end{equation}

We utilize this property in many of our algorithms and refer to it as: \textit{`q-sampling'}.

\subsubsection{Reverse diffusion process}
 Here we reverse the Markovian process and, instead, we sample from $q(\mathbf{x}_{t-1} \vert \mathbf{x}_t)$.  By repeating this process, we should be able to recreate the true sample (image), starting from the pure noise $\mathbf{x}_T \sim \mathcal{N}(\mathbf{0}, \mathbf{I})$. If $\beta_t$ is sufficiently small, $q(\mathbf{x}_{t-1} \vert \mathbf{x}_t)$ too will be an isotropic Gaussian distribution. However, it is not straightforward to estimate $q(\mathbf{x}_{t-1} \vert \mathbf{x}_t)$ in closed form. We, therefore, train a model $p_\theta$ to approximate the conditional probabilities that are required to run the reverse diffusion process. 

\begin{equation}
    p_\theta(\mathbf{x}_{t-1} \vert \mathbf{x}_t) = \mathcal{N}(\mathbf{x}_{t-1}; \bm{\mu}_\theta(\mathbf{x}_t, t), \bm{\Sigma}_\theta(\mathbf{x}_t, t))
\end{equation}
where $\bm{\mu}_\theta$ and $\bm{\Sigma}_\theta$ are the predicted mean and variance of the conditional Gaussian distribution. In the earlier  implementations $\bm{\Sigma}_\theta (x_t, t)$ was kept constant \cite{ho2020denoising}, but later it was shown that it is preferable to learn it through a neural network that interpolates between the upper and lower bounds for the fixed covariance \cite{diffusion_models_dhariwal}.

The reverse distribution is:
\begin{equation}
    p_\theta(\mathbf{x}_{0:T}) = p(\mathbf{x}_T) \prod^T_{t=1} p_\theta(\mathbf{x}_{t-1} \vert \mathbf{x}_t)
\end{equation}
Instead of directly inferring the image through $\bm{\mu}_\theta(x_t, t)$), it might be more convenient to predict the noise ($\bm{\epsilon}_\theta(x_t, t)$) added to the initial noisy sample ( $\mathbf{x}_t$)  to obtain  the denoised sample ($\mathbf{x}_{t-1}$) \cite{ho2020denoising}. Then, $\bm{\mu}_\theta(\mathbf{x}_t, t)$ can be derived as follows:

\begin{equation}
\bm{\mu}_\theta(x_t, t)
=  \frac{1}{\sqrt{\alpha_t}} \Big( \mathbf{x}_t - \frac{\beta_t}{\sqrt{1 - \bar{\alpha}_t}} \bm{\epsilon}_\theta(\mathbf{x}_t, t) \Big)
\end{equation}

\textbf{Sampling:} Mostly, a U-Net neural architecture \cite{unet} is used to predict the denoising amount at each step. A scheduler samples the output from this model. Together with the knowledge of time step $t$, and the input noisy sample $\mathbf{x}_t$, it generates a denoised sample $\mathbf{x}_t$. For sampling through Denoising Diffusion Probabilistic Model (DDPM) \cite{ho2020denoising}, denoised sample is obtained through the following computation:

\begin{equation}
\mathbf{x}_{t-1} =  \frac{1}{\sqrt{\alpha_t}} \Big( \mathbf{x}_t - \frac{\beta_t}{\sqrt{1 - \bar{\alpha}_t}} \bm{\epsilon}_\theta(\mathbf{x}_t, t) \Big) + \sigma_t \bm{\epsilon}
\end{equation}

where $\bm{\Sigma}_\theta(\mathbf{x}_t, t) = \sigma^2_{t} \mathbf{I}$ , and  $\bm{\epsilon} \sim \mathcal{N} (\mathbf{0}, \mathbf{I}) $ is a random sample from the standard Gaussian distribution. 

To achieve optimal results for image quality and speed-ups, besides DDPM, various sampling methods, such as DDIM, LDMS, PNDM, and LMSD \cite{samplers_sd_andrew, samplers_sd_agata} can be employed.

We use DDIM (Denoising Diffusion Implicit Models) as the common method of sampling for all the algorithms discussed in this paper. Using DDIM, we sample $\mathbf{x}_{t-1}$ from $\mathbf{x}_t$ and $\mathbf{x}_0$  via the following equation \cite{song2020denoising}:

\begin{equation}
\label{eq:ddim_sample}
    \mathbf{x}_{t-1} = \sqrt{\bar{\alpha}_{t-1}} \mathbf{x}_0 + \sqrt{1-\bar{\alpha}_{t-1} -\sigma^2_t}  \bm{\epsilon}_\theta(\mathbf{x}_t, t) + \sigma_t \bm{\epsilon}
\end{equation}

Using DDIM sampling, we can produce samples that are comparable to DDPM samples in image quality, while using only a small subset of DDPM timesteps (e.g., $50$ as opposed to $1000$).

\subsubsection{Latent diffusion models(LDM)} 
We can further increase the efficiency of the generative process by running the diffusion process in latent space that is lower-dimensional than but perceptually equivalent to pixel space. Performing diffusion in lower dimensional space provides massive advantages in terms of reduced computational complexity. For this, we first downsample the images into a lower-dimensional latent space and then upsample the results from the diffusion process into the pixel space. For example, the latent diffusion model described in  \cite{stable_diffusion} uses a suitably trained variational autoencoder to encode an RGB pixel-space image ($\mathbf{x} \in \mathbb{R}^{H \times W \times 3}$) into a latent-space representation ($\mathbf{\mathbf{z}} = \mathcal{E}(\mathbf{x})$, $\mathbf{z}\in \mathbb{R}^{h\times w \times c}$  ), where $f=H/h=W/w$ describes the downsampling factor. The diffusion model in the latent space operates similarly to the pixel-space diffusion model described in the previous sections, except that it utilizes a latent space time-conditioned U-Net architecture. The output of the diffusion process ($\tilde {\mathbf{z}}$) is decoded back to the pixel-space ($\tilde {\mathbf{x}} = \mathcal{D}(\tilde{\mathbf{z}})$). 

\subsection{Diffusion guidance}
An unconditional diffusion model, with mean $\mu_{\theta}(x_t)$ and variance $\Sigma_\theta(x_t)$  usually predicts a score function $\nabla_{x_t} \mathrm{log}\, p(x_t)$ which additively perturbs it and pushes it in the direction of the gradient. In conditional models, we try to model conditional distribution $\nabla_{x_t}\mathrm{log}\, p(x_t|y)$, where $y$ can be any conditional input such as class label and free-text. This term, however, can be derived to be a combination of unconditional and conditional terms\cite{dieleman2022guidance}: 
$$\nabla_{x_t}\mathrm{log}\, p(x_t|y) = \nabla_{x_t}\mathrm{log}\, p(x_t) + \nabla_{x_t} \mathrm{log}\, p(y|x_t) $$

\subsubsection{Classifier driven guidance}
We can obtain $\mathrm{log}\, p(y|x_t)$ from an external classifier that can predict a target $y$ from a high-dimension input like an image $x$. A guidance scale $s$ can further amplify the conditioning guidance.
$$\nabla_{x_t}\mathrm{log}\, p_s(x_t|y) = \nabla_{x_t}\mathrm{log}\, p(x_t) + s.\nabla_{x_t} \mathrm{log}\, p(y|x_t) $$
$s$ affects the quality and diversity of samples.
\subsubsection{CLIP driven guidance}
Contrastive Language–Image Pre-training (CLIP) is a neural network that can learn visual concepts from natural language supervision \cite{clip}.  %CLIP model consists of a text and image encoder which are pre-trained together for the task of image-text matching. 
The pre-trained encoders from the CLIP model can be used to obtain semantic image and text embeddings which can be used to score how closely an image and a text prompt are semantically related. 

Similar to a classifier, we can use the gradient of the dot product of the image and caption encodings ( $f(x_t)$  and $g(c)$) with respect to the image to guide the diffusion process  \cite{clipglass, glide, style_clip}.

$$\hat{\mu}_\theta(x_t|c) = \mu_\theta(x_t|c) + s \cdot \Sigma_\theta(x_t|c)\nabla_{x_t} (f(x_t) \cdot g(c))$$

To perform a simple classifier-guided diffusion, Dhariwal and Nichol\cite{diffusion_models_dhariwal} use a classifier that is pre-trained on noisy images to guide the image generation. However, training a CLIP model from scratch on noisy images may not be always feasible or practical. To mitigate this problem we can estimate a clean image $ \hat{x_0}$ from a noisy latent $x_t$ by using the following equation.  

\begin{equation}
\hat{x_0} =  \frac{{x}_t}{\sqrt{\bar{\alpha}_t}} - \frac{\sqrt{1 - \bar{\alpha}_t}\epsilon_\theta(x_t, t)}{\sqrt{ \bar{\alpha}_t}}    
\end{equation}

We can then use this projected clean image $\hat{x}_0$ at each state of diffusion step $t$ for comparing with the target text.
Now, a CLIP-based loss $L_{CLIP}$ may be defined as the cosine distance (or some similar distance measure) between the CLIP embedding of the text prompt ($d$) and the embedding of the estimated clean image $\hat{x}_0$:
 $${L}_{CLIP}(x, d) = D_c(CLIP_{img}(\hat{x}_0), CLIP_{txt}(d)) $$

\subsubsection{Classifier-free guidance}
\label{sec:background_cfg}
Classifier-guided mechanisms face a few challenges, such as: (i) may not be robust enough in dealing with noised samples in the diffusion process,(ii) not all the information in $x$ is relevant for predicting $y$, which may cause adversarial guidance, (iii) do not work well for predicting complex $y$ like `text'.  The classifier-free guidance \cite{ho2022classifier} helps overcome this and also utilizes the knowledge gained by a pure generative model. A conditional generative model is trained to act as both conditional and unconditional (by dropping out the conditional signal by 10-20\% during the training phase). The above equation (section 3.3.1) can be reinterpreted as \cite{dieleman2022guidance, glide}:

\begin{multline}
\nabla_{x_t}\mathrm{log}\, p_s(x_t|y) = \nabla_{x_t}\mathrm{log}\, p(x_t)  \\ + s.(\nabla_{x_t} \mathrm{log}\, p(x_t| y) - \nabla_{x_t}\mathrm{log}\, p(x_t)) 
\end{multline}

For $s=0$, we get an unconditional model, for $s=1$, we get a conditional model, and for $s>1$ we strengthen the conditioning signal. The above equation can be expressed in terms of noise estimates at diffusion timestep $t$, as follows:

\begin{equation}
\label{eq:cond-infer}
    \hat{\epsilon}_{\theta}(x_t|c) = \epsilon_{\theta}(x_t|\emptyset) + s \cdot (\epsilon_{\theta}(x_t|c) - \epsilon_{\theta}(x_t|\emptyset))
\end{equation}

where $c$ is the text caption representing the conditional input, and $\emptyset$ is an empty sequence or a null set representing unconditional output. Our DDIM sampling for conditioned models will utilize these estimates.

%% file: supp_02_base_setup.tex
\section{Our experimental setup}
\label{sec:experimental_setup}
As stated earlier, in this work, we aim to generate a composite image guided entirely  by free-form segments and corresponding natural textual prompts (with optional additional control conditions).  In this section, we summarize our choice of base setup, provide a running example to help explain the working of different algorithms, and provide implementation details of the base setup.

\begin{figure}[t!]
    \centering
    \includegraphics[width= \linewidth]{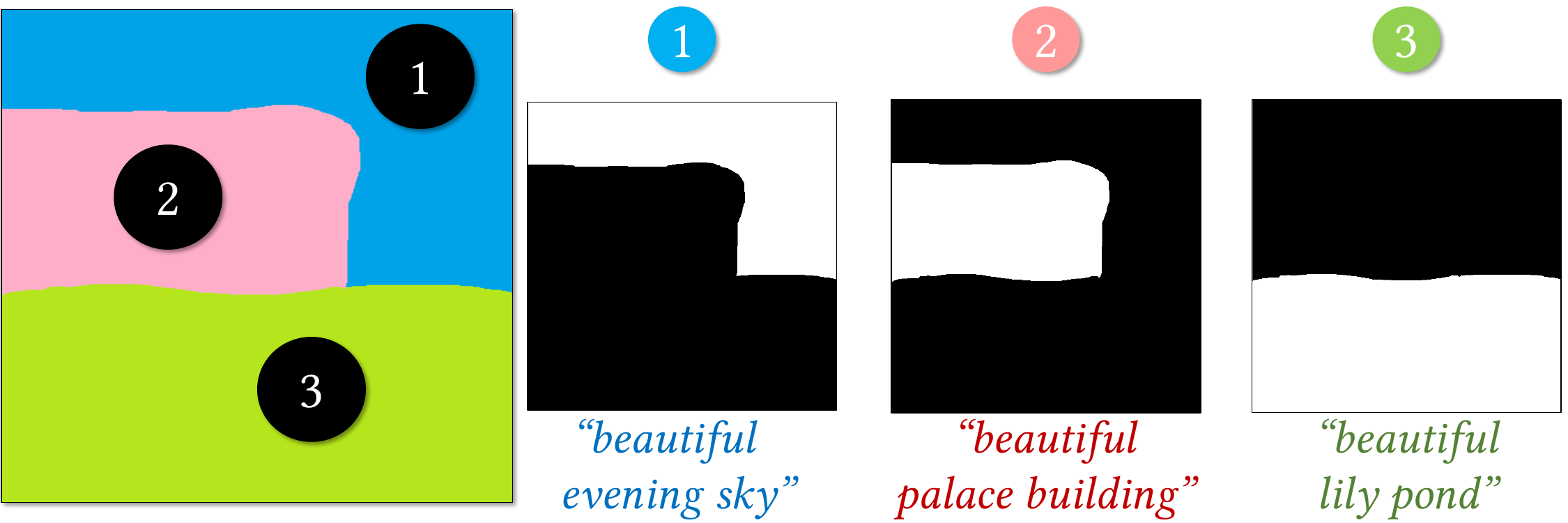}
    \caption {Running Example: Free-form segment layout and natural text input}
    \label{fig:running-example}
\end{figure}

\subsection{Running example} To explain the different algorithms, we will use a common running example. The artist's input is primarily bimodal: free-form segment layout and corresponding natural language descriptions as shown in Figure \ref{fig:running-example}. As a first step common to all the algorithms, the segment layout is converted to segment masks as one-hot encoding vectors where `$0$' represents the absence of pixel information, and `$1$' indicates the presence of image pixels. To standardize the outputs of the generative process, all the initial inputs (noise samples, segment layouts, masks, reference, and background images) and the generated images in this paper are of 512x512 pixel dimensions. Additionally, in the case of latent diffusion setup, we downsize the masks, encode the reference images, and sample the noise into 64x64 pixels corresponding to the latent space dimensions of the model. 

\subsection{Implementation details }
We choose open-domain diffusion model architecture, namely \textit{Stable Diffusion} \cite{stable_diffusion}, to serve as base architectures for our composite diffusion methods. Table \ref{tab:setup_summary} provides a summary of the features of the base setup. 
The diffusion model has a U-Net backbone with a cross-attention mechanism, trained to support conditional diffusion. We use the pre-trained text-to-image diffusion model (Version 1.5) that is developed by researchers and engineers from CompVis, Stability AI, RunwayML, and LAION and is trained on 512x512 images from a subset of the LAION-5B dataset. A frozen CLIP ViT-L/14 text encoder is used to condition the model on text prompts. For scheduling the diffusion steps and  sampling the outputs, we use DDIM \cite{song2020denoising}.

% Table
\begin{table}[ht!]%
\caption{Summary of features of  the base setup}
\label{tab:setup_summary}
\begin{minipage}{\columnwidth}
\begin{center}
\begin{tabular}{ll}
  \toprule
  Feature & Setup  \\ 
  \midrule
  Diffusion Space   & Latent\\ 
  Conditionality & Conditional\\
  Guidance      & Classifier-free\\
  Model Size  & $\approx$ 850 million\\
  Open Domain Models      & StabilityAI \\
  Sampling Method  & DDIM \\
  \bottomrule
\end{tabular}
\end{center}
\bigskip\centering
\end{minipage}
\end{table}

\begin{figure*}[t!]
    \centering
    \includegraphics[width=\linewidth]{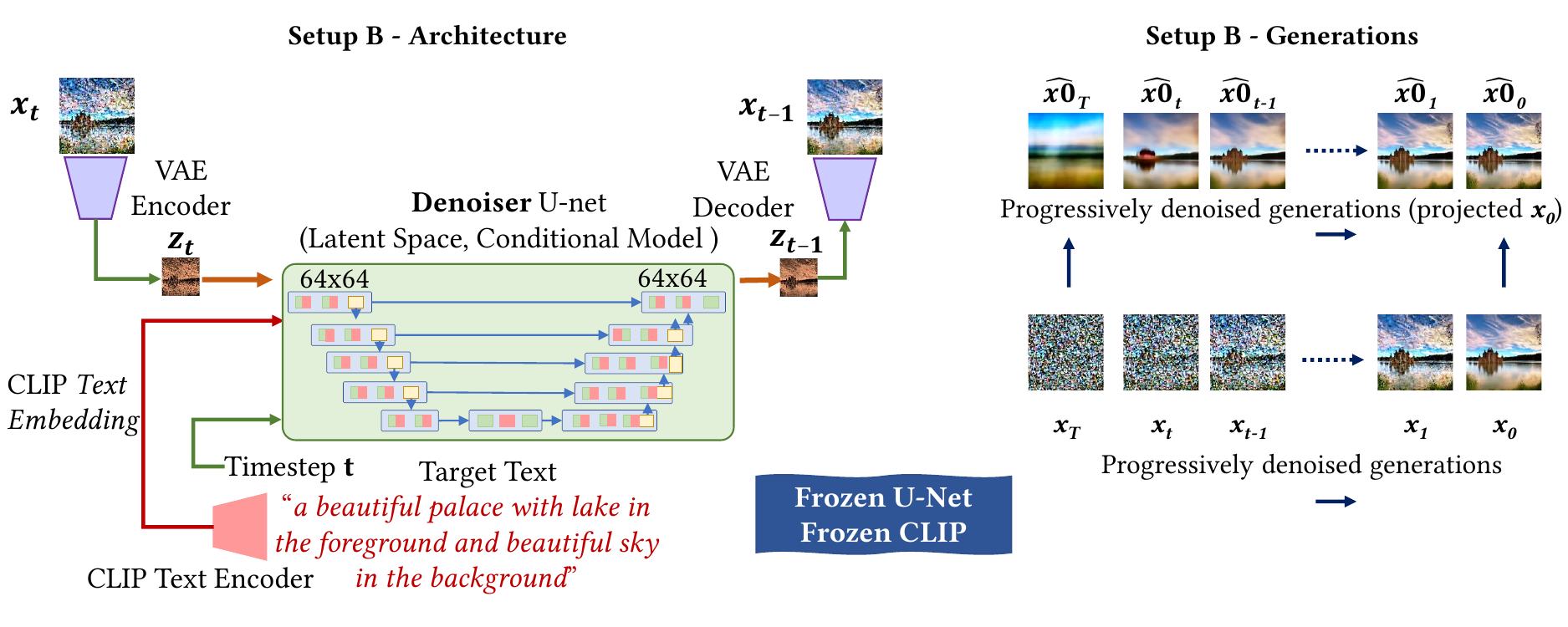}
    \caption {Base setup generation with latent-space diffusion and classifier-free implicit guidance}
    \label{fig:setup-B}
\end{figure*}

\begin{algorithm}[ht!]
\caption{Text-to-Image generation in the base setup}
\label{alg:setup-B}
\textbf{Input}
Target text description $d$,\\
Initial image, $x_T \sim \mathcal{N}(0, \textbf{I})$, Number of diffusion steps = $k$.

\textbf{Output:}  An output image, $x_{0}$, which is sufficiently grounded to input $d$.

$z_T \leftarrow \mathcal{E}(x_T)$, \tcp*{Encode into latent space} 
$d_z \leftarrow \mathcal{C}(d)$ \tcp*{Create CLIP text encoding} 

    \For {\textbf{all} $t$ \textbf{from} $k$ to $1$} {
        $z_{t-1} \leftarrow Denoise(z_t , d_z)$ \tcp*{Denoise using text-condition and DDIM}   
    }
  
\Return  $x_{0} \leftarrow  \mathcal{D} (z_{0})$ \tcp*{Final Image}
\end{algorithm}

We describe image generation through this setup in the next  section.

\subsection{Text-to-Image generation in the base setup}
In this setup (refer to Figure \ref{fig:setup-B}), a pixel-level image ($x$) is first encoded  into a  lower-dimensional latent-space representation with the help of a variational autoencoder(VAE) ($\mathcal{E}(x)\rightarrow  z$). The diffusion process then operates in this latent space. 
This setup uses a conditional\footnote {In practice, the model is trained to act as a both conditional and unconditional model. An empty text prompt is used for unconditional generation along with the input text prompt for conditional generation. The two results are then combined to generate a better quality denoised image. Refer to section \ref{sec:background_cfg}.} diffusion model which is pre-trained on natural text using CLIP encoding.  For a generation,  the model takes CLIP encoding of the  natural text ($\mathcal{C}(d) \rightarrow d_{CLIP} $) as the conditioning input and directly infers a denoised sample $z_t$ without the help of an external classifier  (classifier free guidance) \cite{ho2022classifier}. Mathematically, we use equation \ref{eq:cond-infer} for generating the additive noise $\hat{\epsilon}_{\theta}$ at timestep $t$, and use equation \ref{eq:ddim_sample} for generating $z_t$ from $\hat{\epsilon}_{\theta}$ via DDIM sampling. After the diffusion process is over, the resultant latent $z_0$ is decoded back to pixel-space ($\mathcal{D}(z_0) \rightarrow  x_0$).

 As stated earlier, spatial information cannot be adequately described through only text conditioning. In the next section, we extend the existing in-painting methods to support Composite Diffusion. However, we shall see that these methods do not fully satisfy our quality desiderata which leads us to the development of our approach for Composite Diffusion as described in the main paper.

%% file: supp_02b_serial_inpainting.tex
\section{Composite Diffusion through serial inpainting} 
\label{sec:serial_inpainting}

Inpainting is the means of filling in missing portions or restoring the damaged parts of an image. It has been traditionally used to restore damaged photographs and paintings and (or) to edit and replace certain parts or objects in digital images\cite{bertalmio2000image}.  Diffusion models have been quite effective in inpainting tasks. A portion of the image, that needs to be edited, is marked out with the help of a mask, and then the content of the masked portion is generated through a diffusion model - in the context of the rest of the image, and sometimes with the additional help of a text prompt \cite{repaint, avrahami2022blended, latent_blended}. 

An obvious question is: Can we serially  (or repeatedly ) apply inpainting to achieve Composite Diffusion?  In the following section, we develop our implementation for serial inpainting and discuss issues that arise with respect to Composite Diffusion achieved through these means. The implementation also serves as the baseline for comparing our main Composite Diffusion algorithms (Algo. \ref{alg:scaff} and Algo. \ref{alg:harmon}). 

% described in the main paper.

\begin{figure}[hb!]
    \centering
    \includegraphics[width=\linewidth]{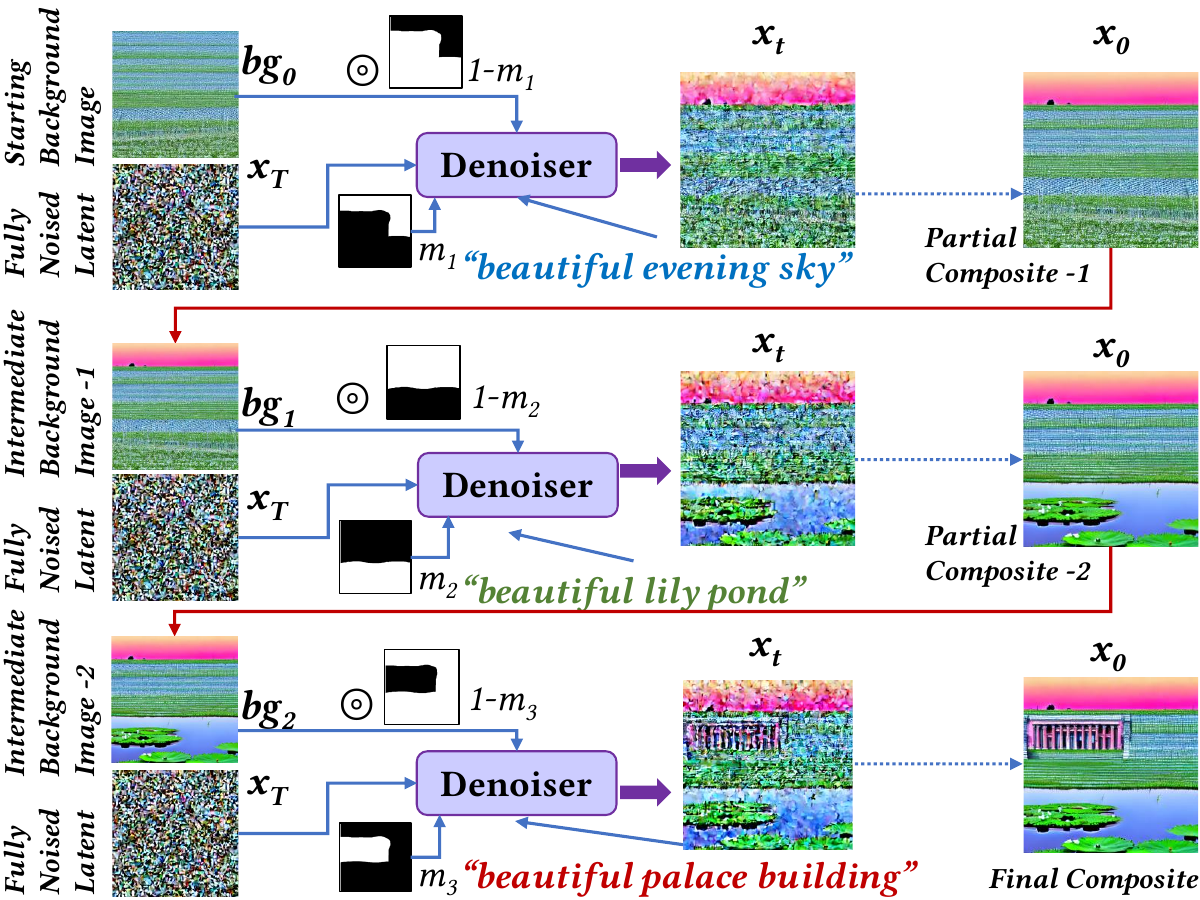}
    \caption {Diffusion steps in the algorithm for Serial Inpainting. Starting with an initial background image ${bg}_0$, we inpaint a segment into it to get $x_0$. The new image $x_0$ serves as the background image for the next stage inpainting process to generate the new $x_0$ with the inpainted second segment. The process is repeated till we have inpainted all the segments. The final $x_0$ is the generated \textit{composite}.}
    \label{fig:serial-inpaint}
\end{figure}

\subsection{Serial Inpainting - algorithm and implementation}
\label{sec:impl_serial_inpainting}

The method essentially involves successive application of the in-painting method for each segment of the layout. We start with an initial background image ($I_{bg}$) and repeatedly apply the in-painting process to generate segments specified in the free-form segment layout and text descriptions (refer to Algo. \ref{alg:serial-inpaint} for details). The method is further explained in Fig. \ref{fig:serial-inpaint} with the help of the running example. 

\begin{algorithm}
\caption{Serial Inpainting for composite creation }
\label{alg:serial-inpaint}

\textbf{Input:} Set of segment masks $m^i \in M$, set of segment descriptions $d^i \in D$, background image $I_{bg}$,
initial image, $x_T \sim \mathcal{N}(0, \textbf{I})$

\textbf{Output:}  An output image, $x_{comp}$, which is sufficiently grounded to  the inputs of segment layout and segment descriptions. \\

$z_T \leftarrow \mathcal{E}(x_T)$, \tcp*{Encode into latent space} 
$\forall i, m^i_z \leftarrow Downsample(m^i)$ \tcp*{Downsample all masks to latent space}
$\forall i, d^i_z \leftarrow \mathcal{C}_{CLIP}(d^i)$ \tcp*{Generate CLIP encoding for all text descriptions}

\For{\textbf{all segments} $i$  \textbf{from} $1$ to $n$}  
    {
    
    $z_{bg}^{masked} \leftarrow \mathcal{E}(I_{bg} \odot (1-m^i))$ \tcp*{Encode masked background image}
    $z_{bg} \leftarrow Inpaint(z_T, z_{bg}^{{masked}}, m^i_z,d^i_z)$\tcp*{Inpaint the segment}  
    $I_{bg} \leftarrow \mathcal{D}  (z_{bg})$ \tcp*{Decode the latent to get the new reference image}
    } 
    
\Return  $x^{comp} \leftarrow I_{bg}$ \tcp*{Final composite}
\end{algorithm}

We base our implementation upon the specialized in-painting method developed by RunwayML for Stable Diffusion \cite{stable_diffusion}. This in-painting method extends the U-net architecture described in the previous section to include additional input of a masked image.  It has 5 additional input channels (4 for the encoded masked image and 1 for the mask itself) and a checkpoint model which is fine-tuned for in-painting. 

\begin{figure*}[t!]
    \centering
    \includegraphics[width = \linewidth]{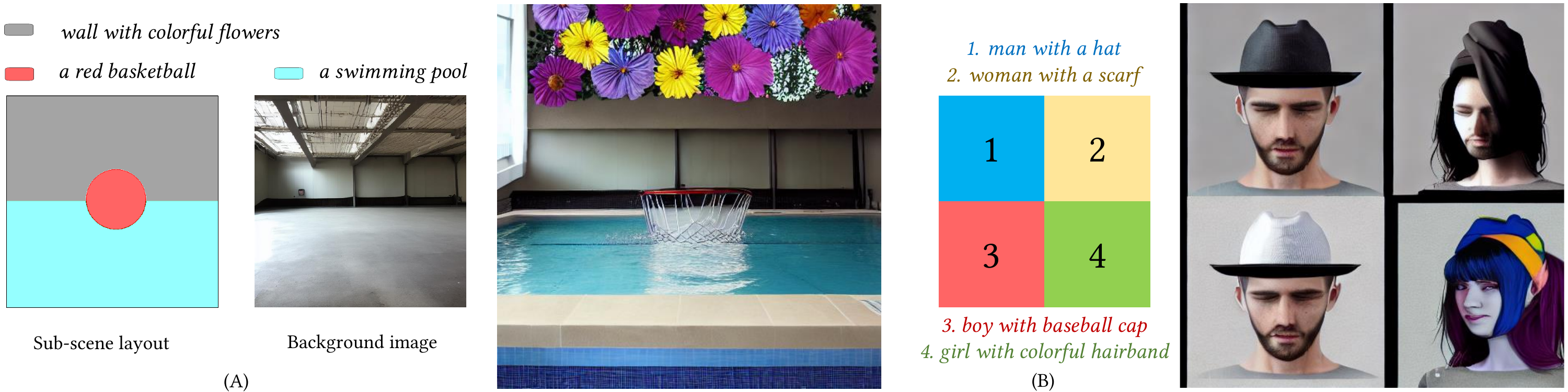}
     \caption {Some of the issues in serial-inpainting:  (A) The background image plays a dominant part in the composition, and sometimes the prompt specifications are missed if the segment text-prompt does not fit well into the background image context, e.g., missing red basketball in the swimming pool, (B) The earlier stages of serial-inpainting influence the later stages; in this case, the initial background image is monochrome black, the first segment is correctly generated but in the later segment generations, the segment-specific text-prompts are missed and duplicates are created.}
    \label{fig:issues-inpainting}
\end{figure*}

\subsection{Issues in Composite Diffusion via serial inpainting}
The method is capable of building good composite images. However, there are a few issues. One of the main issues with the serial inpainting methods for Composite Diffusion is the \textit{dependence on an initial background image}. Since this method is based on inpainting, the segment formation cannot start from scratch. So a suitable background image has to be either picked from a collection or generated anew. If we generate it anew, there is no guarantee that the segments will get the proper context for development. This calls for a careful selection from multiple generations. Also because a new segment will be generated in the context of the underlying image, this sometimes leads to undesirable consequences. Further, if any noise artifacts or other technical aberrations get introduced in the early part of the generation, their effect might get amplified in the repeated inpainting process. Some other issues might arise because of a specific inpainting implementation. For example, in the method of inpainting that we used (RunwayML Inpainting 1.5), the mask text inputs were occasionally  missed and sometimes the content of the segments gets duplicated. Refer to Fig. \ref{fig:issues-inpainting} for visual examples of some of these issues.
 
All these issues motivated the need to develop our methods, as described in the main paper, to support Composite Diffusion. We compare our algorithms against these two baselines of (i) basic text-to-image algorithms, and (ii) serial inpainting algorithms. The results of these comparisons are presented in the main paper with some more details available in the later sections of this Appendix.

%% file: supp_03_our_method_details.tex
\section{Our method: details and features}
\label{sec:our_method_implementation}
In the main paper, we presented a generic algorithm that is applicable to any diffusion model that supports \textit{conditional generation with classifier-free implicit guidance}. Here, we present the implementation details and elaborate on a few downstream applications of Composite Diffusion. 

\subsection{Implementation details of the main algorithm}
\begin{figure*}[t!]
    \centering
    \includegraphics[width=\linewidth]{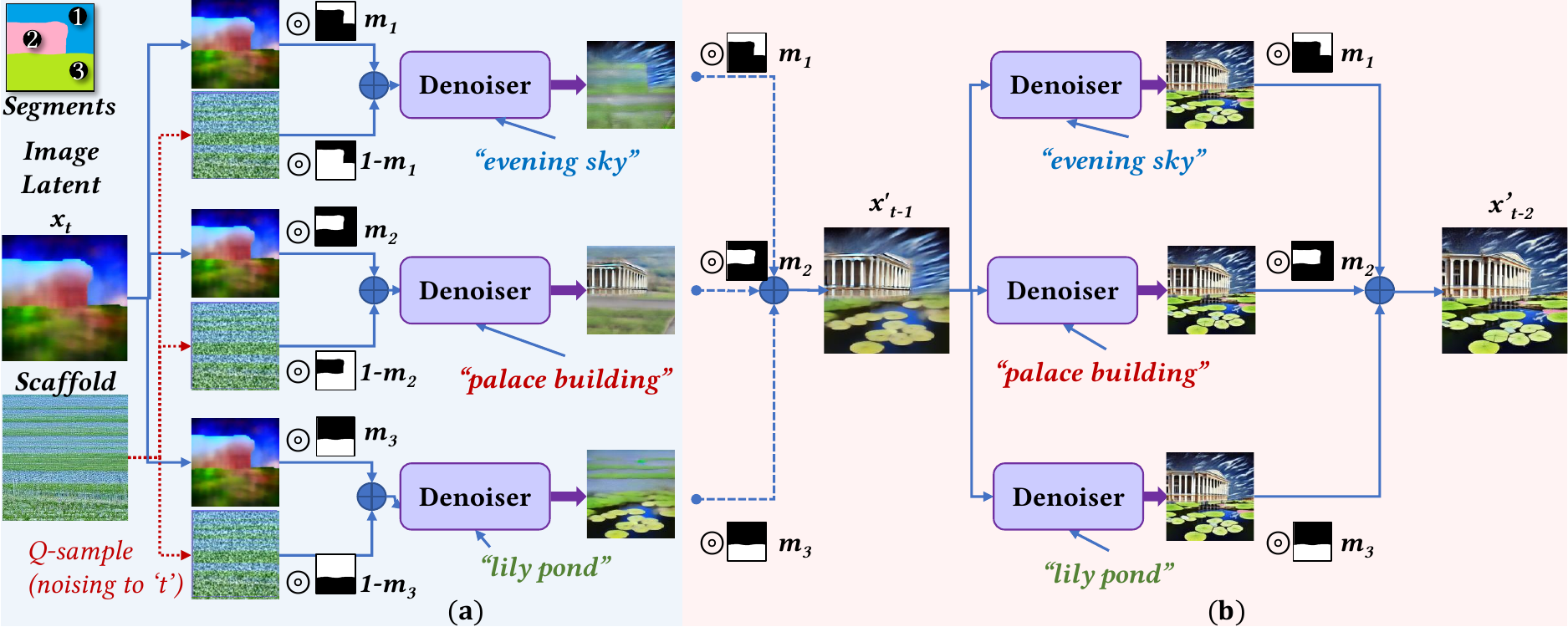}
    \caption {Typical diffusion steps in the two-stage Composite Diffusion process using a scaffolding image: \textbf{(a)} During the scaffolding stage, each segment is generated in \textit{isolation} using a separate diffusion process after composing with the noised scaffolding image. \textbf{(b)} During the harmonizing stage, the final composed latent from the scaffolding stage is iteratively denoised using separate diffusion processes with segment-specific conditioning information; the segments are \textit{composed} after every diffusion timestep for harmonization.}
    \label{fig:our-algo}
\end{figure*}

 In the previous Appendix section \ref{sec:experimental_setup}, we detailed the actual base model which we use as the example implementation of Composite Diffusion. Since the base setup operates in latent diffusion space, to implement our main Composite Diffusion algorithm in this setup, we have to do two additional steps: \textbf{(i)} Prepare the input for latent diffusion by decoding all the image latents through a VAE to 64x64 latent pixel space, \textbf{(ii)} After the Composite Diffusion process (refer to Fig. \ref{fig:our-algo} for the details of typical steps),  use a VAE decoder to decode the outputs of the latent diffusion model into the 512x512 pixel space. Since the VAE encoding maintains the spatial information, we either directly use a 64x64 pixel segment layout, or  downsize the resulting masks to 64x64 pixel image space.

As mentioned in the main paper, for supporting additional control conditions in Composite Diffusion, we use the Stable Diffusion v1.5 compatible implementation of ControlNet \cite{zhang2023adding}. ControlNet is implemented as a parallel U-Net whose weights are copied from the main architecture, but which can be trained on particular control conditions \cite{zhang2023adding} such as canny edge, lineart, scribbles, semantic segmentations, and open poses. 

In our implementation, for supporting \textit{control conditions} in segments, we first prepare a control input for every segment. The controls  that we experimented with included lineart, open-pose, and scribble. Each segment has a separate control input that is designed to be formed in a 512x512 image space but only in the region that is specific to that segment.   Each control input is then passed through an encoding processor that creates a control condition that is embedded along with the text conditioning. ControlNets convert image-based conditions to 64 × 64 feature space to match the convolution size: 
$c_f = \mathcal{E}(c_i)$  (refer to equation 9 of \cite{zhang2023adding}), 
where $c_i$ is the image space condition, and $c_f$ is the corresponding converted feature map. 

Another important aspect is to use a ControlNet model that is particularly trained for the type of control input specified for a segment. However, as shown in the main paper and also illustrated in Fig. \ref{fig:teaser}, more than one type of ControlNets can be deployed for different segments for achieving Composite Diffusion.

\subsection{Example runs}
With reference to the running example shown in the main paper, we present the different stages of the evolution of a composite image using Serial Inpainting and our Composite Diffusion algorithms. Refer to Figures \ref{fig:gen-steps}, \ref{fig:ex-run-si-a}, \ref{fig:ex-run-si-b}, \ref{fig:ex-run-si-c}, and \ref{fig:ex-run-cd}. To standardize our depiction, we run each algorithm for a total of 50 diffusion steps using DDIM as the underlying sampling method. The figures show every alternate DDIM step. 

\begin{figure*}[t!]
\label{fig:serial}
    \centering
    \includegraphics[width= \textwidth]{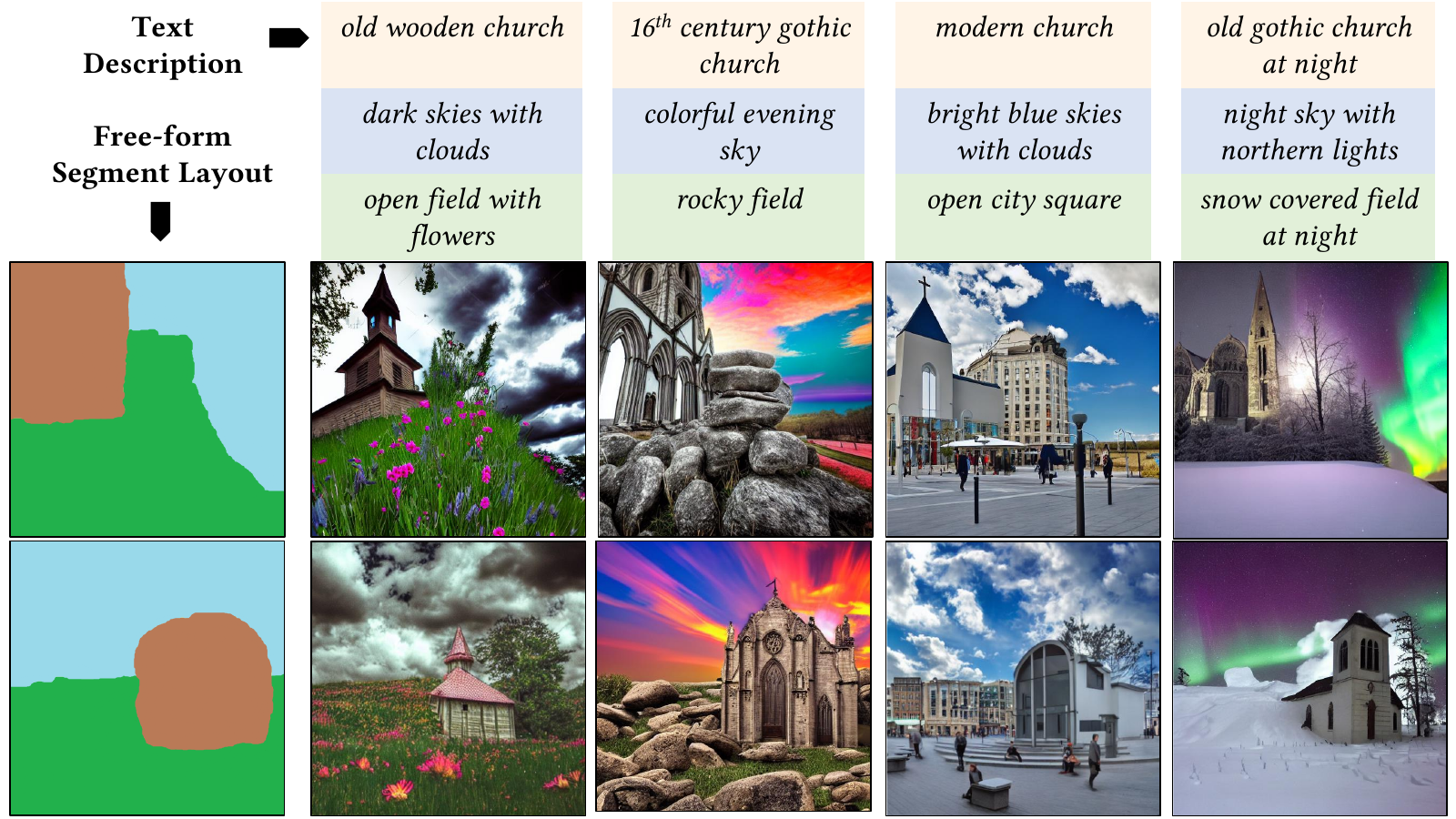}
    \caption {By controlling layout, and/or text inputs independently an artist can produce diverse pictures through Composite diffusion methods. Note how the segment layout is used as a coarse-grained guide for \textit{sub-scenes} within an image and not as an outline of  shapes for the objects as happens in many object segment models.} 
    \label{fig:church-gens}
\end{figure*}

%% file: supp_03b_features_limitations_societal.tex
\subsection{Personalization at a scale}
\label{sec:personalizing_at_scale}
One of the motivations for composite image generation is to produce a controlled variety of outputs. This is to enable customization and personalization at a scale. Our Composite Diffusion models help to achieve variations through: (i) variation in the initial noise sample, (ii) variation in free-form segment layout, (iii) variation through segment content, and (iv) variation through fine-tuned models. 

\subsubsection{Variation through Noise} This is applicable to all the generative diffusion models. The initial noise sample massively influences  the final generated image.  This initial noise can be supplied by a purely random sample of noise or by an appropriately noised (\textit{q-sampled}) reference image. Composite Diffusion further allows us to keep these initial noise variations particular to a segment. This method, however, only gives more variety but not any control over the composite image generations.

\subsubsection {Variation through segment layout}
 We can introduce controlled variation in the spatial arrangement of elements or regions of an image by changing the segment layout while keeping the segment descriptions constant. Refer to figure \ref{fig:church-gens} for an illustration where we introduce two different layouts for any given set of segment descriptions.

\subsubsection {Variation through text descriptions}
Alternatively, we can keep the segment layout constant, and change the description of the segments (through text or control conditions) to bring controlled variation in the content of the segments. Refer to figure \ref{fig:church-gens} for an illustration where each of the three columns represents a different set of segment descriptions for any of the segment layouts.

\begin{figure*}[htp!]
    \centering
    \includegraphics[width=.96\textwidth]{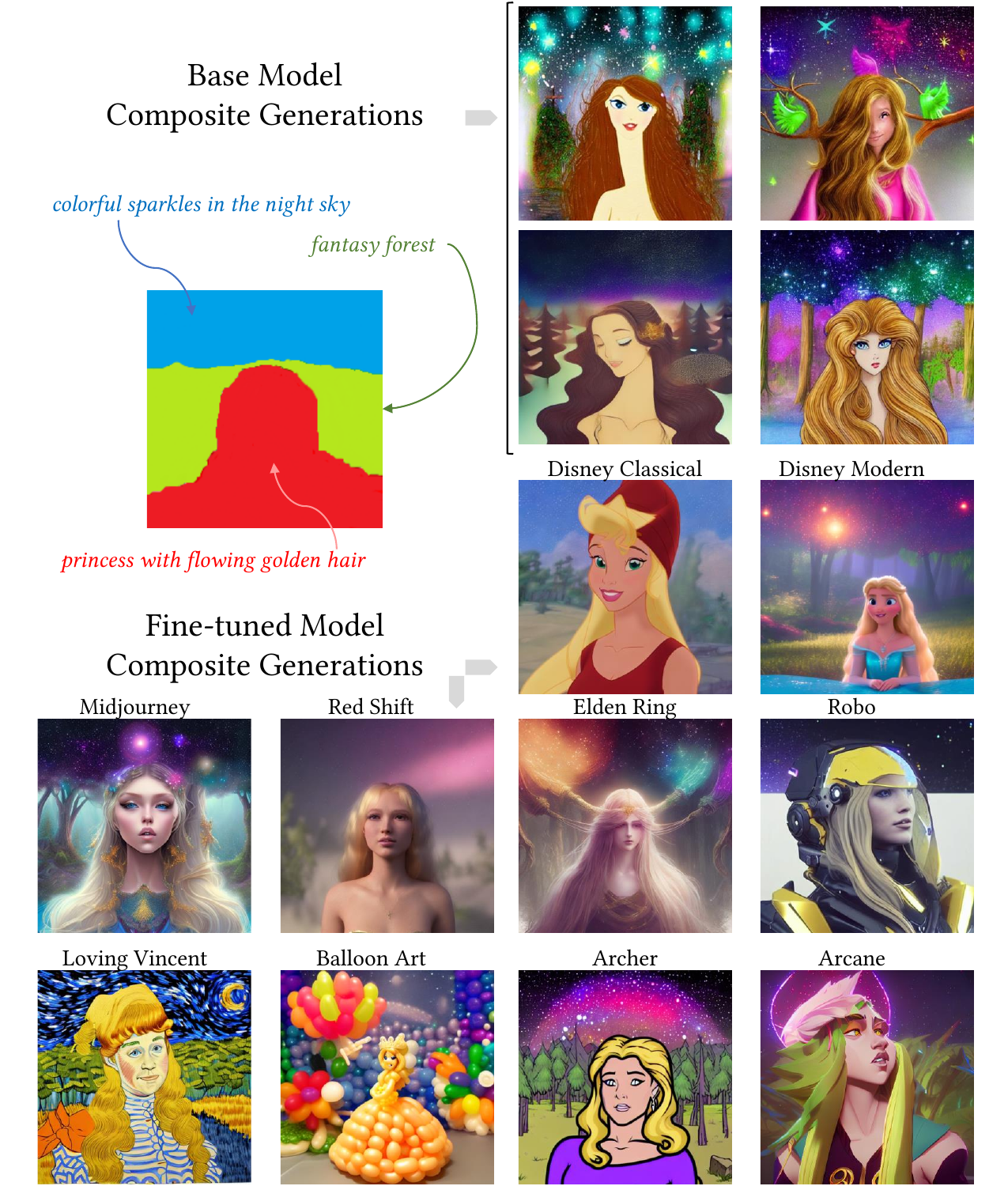}
    \caption {Composite generations using fine-tuned models. Using the same layout and same captions, but different specially trained fine-tuned models, the generative artwork can be customized to a particular style or artform. Note that our Composite Diffusion methods are plug-and-play compatible with these different fine-tuned models.}
    \label{fig:fine_tuned_gens}
\end{figure*}

\begin{figure*}[htp!]
    \centering
    \includegraphics[width=.95\textwidth]{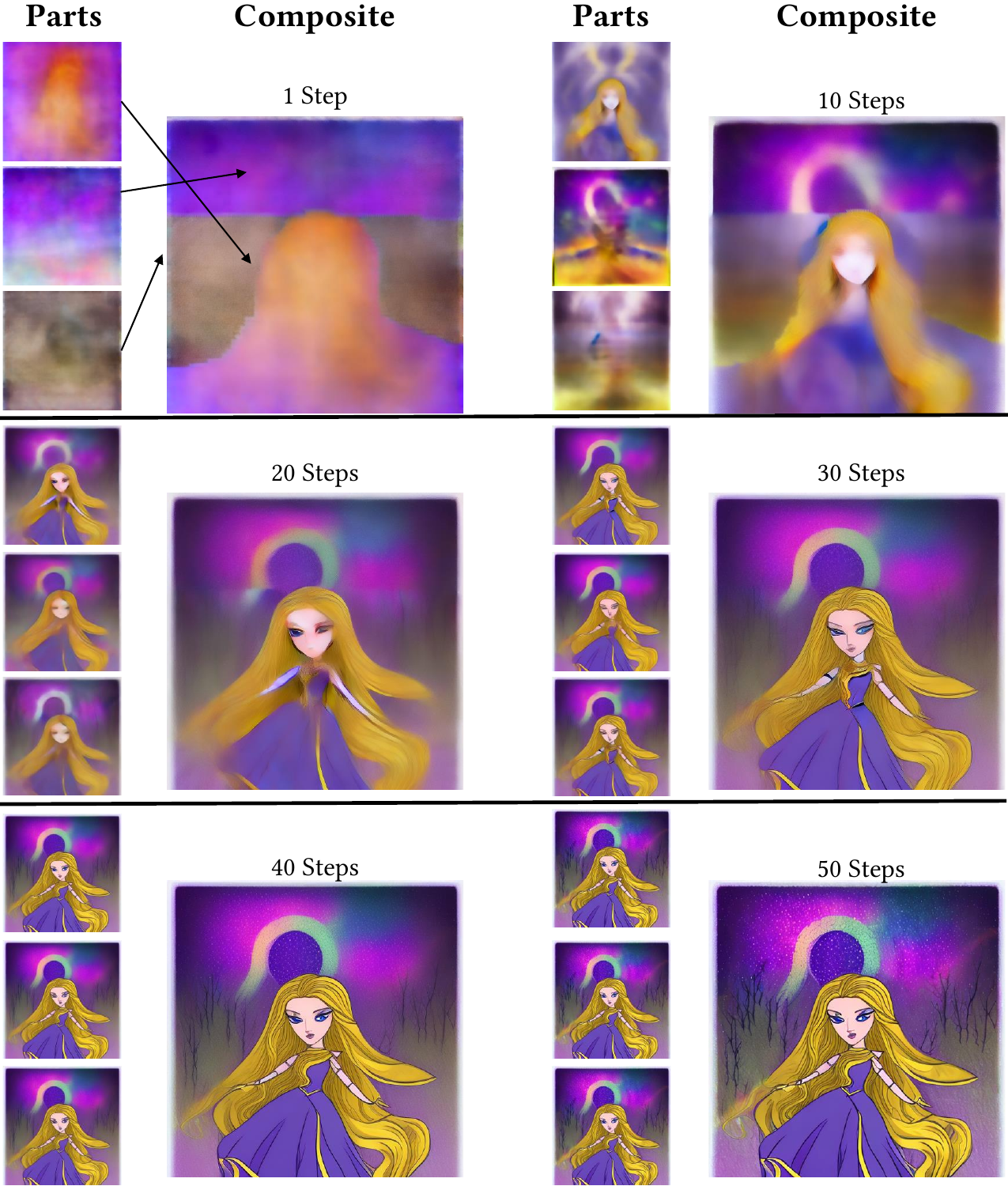}
    \caption {Composite Diffusion generation using the inputs specified in  Fig. \ref{fig:fine_tuned_gens}, a scaffolding factor of $\kappa = 30$, and 50 DDIM diffusion steps. The figure shows segment latents and composites after the timesteps 1, 10, 20, 30, 40, and 50. Note that for the first 15 steps (scaffolding stage), the segment latents develop \textit{independently}, while for the remaining 35 steps (harmonization stage), the latents develop \textit{in-the-context} of all other segments.}
    \label{fig:gen-steps}
\end{figure*}

\subsubsection{Specialized fine-tuned models }
The base diffusion models can be further fine-tuned on specialized data sets to produce domain-specialized image generations. For example, a number of fine-tuned implementations of Stable Diffusion are available in the public domain \cite{FinetunedSD}. This aspect can be extremely useful when creating artwork customized for different sets of consumers. One of the advantages of our composite methods is that as long as the fine-tuning does not disturb the base-model architecture, \textit{our methods allow a direct {plug-and-play} with the fine-tuned models}. 

Figure \ref{fig:fine_tuned_gens} gives an illustration of using 10 different public domain fine-tuned models with our main Composite Diffusion algorithm for generating specific-styled artwork. The only code change required for achieving these results was the change of reference to the fine-tuned model and the addition of style specification in the text prompts.

In the following sections, we discuss some of the limitations of our approach and provide a brief discussion on the possible societal impact of this work.

\subsection{Limitations}
\label{sec:limitations}
Though our method is very flexible and effective in a variety of domains and composition scenarios, we do encounter some limitations which we discuss below:

\textit{Granularity of sub-scenes:} The segment sizes in the segment layout are limited by the diffusion image space. So, as the size of the segment grows smaller, it becomes difficult to support sub-scenes. Our experience has shown that it is best to restrict the segment layout to 2-5 sub-scenes. Some of this is due to the particular model that we use in implementation. Since Stable Diffusion is a latent space diffusion model \cite{stable_diffusion}, the effective size for segment layout is only 64x64 pixels. If we were operating directly in the pixel space, we would have considerably more flexibility because of 8 fold increase in the segment-layout size of 512x512 pixels.

\textit{Shape conformance:} In the only text-only conditioning case,  our algorithms do perform quite well on mask shape conformance. However, total shape adherence to an object only through the segment layout is sometimes difficult. Moreover, in the text-only condition case, while generating an image within a segment the whole latent is in play. The effectiveness of a generation within the segment is influenced by how well the scaffolding image is conducive as well as non-interfering to the image requirements of the segment. This creates some dependency on the choice of scaffolding image. Further, extending the scaffolding stage improves the conformance of objects to mask shapes but there is a trade-off with the overall harmony of the image. 

So in the case where strict object conformance is required, we recommend using the control condition inputs as specified in our algorithm, though this might reduce the diversity of the images that text-only conditioning can produce.

\textit{Training and model limitations:}The quality and variety in generated object configurations are heavily influenced by the variety that the model encounters in the training data. So, as a result, not all object specifications are equal in creating quality artifacts. Although we have tested the models and methods on different kinds of compositions, based on our limited usage we cannot claim that model will equally work well for all domains. For example, we find that it works very well on closeup faces of human beings but the faces may get a bit distorted when we generate a full-length picture of a person or a group of people.

\subsection{Societal impact} 
\label{sec:societal_impact}
Recent rapid advancements in generative models have been so stunning that they have left many people in society (and in particular, the artists)  both worried and excited at the same time. On one hand, these tools, especially when they are getting increasingly democratized and accessible, give artists an enabling tool to create powerful work in lesser time. On the other hand, traditional artists are concerned about losing the business critical for their livelihood to amateurs \cite{artist_impact_wired}. Also, since these models pick off artistic styles easily from a few examples, the affected artists, who take years to build their portfolio and style, might feel shortchanged. Also, there is a concern that AI art maybe be treated at the same level and hence compete with traditional art. 

We feel that generative AI technology is as disruptive as photography was to realistic paintings. Our work, in particular, is based on Generative Models that can add to the consequences. However, since our motivation is to help artists improve their workflow and create images that self-express them, this modality of art may also have a very positive impact on their art and art processes. With confidence tempered with caution, we believe that it should be a welcome addition to an artist's toolkit.

\begin{figure*}[htp!]
    \centering
    \includegraphics[width= .87\textwidth]{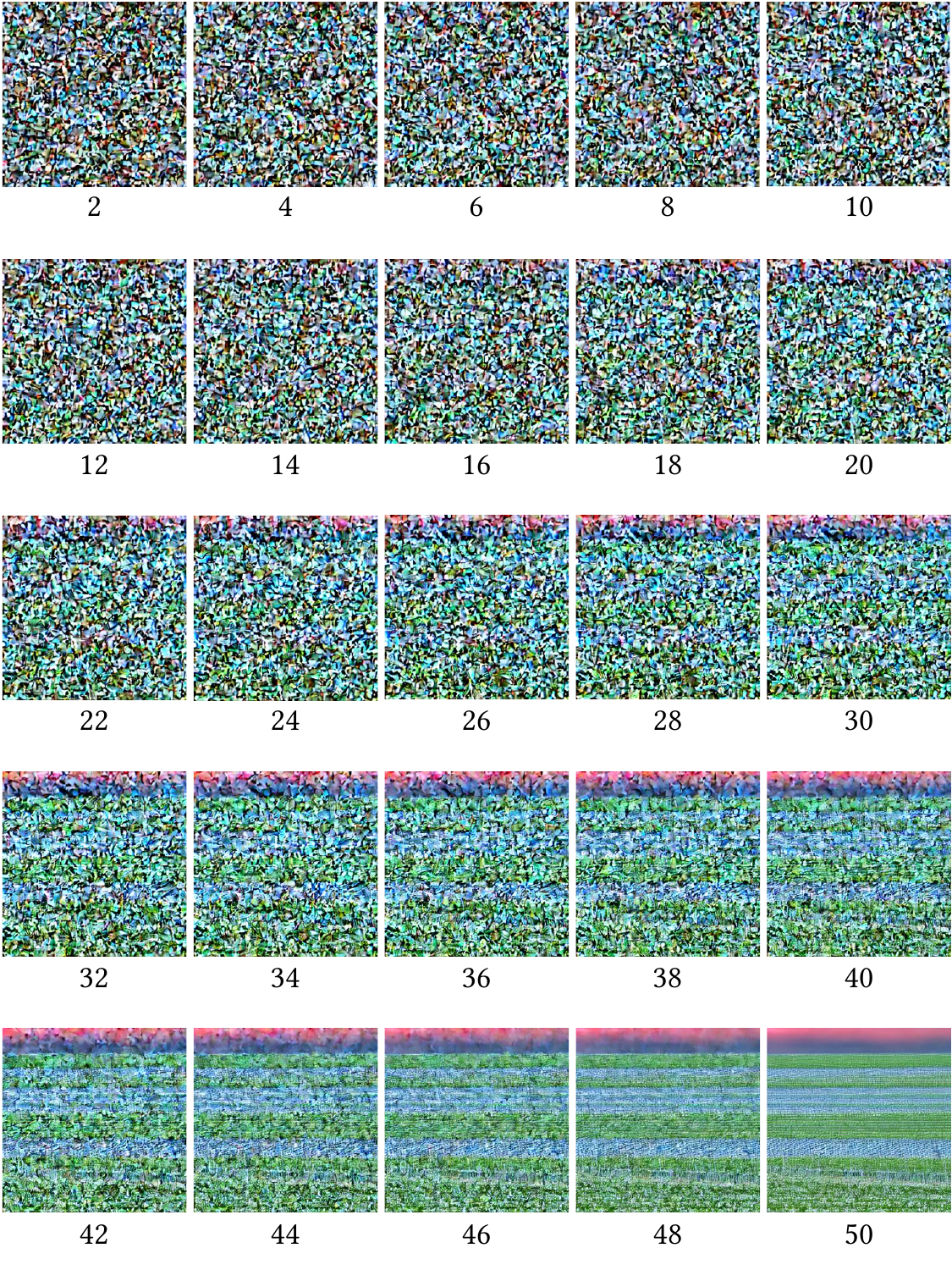}
    \caption {Composite generation using \textbf{Serial Inpainting}. The figure shows the development stages for the \textbf{Segment 1}. The inputs to the model are as shown in the running example of Fig. \ref{fig:running-example}.}
    \label{fig:ex-run-si-a}
\end{figure*}

\begin{figure*}[htp!]
    \centering
    \includegraphics[width= .87\textwidth]{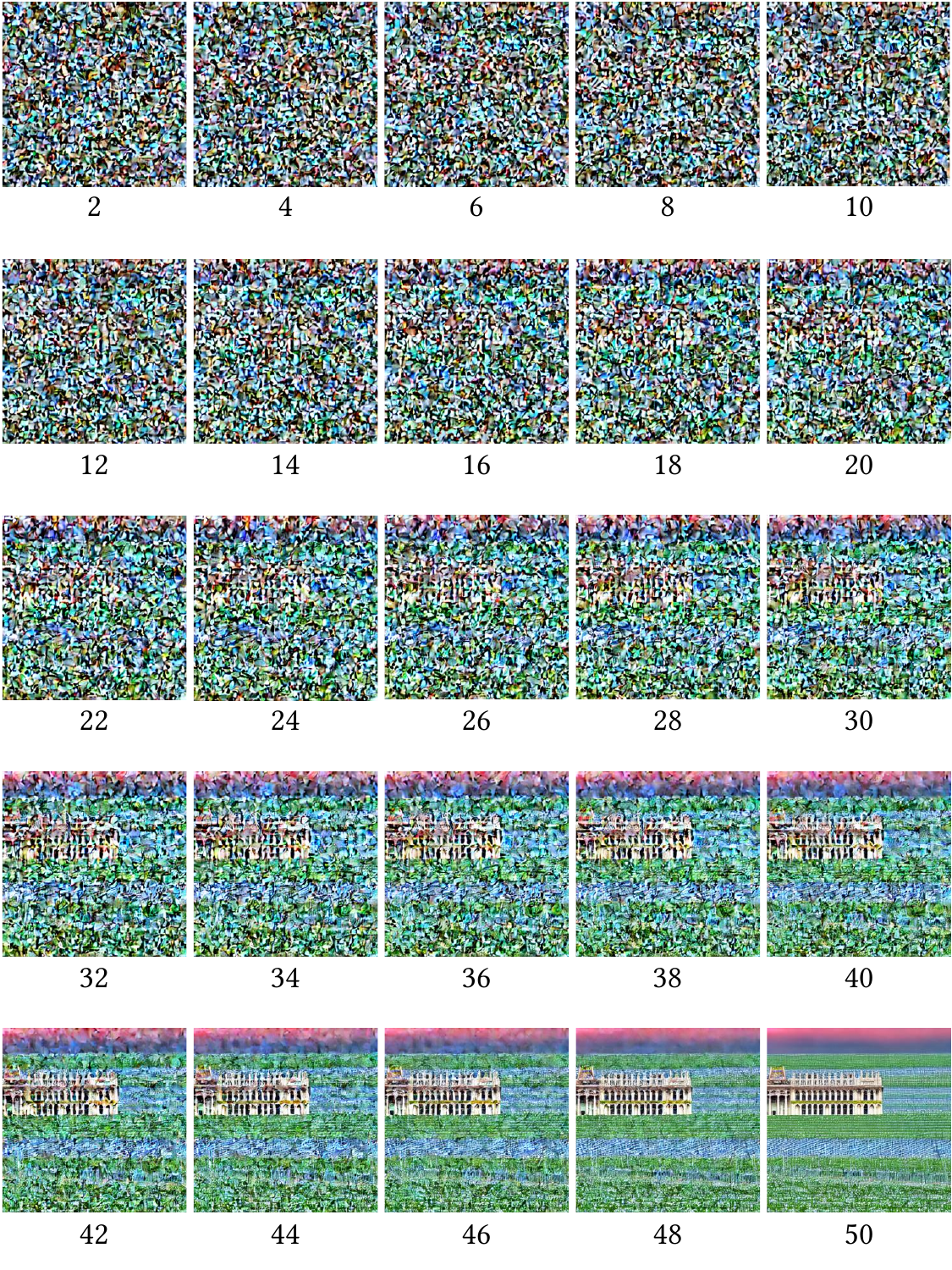}
    \caption {Composite generation using \textbf{Serial Inpainting}. The figure shows the development stages for the \textbf{Segment 2}. The inputs to the model are as shown in the running example of Fig. \ref{fig:running-example}.}
    \label{fig:ex-run-si-b}
\end{figure*}

\begin{figure*}[htp!]
    \centering
    \includegraphics[width= .87\textwidth]{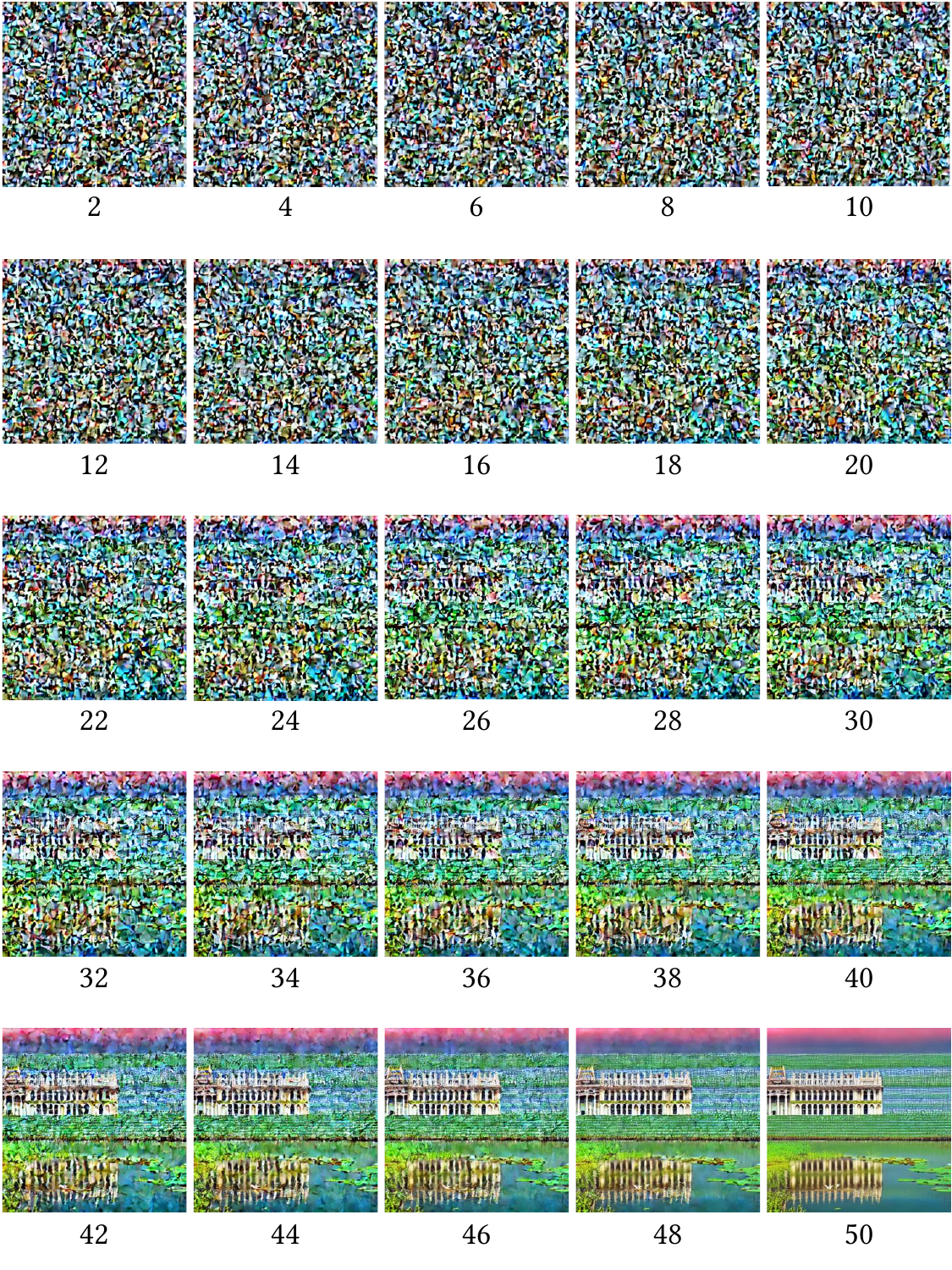}
    \caption {Composite generation using \textbf{Serial Inpainting}. The figure shows the development stages for the \textbf{Segment 3}. The inputs to the model are as shown in the running example of Fig. \ref{fig:running-example}.}
    \label{fig:ex-run-si-c}
\end{figure*}

\begin{figure*}[htp!]
    \centering
    \includegraphics[width= .87\textwidth]{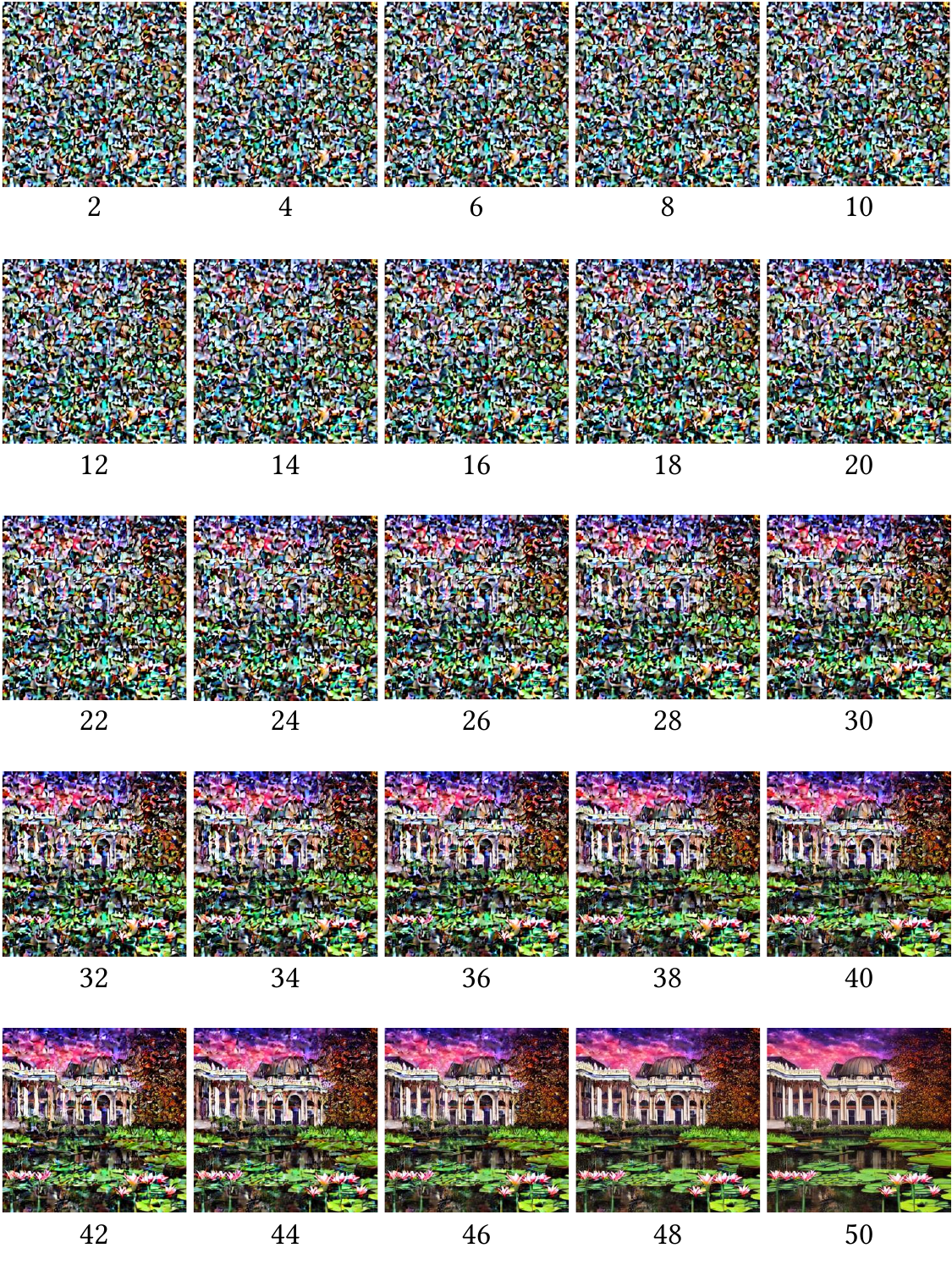}
    \caption {Composite generation using \textbf{Composite Diffusion}. The figure shows the development stages of the composite image. The inputs to the model are as shown in the running example of Fig. \ref{fig:running-example}.}
    \label{fig:ex-run-cd}
\end{figure*}

%% file: supp_04_evaluation_details.tex
\section{Human evaluation and survey details}
\label{sec:survey_results}

During the course of the project, we conducted a set
of three different surveys. A preliminary and a revised
survey were conducted on the general population, and a separate survey was conducted on artists and designers. In these surveys,
we evaluated the generated outputs of text-to-image
generation, serial inpainting generation methods, and
our composite generation methods. The text-to-image
generations serve as the first baseline (\textbf{B1}) and serial inpainting generations serve as the second baseline (\textbf{B2})
for comparison.

\begin{figure*}[ht!]
    \centering
    \includegraphics[width=\textwidth]{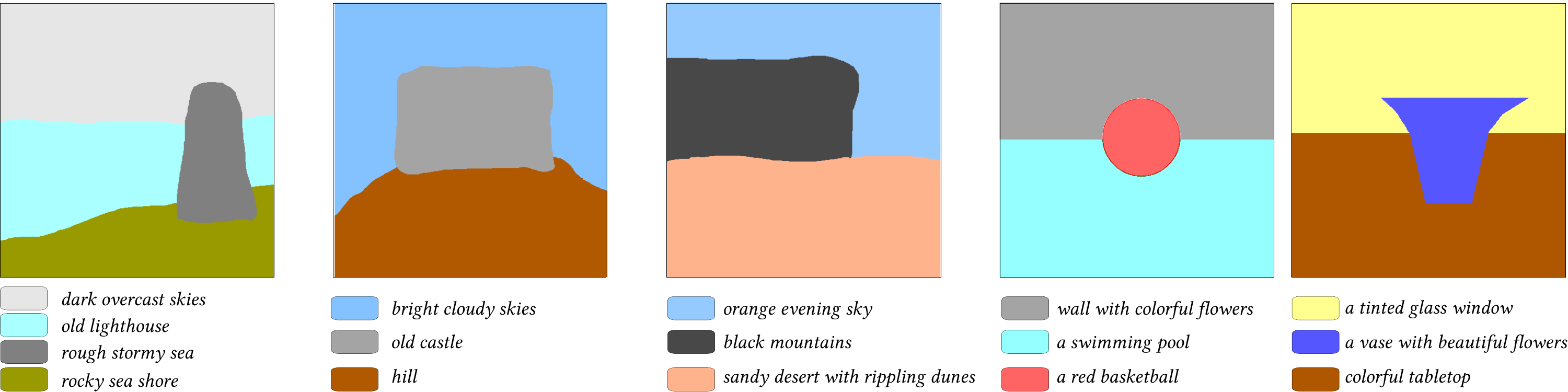}
    \caption {Segment layouts and segment text prompts as inputs for Survey sample generations }
    \label{fig:survey-segment-layouts}
\end{figure*}

\subsection{Survey design}
The survey design involved the following parts:

\subsubsection{Method for choosing survey samples} 
We designed five different input sets for the survey. The free-form segment layouts and corresponding text descriptions were chosen to bring in a variety of scenarios for the input cases. Refer to Fig. \ref{fig:survey-segment-layouts} for the inputs and to Fig. \ref{fig:image-compare} for a sample of generated images for the survey. Since we also wanted to compare our generations with the base model generations, and text-to-image model only allows a single text prompt input, we manually crafted the prompts for the first base case. This was done by: (i) creating a text description that best tries to capture the essence of different segment prompts, (ii) concatenating the different segment descriptions into a single text description. The best of these two generations were taken as the representative pictures for base models. For selecting the samples from the different algorithms we followed the following protocol. Since the underlying models are of different architecture (for example, the serial inpainting method uses a specialized inpainting model and requires a background image), we generated 3 images using random seeds for each algorithm and for each set of inputs. We then chose the best representatives (1 out of 3) from each algorithm for the survey samples. 

\begin{figure*}[ht!]
    \centering
    \includegraphics[width=\textwidth]{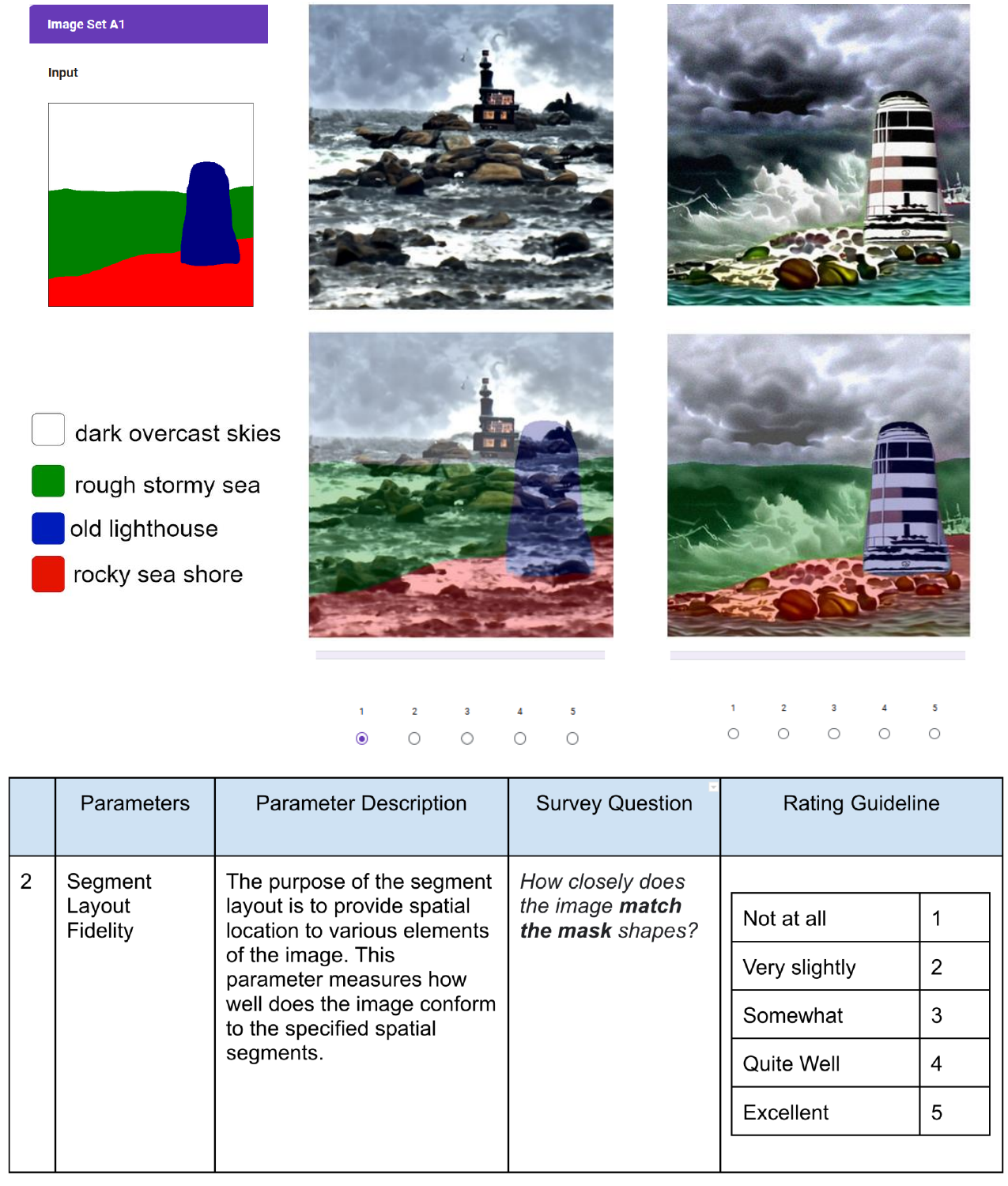}
    \caption {Snippets from the interface used for collecting responses in user evaluation}
    \label{fig:survey_interface}
\end{figure*}

\subsubsection{Survey questions:} In each of the surveys, the survey takers were presented with a Google form on the web containing anonymized and randomly sorted images generated from these three algorithms with corresponding inputs. The respondents were asked to rate each of these images on five quality parameters. We explained each quality parameter and  asked a corresponding quality question as listed below:

\begin{enumerate}
  \item \textit{Text Fidelity:} How closely does the image match the text prompts? 
  \item \textit{Mask Fidelity:}  How closely does the image match the mask shapes?
  \item \textit{Blending \& Harmony:} How well do the segments blend together and how harmonious is the overall image?
  \item \textit{Technical Quality:} How would you rate the overall technical quality of the image?  
  \item \textit{Aesthetic Quality:} How would you rate the overall aesthetic quality of the image?
\end{enumerate}

The respondents were asked to rate a generated image for a given quality parameter on a scale of 1 to 5 (semantic differential scales). We also provided a rough rating 
guideline for these parameters. Refer to Fig.\ref{fig:survey_interface} for a snapshot of the web survey.

\subsection{Survey execution:}
The details of the execution of  the three surveys are as follows:

\subsubsection{Phase 1: Preliminary survey:} We conducted this survey on a diverse set of 14 respondents who were spread across age (20-80), gender, and profession.  Our experience with the first survey gave us vital feedback on how to design the survey more effectively. For example, many surveyors said that they found it tough to take the survey as it was lengthy. There were a total of 75 rating questions that needed to be answered. So there was some fatigue due to the cognitive load. The first survey was organized in the following manner: Each set of inputs was a separate page and contained all five quality questions. On each page, the respondents were presented with 3 pics from 3 different algorithms(anonymized and randomly sorted) and were asked to rate each of the pictures on five quality parameters. We also received feedback that all the guidance information was on the front page,  and they had to revisit it several times to understand the rating guidelines and the meaning of each quality parameter. Further, some users told us that `aesthetics' influenced their rating of the other qualities; They tended to rate an image with higher aesthetics higher for other qualities as well.

\subsubsection{Phase 2: Revised survey}
We built upon this feedback, and without changing the content, restructured the survey to make it more modular for our final assessment. We also found the guidelines in \cite{bylinskii2022towards} relevant and followed them to fine-tune the survey organization. The two major changes were: (1) Each quality parameter was made into a separate survey. This was done to help the surveyors focus on one quality parameter at a time. (2) We provided guidelines for the score rating on each of the survey pages as a ready reference. 

The survey was further divided into two sets of Surveyors representing different sets of professional skills. 

\begin{itemize}

\item \textbf{Survey population: Artists and Designers (AD)}: We conducted this survey during the final phase of our project. We used the same set of images as used in the preliminary survey to collect responses from artists and designers. We took the help of Amazon M-Turk for collecting responses for this survey.  There was no restriction on whether a person took all 5 surveys or only a subset of them. There were a total of 20 respondents for each of the five surveys (where one survey was comprised of a distinct quality parameter). 

\item \textbf{Survey population: General (GP)}: 
We conducted this survey simultaneously with the above survey.  The participants in this survey were chosen from a larger general population that also included professionals such as engineers and software developer. In this case, 22 respondents completed all the five survey sets, while 48 completed at least one set.

\end{itemize}

\begin{table}[t!]
\resizebox{\columnwidth}{!}{
\begin{tabular}{|l|c|c|c|}
\hline
\textbf{Method}         & \textbf{B1} & \textbf{B2} & \textbf{Ours} \\ \hline
Content Fidelity $\uparrow$         & 3.81±1.0    & 3.22±1.07   & 3.92±0.97     \\ \hline
Spatial Layout Fidelity $\uparrow$ & 3.21±1.08   & 3.14±1.15   & 3.62±0.97     \\ \hline
Blending \& Harmony $\uparrow$     & 3.87±0.96   & 4.02±0.99   & 3.86±1.03     \\ \hline
Technical Quality $\uparrow$        & 3.85±1.02   & 3.75±1.15   & 3.6±0.98      \\ \hline
Aesthetic Quality $\uparrow$       & 3.55±0.92   & 3.55±1.0    & 3.52±0.99     \\ \hline
\end{tabular}}
\caption{Results of the survey conducted on Artists and Designers}
\label{tab:artists_designers}
\end{table}

\begin{table}[t!]
\resizebox{\columnwidth}{!}{
\begin{tabular}{|l|c|c|c|}
\hline
\textbf{Method}         & \textbf{B1} & \textbf{B2} & \textbf{Ours} \\ \hline
Content Fidelity $\uparrow$         & 2.8±1.28   & 2.38±1.13   & 3.12±1.45     \\ \hline
Spatial Layout Fidelity $\uparrow$ & 2.19±1.11   & 2.99±1.44   & 3.82±1.08     \\ \hline
Blending \& Harmony $\uparrow$     & 3.47±1.07   & 2.94±1.22   & 3.62±1.14     \\ \hline
Technical Quality $\uparrow$        & 3.33±1.11   & 2.78±1.16   & 3.39±1.14     \\ \hline
Aesthetic Quality $\uparrow$       & 3.16±1.19   & 2.66±1.26   & 3.36±1.28     \\ \hline
\end{tabular}}
\caption{Results of the survey conducted on General Population.}
\label{tab:general_population}
\end{table}

\subsection{Survey Results}
Table \ref{tab:artists_designers} presents the results of the survey for the artists and designers population, and Fig. \ref{fig:evaluation-ad} presents a graphical representation of the same for easy comparison. Since the set of images and the survey questions were the same across the two phases of the survey, we consolidate the results of general population responses. Table \ref{tab:general_population} presents the consolidated results of the survey of the general population, and Fig. \ref{fig:evaluation-gp} gives a graphical representation of the same.

\begin{table*}[htbp!]
\centering
\resizebox{\textwidth}{!}{%
\begin{tabular}{|c|c|c|c|c|c|c|}
\hline
\textbf{Kappa} & \textbf{Content Fidelity $\uparrow$} & \textbf{Spatial Layout Fidelity $\uparrow$} & \textbf{Technical Quality $\uparrow$} & \textbf{Human Preference $\downarrow$} & \textbf{Aesthetic Score$\uparrow$} & \textbf{Blending \& Harmony $\downarrow$} \\ \hline
0              & 0.2634                 & 0.278                     & 1.2612                & 3                         & 6.1809                  & 5321              \\ \hline
20             & 0.2629                 & 0.278                     & 1.2079                & 3                         & 6.1487                  & 6137              \\ \hline
40             & 0.2596                 & 0.2726                    & 1.6987                & 3                         & 6.296                   & 8078              \\ \hline
60             & 0.2627                 & 0.2757                    & 1.4186                & 4                         & 6.2565                  & 6827              \\ \hline
80             & 0.2594                 & 0.2744                    & 1.3123                & 4                         & 5.9693                  & 7235              \\ \hline
100            & 0.2579                 & 0.2773                    & 1.7702                & 3                         & 6.0798                  & 7699              \\ \hline
\end{tabular}}
\caption{Automated Method evaluation across different scaffolding factor $\kappa$ values. We observe that the general trend is that Blending \& Harmony (lower is better) progressively gets slightly worse as we move from  lower to higher $\kappa$, while the other factors remain quite similar across different $\kappa$ values.}
\label{tab:kappa_ablation}
\end{table*}

\section{Automated evaluation methods}
\label{sec:automated_evaluation_methods_details}
We find that the present methods of automated quality comparisons such as FID and IS aren't well suited for the given quality criteria. In the section below we discuss a few of the methods that are widely used in measuring the capabilities of generative models, point out their drawbacks,  and then detail our methods for automated evaluation. 

\subsection{Current approaches for automated evaluation} 
Inception score (IS), Fréchet inception distance (FID), precision and recall are some of the commonly used metrics for assessing the quality of synthetically generated images \cite{FID_NIPS, IS_NIPS, sajjadi2018assessing, borji2022pros}. IS score jointly measures the diversity and quality of generated images. FID measures the similarity between the distribution of real images and generated images. Metrics like precision and recall \cite{sajjadi2018assessing} separately capture the quality and diversity  aspects of the generator. Precision is an indicator of how much the generated images are similar to the real ones, and recall measures how good the generator is in synthesizing all the instances of the training data set \cite{borji2022pros}.

These approaches have some drawbacks to our requirement of assessing the quality of Composite Diffusion generations:  (i) These approaches require a large set of reference images to produce a statistically significant score. The distribution of the training set images is not relevant to us. We need datasets that have - an input set of sub-scene layouts along with textual descriptions of those sub-scenes, and a corresponding set of reference images., (ii) Even if we had the facility of a relevant large dataset, these methods assume that the reference images provide the highest benchmark for quality and diversity. This might not be always true as the generated images can exceed the quality of reference images and have a variety that is different from the reference set., and (iii) These methods don't measure the quality with the granularity as described in the quality criteria that we use in this paper.

\subsection{Our approach for automated evaluation}

We devise the following automated methods to evaluate the generated images based on our quality criteria. 

\textbf{Content Fidelity $\uparrow$:} The objective here is to obtain a measure of how similar the image is to each of the artist's intended content, and in this case, we use the textual descriptions as content. We compute the cosine similarity between the CLIP embeddings of the image and the CLIP embeddings of each segment's description. We then take the mean of these similarity scores. Here \textit{a greater score indicates greater content fidelity}.

\textbf{Spatial-layout Fidelity $\uparrow$:} The objective here is to measure how accurately we generate a segment's content. We use masking to isolate a segment from the image. We find the CLIP similarity score between the masked image and that segment's description. We do this for all the segments and then take the mean of these scores. Here \textit{a greater score indicates greater spatial-layout fidelity}.

\textbf{Technical Quality $\downarrow$:} The goal here is to measure if there are any degradations or the presence of unwanted artifacts in the generated images. It is difficult to define all types of degradations in an image. We consider the presence of noise as a vital form of degradation. We estimate the Gaussian noise level in the image by using the method described in \cite{chen2015efficient}. Here \textit{a lower score indicates greater technical quality}.

\textbf{Aesthetics $\uparrow$:} We use the aesthetic quality estimator from \cite{laion_aesthetic} to get an estimate of the aesthetic quality of the image. This method uses a linear layer on top of the CLIP embedding model and is trained on 4000 samples to estimate if an image is looking good or not. Here \textit{a greater score indicates greater perceived aesthetics}.

\textbf{Blending \& Harmony $\downarrow$:} We detect the presence of edges around the segment boundaries as a measure of better blending. Hence \textit{a lower value in this case indicates better blending}. 

\textbf{Human Preference $\downarrow$:} To additionally estimate the human preference we rank the images generated by the different algorithms using ImageReward\cite{xu2023imagereward}. This method uses a data-driven approach to score human preferences for a set of images. Here \textit{a greater score indicates lower preference}.

\subsection{Limitations of automated evaluation methods}
As stated in the main paper, these measures are the initial attempts and may give only a ballpark estimation of the qualities under consideration. Content Fidelity and Spatial-layout metrics are only as good as the capability underlying the image-text model - OpenAI's CLIP model \cite{clip}. Technical quality should give an overall measure of technical aberrations like color degradation, unwanted line artifacts, etc. However, we limit ourselves to only measuring the overall noise levels. Aesthetics is a highly subjective aspect of image quality and the CLIP aesthetic model \cite{laion_aesthetic}, though effective, has been trained on a relatively small-sized dataset. Blending \& Harmony in our case is limited to measuring the presence of edges around the boundaries of a segment. Measuring harmony in images is a challenging problem as one needs to also consider the positioning, scale, and coloring of the elements and segments in the context of the overall image. Human preference scoring utilizes ImageReward \cite{xu2023imagereward}, which models the ranking that humans would assign to a group of images.  Although this method performs better than CLIP and BLIP in this aspect, it lacks the explainability of why one image is ranked higher over the other.

Finding better, more precise, and holistic machine-assisted methods for measuring the qualities presented in this paper is an opportunity for future research.

\subsection{Benchmark dataset}
A notable challenge in the automated evaluation of the composite diffusion method is the lack of benchmark datasets. Currently, there do not exist any datasets which consist of segment (or sub-scene) layouts with rich textual descriptions for each segment. Creation of such a dataset is non-trivial using automated methods and requires expensive annotation \cite{Avrahami_2023_CVPR}. 

We handcraft a dataset containing 100 images where we segment each representative image into sub-scenes and manually annotate each sub-scene with a relevant textual description. This enables us to build a benchmark dataset for composite image generation with sufficiently high-quality data. We use this dataset to generate images using our baseline and Composite Diffusion methods. We  use the automated methods described above to get the automated evaluation results (Table \ref{tab:kappa_ablation}).

%% file: supp_04b_artworks.tex
\section{Artworks exercise}
\label{appendix_sec:artworks}
Here, we present the outcomes of a brief collaboration with an artist. The aim of this exercise was to expose the modalities of our system to an external artist and gather early-stage user feedback. To achieve this, we explained the workings of the Composite Diffusion system to the artist and asked her to provide us with 2-3 specific scenarios of artwork that she would normally like to work on. The artist's inputs were given to us on plain sheets of paper in the form of rough drawings of the intended paintings, with clear labels for various objects and sections.

\begin{figure*}[ht!]
    \centering
    \includegraphics[width=\textwidth]{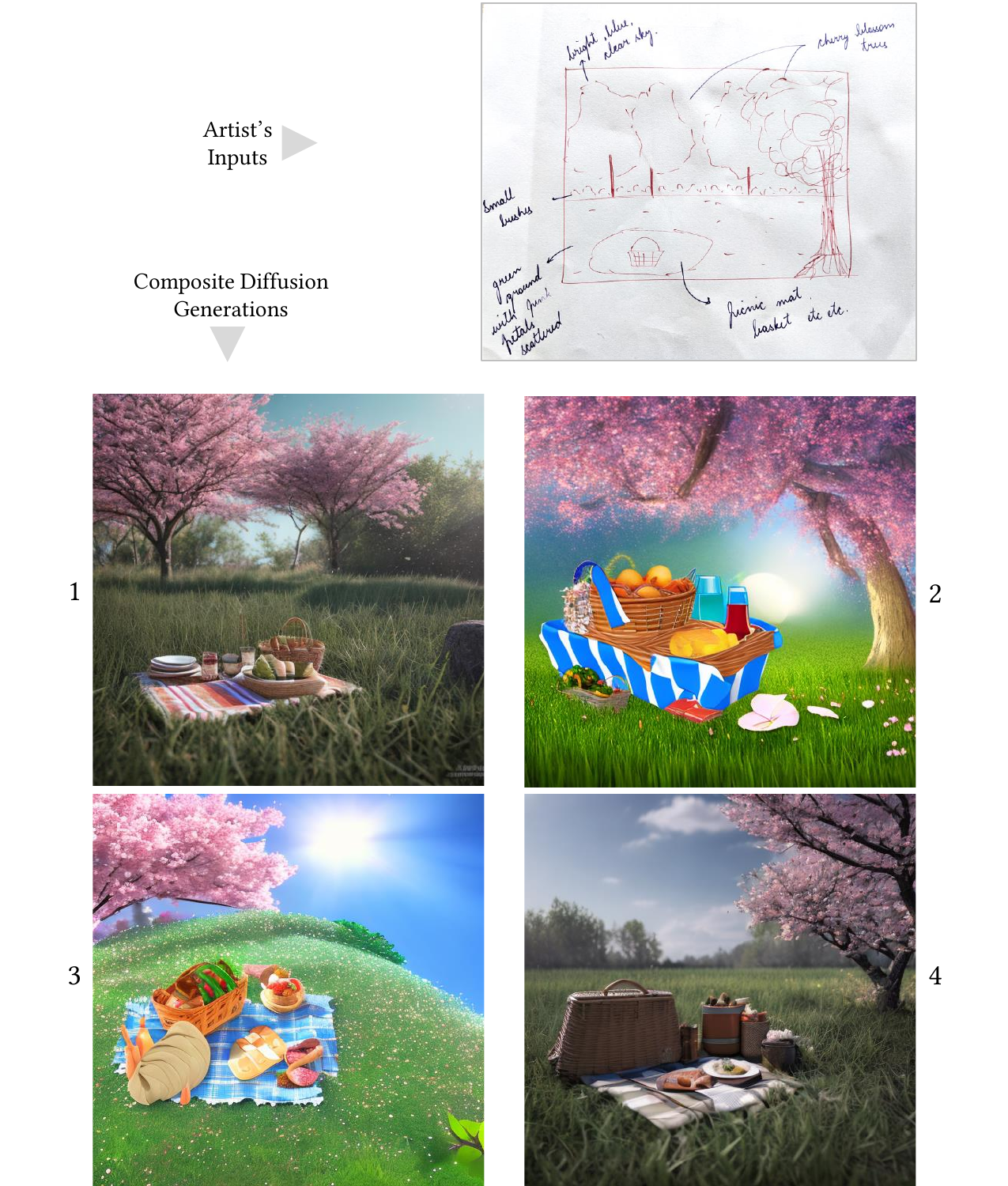}
    \caption {Artwork Exhibit 1a}
    \label{fig:artwork-picnic-a}
\end{figure*}

\begin{figure*}[ht!]
    \centering
    \includegraphics[width=\textwidth]{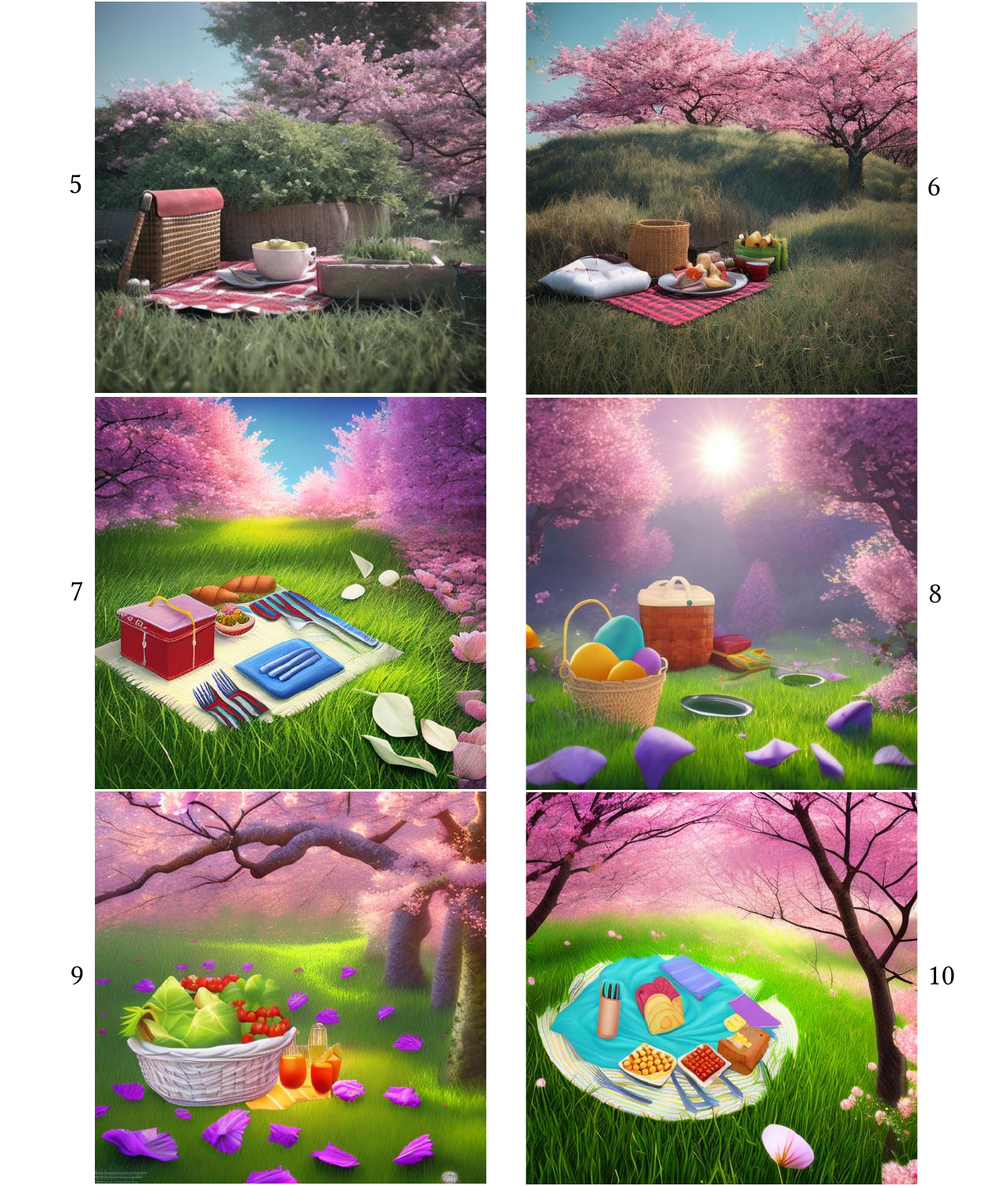}
    \caption {Artwork Exhibit 1b}
    \label{fig:artwork-picnic-b}
\end{figure*}

\begin{figure*}[ht!]
    \centering
    \includegraphics[width=\textwidth]{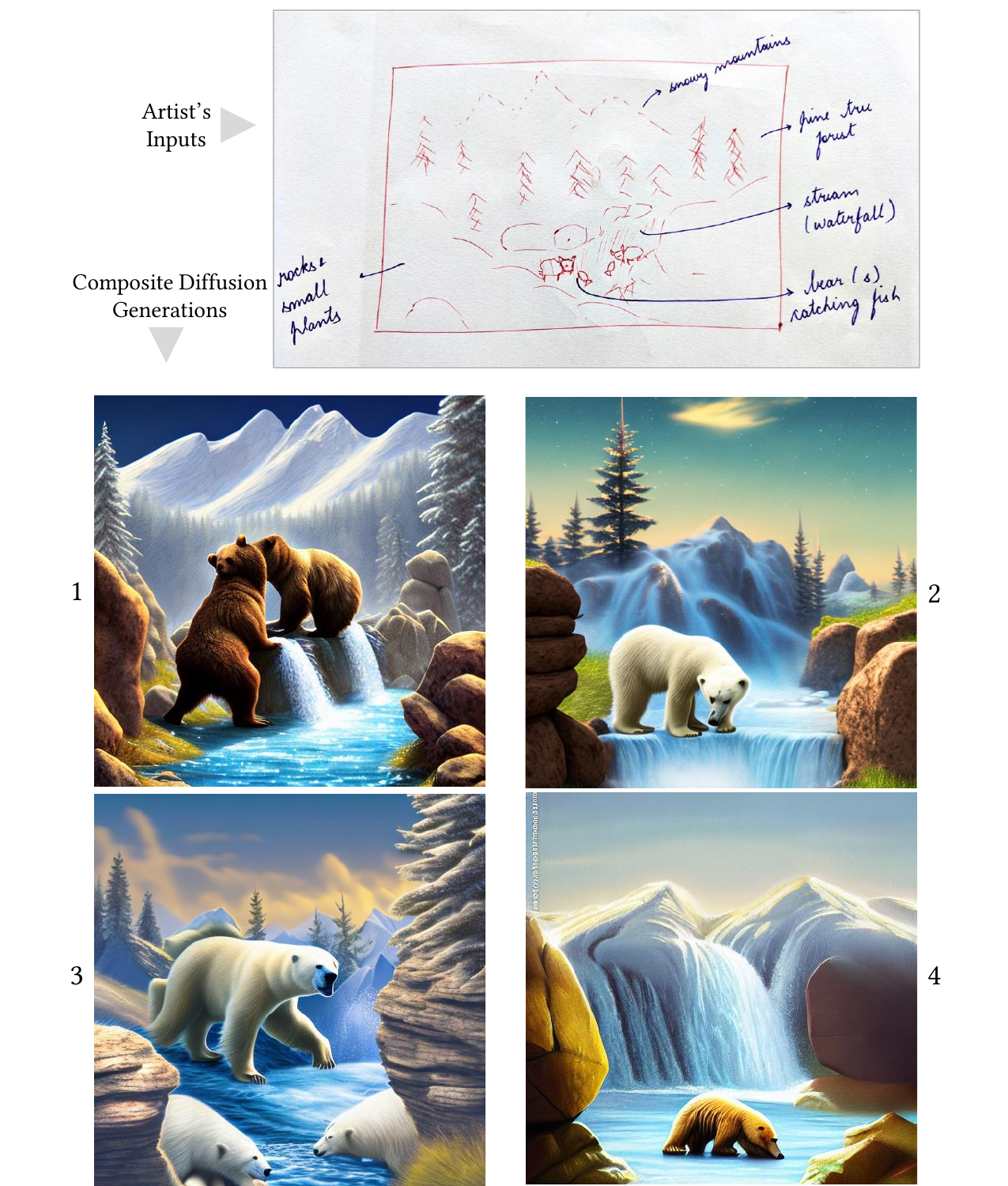}
    \caption {Artwork Exhibit 2a}
    \label{fig:artwork-bear-a}
\end{figure*}

\begin{figure*}[ht!]
    \centering
    \includegraphics[width=\textwidth]{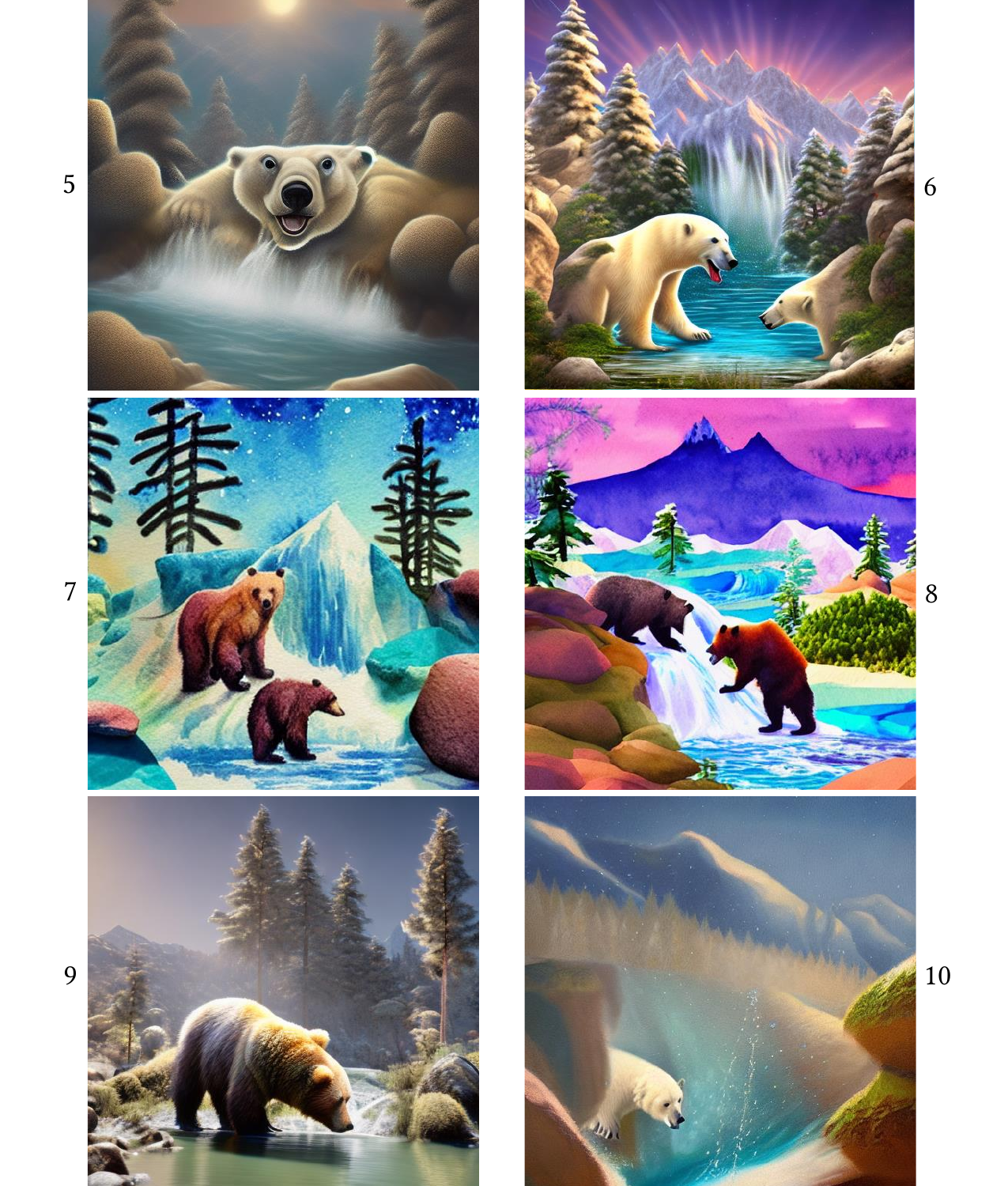}
    \caption {Artwork Exhibit 2b}
    \label{fig:artwork-bear-b}
\end{figure*}

\begin{figure*}[ht!]
    \centering
    \includegraphics[width=\textwidth]{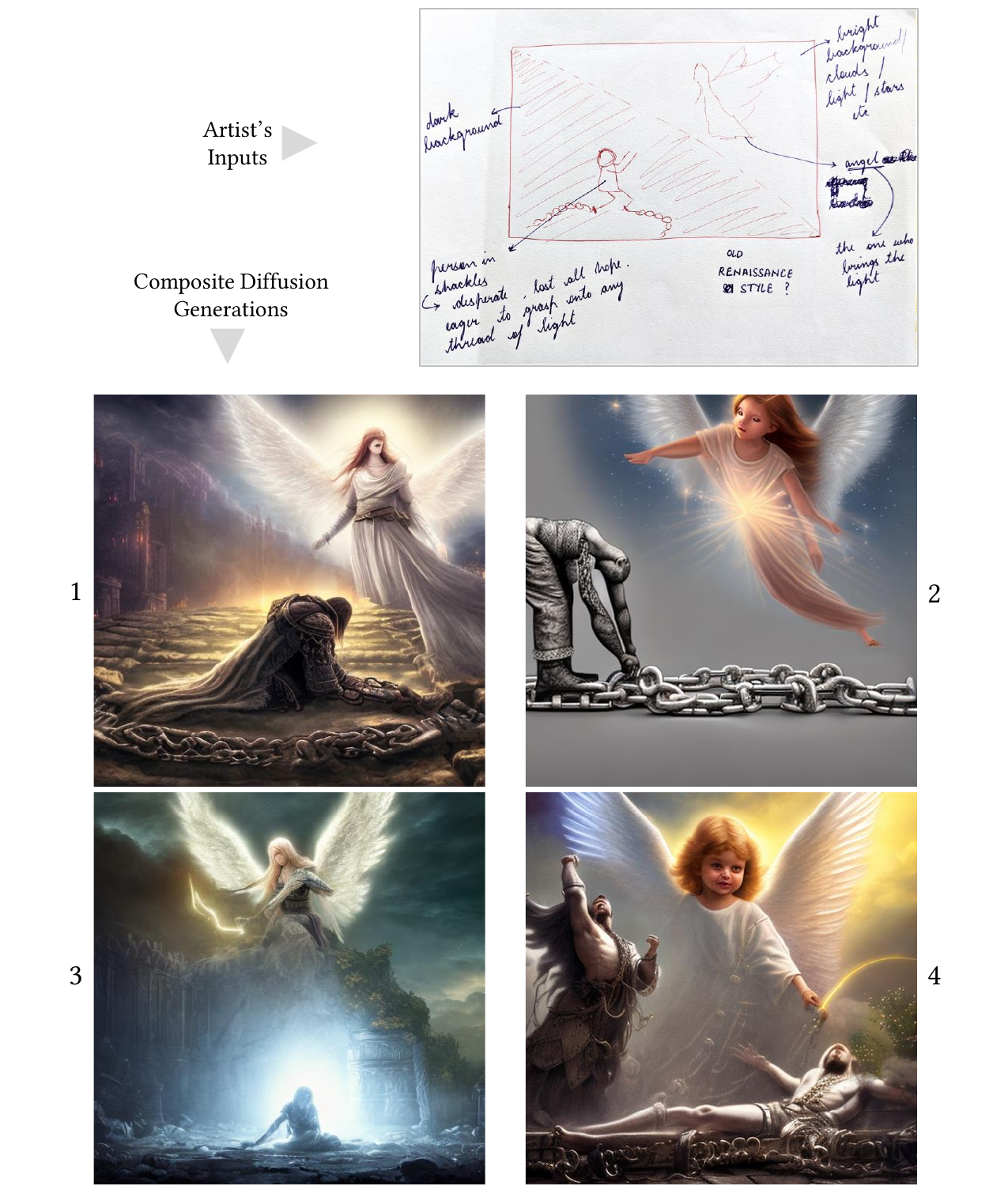}
    \caption {Artwork Exhibit 3a}
    \label{fig:artwork-fairy-a}
\end{figure*}

\begin{figure*}[ht!]
    \centering
    \includegraphics[width=\textwidth]{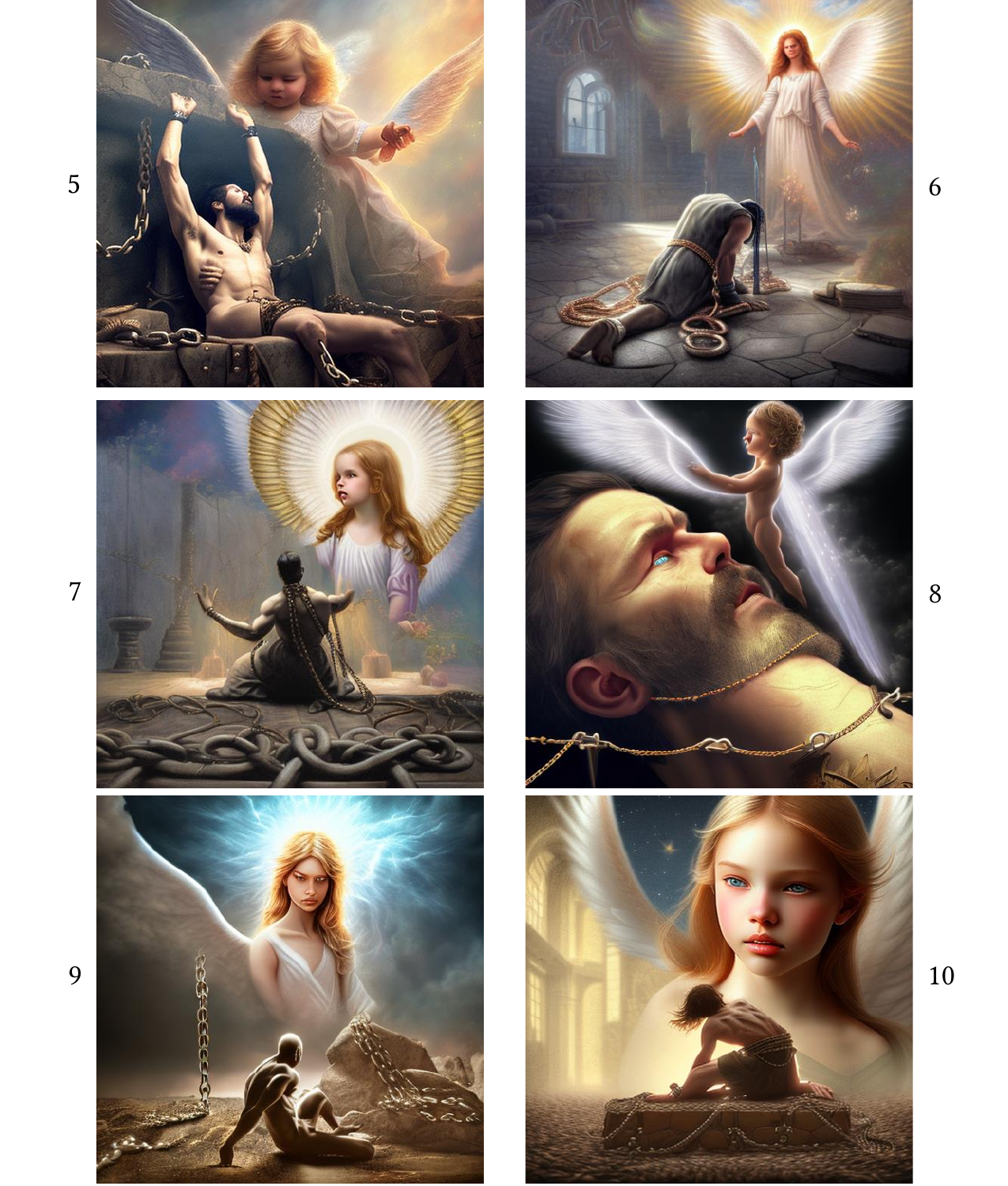}
    \caption {Artwork Exhibit 3b}
    \label{fig:artwork-fairy-b}
\end{figure*}

\begin{figure*}[ht!]
    \centering
    \includegraphics[width=\textwidth]{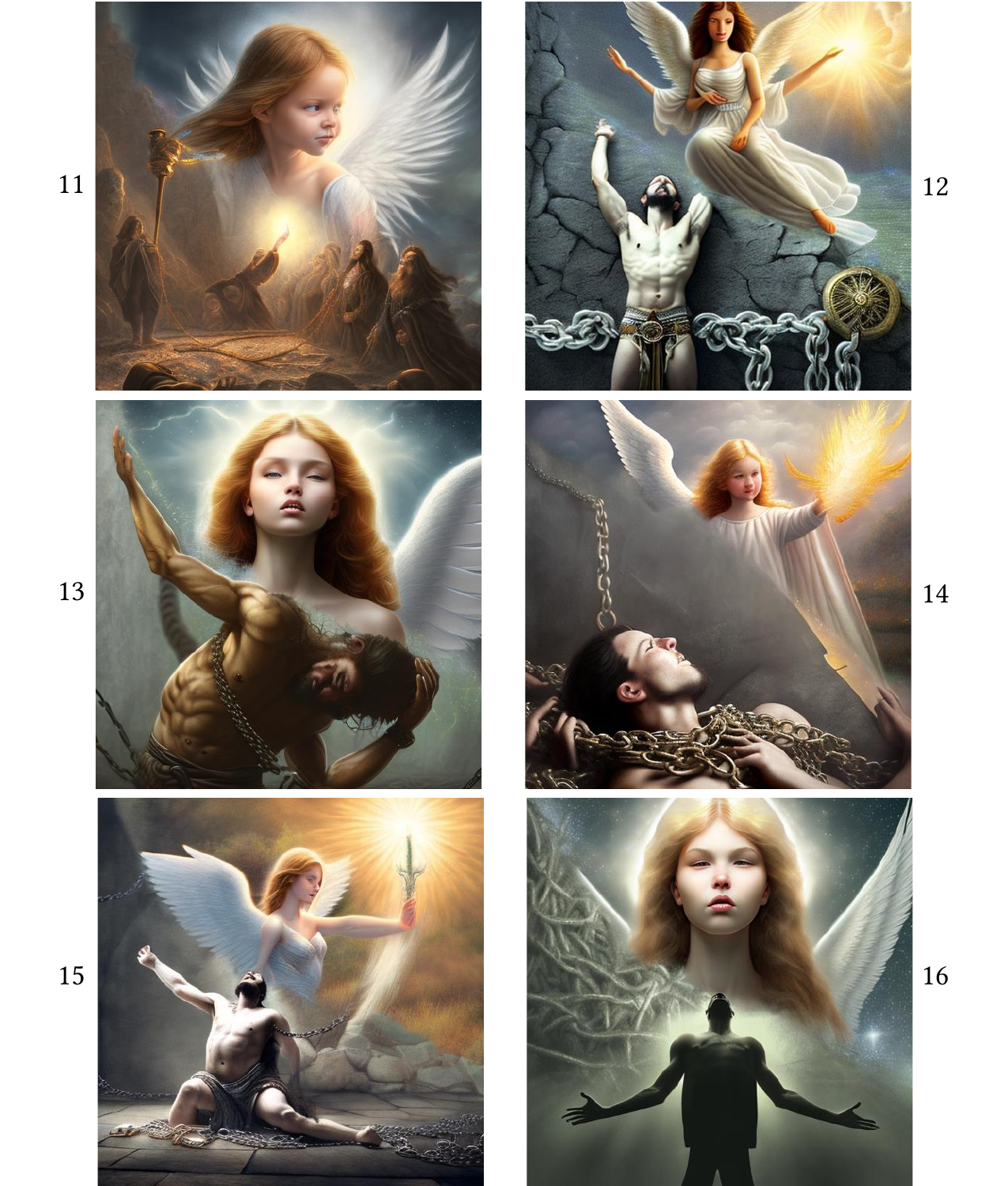}
    \caption {Artwork Exhibit 3c}
    \label{fig:artwork-fairy-c}
\end{figure*}

We converted these inputs into the bimodal input - the free-form segment layouts and text descriptions for each segment.  We did not create any additional control inputs.  We then supplied these inputs to our Composite Diffusion algorithm and performed many iterations with base and a few fine-tuned models, and also at different scaffolding values. The outputs were first shown to a few internal artists for quick feedback, and the final selected outputs were shared with the original artist.  For the final shared outputs, refer to Figures \ref{fig:artwork-picnic-a} and \ref{fig:artwork-picnic-b}  for input 1, Figures \ref{fig:artwork-bear-a} and \ref{fig:artwork-bear-b} for input 2, and Figures \ref{fig:artwork-fairy-a}, \ref{fig:artwork-fairy-b}, and \ref{fig:artwork-fairy-c} for input 3. Please note that the objective here was to produce an artwork with artistically satisfying outputs. So, for some of our generations, we even allowed the outputs which were creative and caught the overall intent of the artist, but did not strictly conform to the prescribed inputs by the artist.

The feedback that we received from the artist at the end of the exercise (as received on Jan 25, 2023), is presented here verbatim:

\textbf{For Artwork 1} (refer to Figures \ref{fig:artwork-picnic-a} and \ref{fig:artwork-picnic-b}):
% \begin{quote}
\textit{``The intended vision was of a lively scene with bright blue skies, picnic blossoms blooming, soft green grass with fallen pink petals, and a happy meal picnic basket. All the images are close enough to the description. Colors are bright and objects fit harmoniously with each other."}
% \end{quote}

\textbf{For Artwork 2} (refer to Figures \ref{fig:artwork-bear-a} and \ref{fig:artwork-bear-b}):
% \begin{quote}
\textit{``The intended vision was of bears in their natural habitat, surrounded by forest trees and snow-clad mountains, catching fish in the stream. Overall, the bears, mountains, trees, rocks, and streams are quite realistic. However, not a single bear could catch a fish. Bear 4 looks like an Afgan hound (dog breed with long hair) and bear 5 itself became a mountain. In image 10, the objects have merged into one another, having ill-defined margins and boundaries."}
% \end{quote}

\textbf{For Artwork 3}(refer to Figures \ref{fig:artwork-fairy-a}, \ref{fig:artwork-fairy-b}, and \ref{fig:artwork-fairy-c}):
% \begin{quote}
\textit{``The intended vision was of an angel - symbolic of hope \& light, salvaging a dejected man, bound by the shackles of hopelessness and despair. Each and every angel is a paragon of beauty. Compared to the heavenly angels, the desolate men look a bit alienish. There is a slight disproportion seen between man and angel in some images. My personal best ones are no. 1,3,6,10."}
% \end{quote}

\begin{figure*}[ht!]
    \centering
    \includegraphics[width=\textwidth]{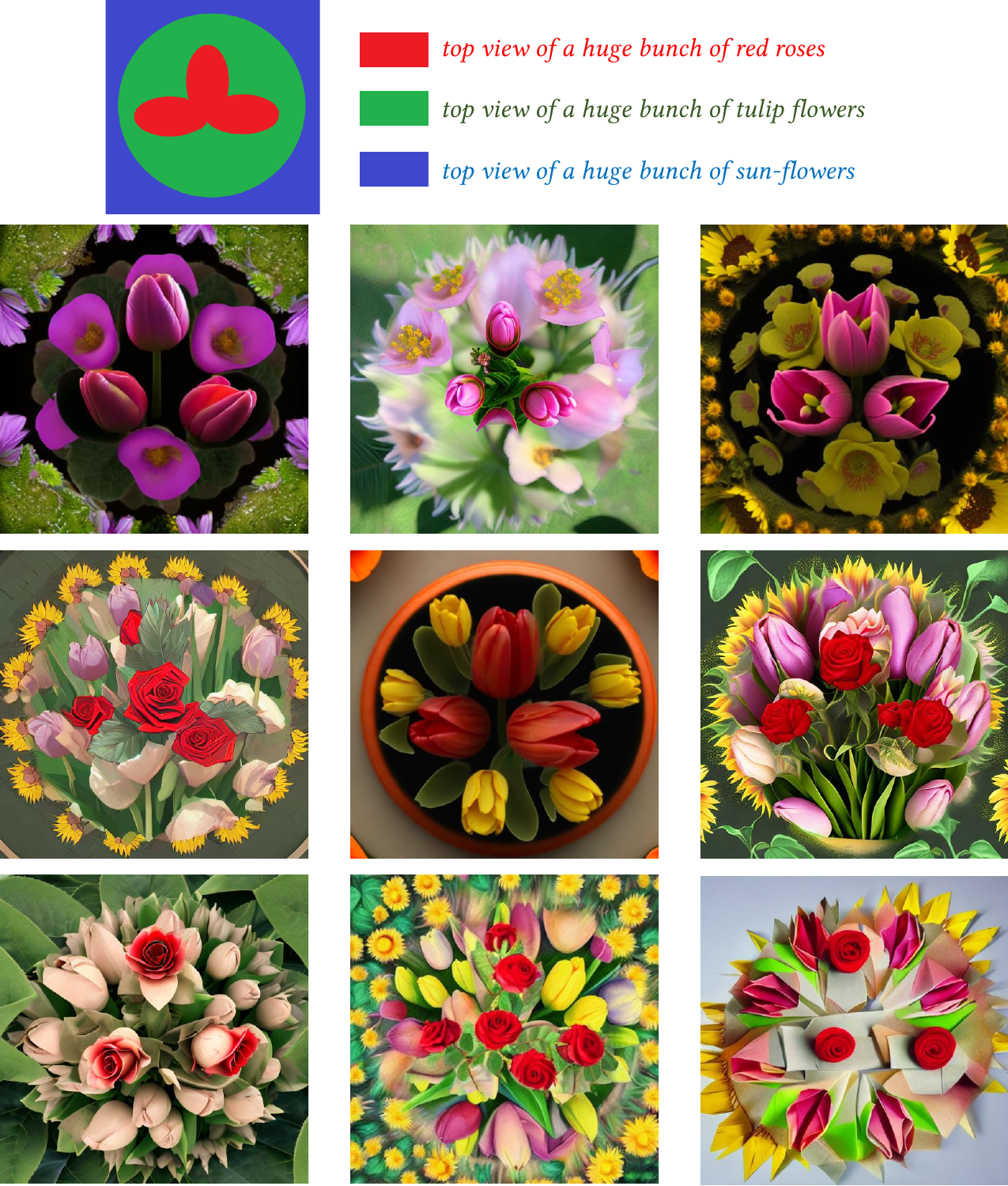}
    \caption {Artwork Exhibit 4: Flower Patterns}
    \label{fig:artwork-flowers}
\end{figure*}

In another case, we wanted to check if our system can be effectively used for artistic pattern generation. Here we gave the system an abstract pattern in the form of a segment layout and specified the objects that we want to fill within those segments. Figure \ref{fig:artwork-flowers} shows a sample output of such an exercise where we fill in the pattern with different forms of flowers.

While the exercise of interactions with the artist and the application of the system for creating the artwork was informal and done in a limited manner, still, it demonstrated to us at an early stage the effectiveness of Composite Diffusion in creating usable art in real-life scenarios. It also validated that the art workflow was simple and intuitive enough to be adopted by people with varied levels of art skills.

